\documentclass{article} %
\usepackage{iclr2025_conference,times}

\usepackage{amsmath,amsfonts,bm}

\def\eqref#1{equation~\ref{#1}}

\def\1{\bm{1}}

\DeclareMathAlphabet{\mathsfit}{\encodingdefault}{\sfdefault}{m}{sl}
\SetMathAlphabet{\mathsfit}{bold}{\encodingdefault}{\sfdefault}{bx}{n}

\usepackage{graphicx}
\usepackage[pagebackref=true,breaklinks=true,colorlinks=true,bookmarks=false]{hyperref}
\definecolor{deepred}{HTML}{940000}
\hypersetup{linkcolor=deepred}
\hypersetup{urlcolor  = [rgb]{0.4,0.15,0.95}}
\hypersetup{citecolor=[rgb]{0.4,0.15,0.95}}

\usepackage{url}
\usepackage{xspace}
\usepackage{slashed}
\usepackage{multirow}
\usepackage{makecell}
\usepackage{algorithm,algorithmicx,algpseudocode}
\usepackage{amsfonts}
\usepackage{booktabs}
\usepackage{siunitx}
\usepackage{amsthm}
\usepackage{wrapfig}
\usepackage{adjustbox}
\usepackage{scalerel}
\usepackage{enumitem}

\usepackage{amssymb}
\usepackage{mathtools}
\usepackage{amsthm}
\usepackage{float}

\usepackage{color}
\usepackage{colortbl}
\definecolor{Gray}{gray}{0.94}

\usepackage{caption}

\usepackage{tocloft}
\usepackage[toc,page,header]{appendix}
\usepackage{minitoc}

\renewcommand \thepart{}
\renewcommand \partname{}

\usepackage[T1]{fontenc}    %

\newlength\savewidth\newcommand\shline{\noalign{\global\savewidth\arrayrulewidth
  \global\arrayrulewidth 1pt}\hline\noalign{\global\arrayrulewidth\savewidth}}

\title{\vspace{-4mm}
\fontsize{15.45pt}{\baselineskip}\selectfont 
Efficient Diversity-Preserving Diffusion Alignment via Gradient-Informed GFlowNets
}

\author{
Zhen Liu\textsuperscript{1,2,3,\textdagger}~~~~Tim Z. Xiao\textsuperscript{2,4,*}~~~~Weiyang Liu\textsuperscript{2,5,*}~~~~Yoshua Bengio\textsuperscript{1}~~~~Dinghuai Zhang\textsuperscript{1,6,\textdagger}
\\[.5mm]
\textsuperscript{1}Mila, Université de Montréal~~~\textsuperscript{2}Max Planck Institute for Intelligent Systems - Tübingen\\
\textsuperscript{3}The Chinese University of Hong Kong (Shenzhen)~~~\textsuperscript{4}University of Tübingen\\
\textsuperscript{5}University of Cambridge~~~\textsuperscript{6}Microsoft Research~~~\textsuperscript{\textdagger}Corresponding author~~~\textsuperscript{*}Equal contribution
}

\newcommand{\expt}{\mathop{\mathbb{E}}}
\newcommand{\Bignorm}[1]{\Bigl\lVert #1 \Bigr\rVert}

\usepackage{xspace}

\newcommand*{\ie}{{\it i.e.}\@\xspace}

\newcommand{\graddb}{$\nabla$-DB\xspace}
\newcommand{\resgraddb}{\textit{residual }$\nabla$-DB\xspace}
\newcommand{\resdb}{\textit{residual} DB\xspace}
\newcommand{\methodname}{$\nabla$-GFlowNet\xspace}

\theoremstyle{plain}
\newtheorem{theorem}{Theorem}%
\newtheorem{proposition}[theorem]{Proposition}
\newtheorem*{proposition*}{Proposition}

\theoremstyle{definition}

\newtheorem{remark}[theorem]{Remark}
\newtheorem*{remark*}{Remark}

\iclrfinalcopy %
\begin{document}

\doparttoc %
\faketableofcontents
\maketitle

{
\vspace{-8mm}
\begin{center}
        \fontsize{9pt}{\baselineskip}\selectfont
        Project page:~{\tt\href{https://nabla-gfn.github.io/}{\textbf{nabla-gfn.github.io}}}
        \vspace{2mm}
\end{center}
}

\begin{abstract}
\vspace{-0.5mm}
While one commonly trains large diffusion models by collecting datasets on target downstream tasks, it is often desired to align and finetune pretrained diffusion models with some reward functions that are either designed by experts or learned from small-scale datasets. Existing post-training methods for reward finetuning of diffusion models typically suffer from lack of diversity in generated samples, lack of prior preservation, and/or slow convergence in finetuning. In response to this challenge, we take inspiration from recent successes in generative flow networks (GFlowNets) and propose a reinforcement learning method for diffusion model finetuning, dubbed Nabla-GFlowNet (abbreviated as \methodname), that leverages the rich signal in reward gradients for probabilistic diffusion finetuning. We show that our proposed method achieves fast yet diversity- and prior-preserving finetuning of Stable Diffusion, a large-scale text-conditioned image diffusion model, on different realistic reward functions.
\end{abstract}

\section{Introduction}
\vspace{-0.5mm}

Diffusion models \cite{ho2020denoising, song2021score, rombach2022high} are a powerful class of generative models that model highly complex data distributions with sequential denoising steps. They prove capable of modeling distributions of images \cite{rombach2022high, dhariwal2021diffusion,qiu2023controlling}, videos \cite{ho2022video,ho2022imagen,blattmann2023stable}, 3D objects \cite{zhang2024clay, poole2023dreamfusion, liu2023meshdiffusion, liu2024gshell,gao2024graphdreamer}, molecules \cite{xu2022geodiff, hoogeboom2022equivariant,liu2024manifold}, languages \cite{lovelace2024latent,saharia2022photorealistic,lou2024discrete} and many others. State-of-the-arts diffusion models for downstream applications are typically large in network size and demand a significant amount of data to train. 

It is often desirable that one finetunes pretrained diffusion models with some reward --- either learned from human preferences as in reinforcement learning from human feedback (RLHF)~\cite{christiano2017deep, ouyang2022training} or designed by human experts \cite{Silver2016MasteringTG,mirhoseini2020chip}. 
While existing methods achieve fast reward convergence~\cite{xu2024imagereward, clark2024directly}, many of them are diversity-lacking,
prior-ignoring,
and/or computationally expensive.

To address this issue, we take the best from both traditional reinforcement learning (RL) and direct reward maximization approaches and propose a novel reinforcement learning objective, dubbed \graddb, that leverages the rich information in reward gradients. Our method is rooted in both generative flow networks (GFlowNets)~\cite{bengio2023gflownet}, a framework and language for generative modeling on a direct acyclic transition graph, and soft RL~\cite{haarnoja2017reinforcement, haarnoja2018soft}. Compared to existing gradient-informed methods, our probabilistic finetuning method achieves better sample diversity. We further propose \methodname with the objective \resgraddb that, by leveraging the structure of diffusion models, 
allows us to perform fast, diversity preserving and prior-preserving amortized finetuning of diffusion models with rather long sampling sequences.

We summarize our major contributions below:

\begin{itemize}[leftmargin=*,nosep]
\setlength\itemsep{0.6em}
    \item We propose \graddb and the pretrained-model-aware variant \resgraddb, the objectives that efficiently leverage the rich information in reward gradients in a principled and probabilistic way.
    \item Our \methodname is the first GFlowNet method that considers first-order information in reward signals, and is closely connected to soft reinforcement learning.
    \item We empirically show that with the proposed \resgraddb objective, we may achieve diversity- and prior-preserving yet fast finetuning and alignment of diffusion models.
\end{itemize}

{
\begin{figure}[H]
    \centering
    \includegraphics[width=0.98\linewidth]{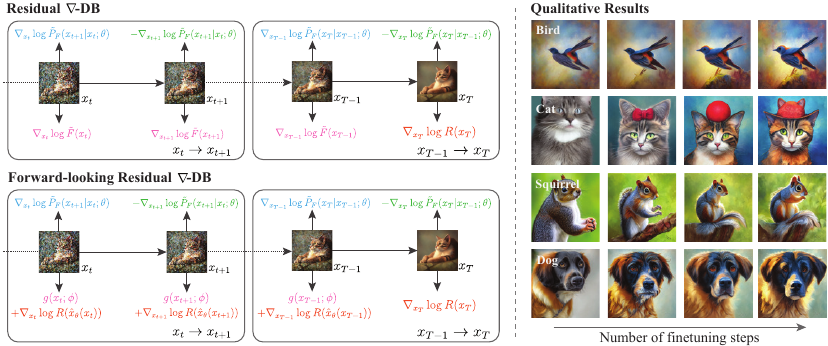}
    \vspace{-2mm}
    \caption{\footnotesize Left: Illustration of the proposed \resgraddb objective, along with its forward-looking variant. The two ``forces'' on each image in each transition $x_t \rightarrow x_{t+1}$ out of a trajectory $\tau = (x_0, x_1, ..., x_T)$ are expected to sum to zero. Green and blue terms represent forward and reverse residual policy scores (respectively), orange terms represent signals from terminal rewards and pink terms represent flow scores or residual flow scores, each of which is defined in Section~\ref{sec:method}. Notice that the reward term on the $x_T$ in the final transition are different from the others. Right: Generated image from a model finetuned with the proposed \resgraddb on the Aesthetic Score reward. The text prompt for each row is shown on the left. The leftmost figure is the image generated by the pretrained model while the rightmost one is from the model finetuned for 200 iterations.}
    \label{fig:teaser2}
    \vspace{-3.5mm}
\end{figure}
}

\section{Preliminaries}
\vspace{-1mm}
\subsection{Diffusion models and RL-based finetuning}

\vspace{-.5mm}

Diffusion models \cite{ho2020denoising, song2021denoising, song2021score} are a class of hierarchical latent models that model the generation process as a sequence of denoising steps. Different from the convention in diffusion model literature, for convenience we adopt in this paper the reverse time of arrow where $x_T$ means samples from the data distribution and the sampling process starts from $t=0$. Under this convention, the probability of the generated samples is:
\begin{align}
\label{eqn:diffusion_likelihood}
    P_F(x_T) = \int_{x_{0:T-1}} P_0(x_0) \prod P_F(x_t | x_{t-1}) dx_{0:T-1}.
\end{align}
Here $P_0(x_0)$ is a fixed initial distribution, $P_F(x_T)$ is the likelihood of the model generating data $x_T$, and the noisy states $x_t$ in the intermediate time steps are constructed by a pre-defined noising process, and the \textit{forward policy} $P_F(x_t | x_{t-1})$ is the denoising step of the diffusion model\footnote{To clarify, 
the ``forward'' and ``backward'' directions in the GFlowNet literature are typically the opposite of those in the diffusion model literature.
}. 
Take DDPM \cite{ho2020denoising} as an example: the corresponding noising process is $q(x_{t-1} | x_{t}) = \mathcal{N}(\sqrt{\alpha_{t-1} / \alpha_{t}}x_t, \sqrt{1 - \alpha_{t-1} / \alpha_t}I)$ and induces $q(x_t | x_{T}) = \mathcal{N}(\sqrt{\alpha_{T-t}}x_T, \sqrt{1 - \alpha_{T-t}}I)$, where $\{\alpha_t\}_t$ is a noise schedule set. With this noising process defined, the training loss is:
\begin{align}
\label{eqn:diffusion_objective}
    \expt_{\substack{t \sim \text{Uniform}(\{1, \ldots,T\}), \epsilon \sim \mathcal{N}(0, I), x_T \sim \mathcal{D}}} w(t) \Bignorm{x_\theta(\sqrt{\alpha_{T-t}}x_T + \sqrt{1 - \alpha_{T-t}} \epsilon, t) - x_T}^2,
\end{align}
where $\mathcal{D}$ is a dataset, $w(t)$ is a certain schedule weighting function, and $x_\theta(x_t, t)$ is a data prediction model that predict the clean data $x_T$ given a noisy data $x_t$ at time step $t$.

The sequential sampling process of diffusion models is Markovian and one could construct an Markov desicion process (MDP) to describe its denoising process. Similar to existing approaches \cite{black2024training,zhang2025improving}, given a Markovian diffusion sampling algorithm, we construct an MDP by treating the noisy sample $(x_t, t)$ at each diffusion inference step $t$ as a state and a denoising step from $x_{t}$ to $x_{t+1}$ as a transition.
Given such an MDP defined, one may finetune diffusion models with techniques like DDPO \cite{black2024training} by collecting on-policy sample trajectories $\{(x_1, ..., x_T)\}$ and optimize the forward (denoising) policy with a given terminal reward $R(x_T)$. We give a more detailed overview of MDP construction and DDPO in Appendix~\ref{sec:mdp_construction}.

\vspace{-0.5mm}
\subsection{RL finetuning in the GFlowNet framework}
\label{sec:gfn_prelim}
\vspace{-0.5mm}

It is well known~\cite{haarnoja2017reinforcement, rafailov2023direct} in the soft RL literature that, by maximizing $\expt_{\pi_\theta}\sum_t [R(s_T) + \frac{1}{\beta} \mathcal{H}(\pi_\theta(\cdot | s_t)]$ in an MDP (with state $s_t$, terminating state $s_T$, action $a_t$, terminal reward $R(\cdot)$, discount rate $\gamma$, policy $\pi_\theta$ and entropy function $\mathcal{H}$), we have $P(x) \propto \exp(\beta R(x))$ for any terminating state $x$. 
Finding optimal policies to sample from $R(\cdot)$ can therefore be achieved with appropriate designs of MDPs. While we may derive our objective with the soft RL language, here we follow a recent work~\cite{zhang2025improving} and adopt the GFlowNet \cite{bengio2023gflownet} framework for generative models that sample terminal states with respect to reward with MDP defined on a directly acyclic graph (DAG). The GFlowNet framework uniquely motivates the introduction of a special intermediate reward associated to a reference backward policy, which is not obvious in the soft RL framework. Furthermore, in modeling distributions with a DAG-based MDP, the GFlowNet framework is more general and perhaps more natural than soft RL. We discuss the connections between two frameworks in Appendix~\ref{sec:soft-rl}.

In a GFlowNet, the forward sampling policy $P_F(s' | s)$ on a DAG from an initial state $s_0$ to a terminal state $s_f$ is paired with a backward ``reference'' policy $P_B(s | s')$. One can view $P_F$ as distributing probability mass from $s$ to its successors, and $P_B$ as moving mass backward from $s'$ to its ancestors. When $P_F$ and $P_B$ are matched, they induce an unnormalized density $F(s)$ at each state, called the flow function. One way to specify the matching between $P_F$ and $P_B$ is through the following:
\vspace{-0.2cm}
\paragraph{Detailed Balance (DB).} A valid GFlowNet with a forward policy $P_F(s' | s)$, a backward policy $P_B(s | s')$, and a flow function $F(s)$ satisfies the following DB condition for all transition $(s\to s')$
\begin{align}
\label{eqn:db_condition_orig}
    P_F(s' | s) F(s) = P_B(s | s') F(s').
\end{align}
Hence we have the following GFlowNet DB loss in the logarithm probability space:
\begin{align}
    L_\text{DB}(s, s') = \Big(\log P_F(s' | s) + \log F(s) - \log P_B(s | s') - \log F(s')\Big)^2
\end{align}
with an extra terminal constraint $F(s_f) = R(s_f)$ to incorporate the target reward information.

In the context of time-indexed sampling processes such as diffusion models, the transition graph of states $s\triangleq(x_t, t)$ is naturally acyclic, as it adheres to the arrow of time~\citep{zhang2022unifying}. With slight abuse of terminology that we use $x_t$ to represents the tuple $(x_t, t)$ when necessary, for time-indexed settings the \textit{forward policy} is $P_F(x_{t+1} | x_t)$, the \textit{backward policy} is $P_B(x_t | x_{t+1})$, and the \textit{flow function} is $F(x_t)$\footnote{We write $F_t(x_t)$ as $F(x_t)$ for the sake of notation simplicity.}. The corresponding DB condition is therefore
\begin{align}
\label{eqn:db_condition}
    P_F(x_{t+1} | x_t) F(x_t) = P_B(x_t | x_{t+1}) F(x_{t+1}).
\end{align}
To finetune a diffusion model with DB losses \cite{zhang2025improving}, one can simply set $P_F(x_{t+1} | x_t)$ to be the sampling process 
and fix $P_B(x_t | x_{t+1})$ to be the noising process used by the pretrained diffusion model. If $P_B$ is set to be fixed, we can similarly derive this condition using soft Q-learning~\cite{haarnoja2017reinforcement}, with details provided in Appendix~\ref{sec:soft-rl}.

\vspace{-0.5mm}
\section{Reward Finetuning via Gradient-informed GFlowNets}
\label{sec:method}
\vspace{-0.5mm}
\subsection{\graddb: the gradient-informed Detailed Balance}
\vspace{-0.5mm}

In our setting, we do not have access to any dataset of images, but are given an external positive-valued reward function $R(\cdot)$ to which a generative model is trained to adapt. 
While a typical GFlowNet-based algorithm can effectively achieve this objective with diversity in generated samples, it only leverages the zeroth-order reward information and does not leverage any differentiability of the reward function.
Yet, whenever the reward gradients are available, it is often beneficial 
to incorporate them into the finetuning objective since they help navigate the finetuned model on the optimization landscape and may significantly accelerate the optimization process. 
We are therefore motivated to develop \methodname, a method that builds upon GFlowNet-based algorithms to take full advantage of the reward gradient signal.
To achieve this, we take derivatives on the logarithms of both sides of the DB condition (logarithm of Equation~\ref{eqn:db_condition}) with respect to $x_{t+1}$ and obtain a necessary condition, which we call the forward\footnote{This is because the derivative is taken with respect to $x_{t+1}$. 
} \graddb condition:
\begin{align}
\label{eqn:graddb-condition}
    \nabla_{x_{t+1}}\log P_F(x_{t+1} | x_t) = \nabla_{x_{t+1}} \log P_B(x_t | x_{t+1}) + \nabla_{x_{t+1}} \log F(x_{t+1}),
\end{align}
and hence the corresponding forward \graddb objective $L_{\overrightarrow{\nabla}\text{DB}}(x_t, x_{t+1})$ to be
\begin{align}
\label{eqn:grad-db}
    \Bignorm{ \nabla_{\small{x_{t+1}}}\log P_F(x_{t+1} | x_t) - \nabla_{\small{x_{t+1}}} \log P_B(x_t | x_{t+1}) - \nabla_{\small{x_{t+1}}} \log F(x_{t+1}) }^2,
\end{align}
with the terminal flow loss on the logarithm scale
\begin{align}
\label{eqn:grad-db-terminal}
    L_{\nabla\text{DB-terminal}}(x_T) = \Bignorm{
        \nabla_{x_T} \log F(x_T) - \beta\nabla_{x_T} \log R(x_T)
    }^2,
\end{align}
where $\beta$ is a temperature coefficient and serve as a hyperparameter in the experiments.
Notice that, by taking derivatives on the logarithms, we obtain the (conditional) score function $\nabla_{x_{t+1}} \log P_F(x_{t+1} | x_t)$ of the finetuned diffusion model. Indeed, the \graddb loss is closely related to a Fisher divergence, also known as Fisher information score \cite{johnson2004information} (see Appendix~\ref{sec:fisher-div}).

Similarly, by taking the derivative of both sides in Equation~\ref{eqn:db_condition} with respect to $x_{t}$, one obtains the reverse \graddb objective:
\begin{align}
\label{eqn:graddb-rev}
     L_{\overleftarrow{\nabla}\text{DB}}(x_{t}, x_{t+1}) = \Bignorm{
        \nabla_{x_{t}}\log P_F(x_{t+1} | x_t) - \nabla_{x_{t}} \log P_B(x_t | x_{t+1}) + \nabla_{x_{t}} \log F(x_{t})
    }^2.
\end{align}
Such \methodname objectives constitute a valid GFlowNet algorithm (see the proof in Section~\ref{sec:proof_prop_validgfn}):

\begin{proposition}
\label{prop:valid_gfn}
If $L_{\overrightarrow{\nabla}\text{DB}}(x_t, x_{t+1})= L_{\overleftarrow{\nabla}\text{DB}}(x_{t}, x_{t+1})=0$ for any denoising transition $(x_t, x_{t+1})$ over the state space and $L_{\nabla\text{DB-terminal}}(x_T)=0$ for all terminal state $x_T$, then the resulting forward policy generate samples $x_T$ with probability proportional to the reward function $R(x_T)^\beta$.
\end{proposition}

\begin{remark}
The original detailed balance condition propagates information from the reward function to each state flow function in the sense of $F(x_{t+1}) \to (F(x_t), P_F(x_{t+1}|x_t))$, assuming the backward (noising) policy is fixed (\ie, there is no learning component in the diffusion noising process). In our case, if we take a close look at Equation~\ref{eqn:grad-db}, we can see that $L_{\overrightarrow{\nabla}\text{DB}}(x_t, x_{t+1})$ could propagate the information from $F(x_{t+1})$ to the forward policy $P_F(x_{t+1}|x_t)$ but not $F(x_t)$. 
\end{remark}

\begin{remark}
Compared to previous GFlowNet works which use a scalar-output network to parameterize the (log-) flow function, in \methodname we can directly use a U-Net~\citep{ronneberger2015unetcn}-like architecture that (whose output and input shares the same number of dimension) to parameterize $\nabla \log F(\cdot)$, which potentially provides more modeling flexibility. Furthermore, it is possible to initialize $\nabla \log F(\cdot)$ with layers from the pretrained model so that it can learn upon known semantic information.
\end{remark}

\subsection{Residual \graddb for reward finetuning of pretrained models}
\label{residual-finetune}

With the \graddb losses, one can already finetune a diffusion model to sample from the reward distribution $R(x)$. 
However, the finetuned model may eventually over-optimize the reward and thus forget the pretrained prior (\textit{e.g.}, how natural images look like).
Instead, similar to other amortized inference work~\cite{zhao2024twist, wu2024practical}, we consider the following objective with an augmented reward: 
\begin{align}
\label{eqn:finetuning-objective}
P_F(x_T) \propto R(x_T)^\beta P_F^\#(x_T),
\end{align}
where $R(x)$ is the positive-valued reward function, $\beta$ is the temperature coefficient, $P_F^\#(x_T)$ is the marginal distribution of the pretrained model\footnote{We use the notation of $\#$ to indicate quantities of the pretrained model.} and $P_F(x_T)$ is the marginal distribution of the finetuned model (as defined in Equation~\ref{eqn:diffusion_likelihood}).

Because both the finetuned and pretrained model share the same backward policy $P_B$ (the noising process of the diffusion models), 
we can remove the $P_B$ term and obtain the forward \resgraddb condition
by subtracting the forward \graddb equation for the pretrained model from the that of the finetuned model:
\begin{align}
\label{eqn:grad-db-res-cond}
        \underbrace{
                    \nabla_{x_{t+1}}\log P_F(x_{t+1} | x_{t})  \!-\! \nabla_{x_{t+1}}\log P_F^\#  (x_{t+1} | x_{t}) 
            }_{\scalebox{0.8}{$\nabla_{x_{t+1}} \log \tilde{P}_F(x_{t+1} | x_t)$}: \ \text{\small residual policy score function}}
        \!=\!
        \underbrace{
                \nabla_{x_{t+1}} \log F(x_{t+1}) \!-\! \nabla_{x_{t+1}} \log F^\#(x_{t+1}) 
        }_{\scalebox{0.8}{$\nabla_{x_{t+1}} \log \tilde{F}(x_{t+1})$}: \ \text{\small residual flow score function}}.
\end{align}

With the two residual terms defined above, we obtain the forward \resgraddb objective: 
\begin{align}
\label{eqn:grad-db-res}
    L_{\overrightarrow{\nabla}\text{DB-res}}(x_{t}, x_{t+1}) =
    \Bignorm{ 
        \nabla_{x_{t+1}} \log \tilde{P}_F(x_{t+1} | x_t) - \nabla_{x_{t+1}} \log \tilde{F}(x_{t+1}),
    }^2.
\end{align}
Similarly, we have the reverse \resgraddb loss:
\begin{align}
\label{eqn:grad-db-res-rev}
    L_{\overleftarrow{\nabla}\text{DB-res}}(x_{t}, x_{t+1})  
    =
    \Bignorm{ 
        \nabla_{x_{t}} \log \tilde{P}_F(x_{t+1} | x_t) + \nabla_{x_{t}} \log \tilde{F}(x_{t})
    }^2.
\end{align}
The terminal loss of the residual \graddb method stays the same form as in Equation~\ref{eqn:grad-db-terminal}
\begin{align}
     L_{\nabla\text{DB-terminal}}(x_T) = \Bignorm{
        \nabla_{x_T} \log \tilde{F}(x_T) - \beta \nabla_{x_T} \log R(x_T)
    }^2.
\end{align}
\begin{proposition}
\label{prop:valid-res-gfn}
If $L_{\overrightarrow{\nabla}\text{DB-res}}(x_t, x_{t+1})= L_{\overleftarrow{\nabla}\text{DB-res}}(x_{t}, x_{t+1})=0$ for any denoising transition $(x_t, x_{t+1})$ over the state space and $L_{\nabla\text{DB-terminal}}(x_T)=0$ for all terminal state $x_T$, then the resulting forward policy generate samples $x_T$ with probability proportional to $R(x_T)^\beta  P_F^\#(x_T)$.
\end{proposition}

\begin{remark}
\label{remark:relative-tb}
We point out that perform the same way of deriving Equation~\ref{eqn:grad-db-res-cond}, \ie, subtraction between GFlowNet conditions from the finetuned and pretrained model, on the DB condition without gradient, we can obtain a \textit{residual} DB condition $\tilde F(x_t)P_F(x_{t+1}|x_t)=P_F^\#(x_{t+1}|x_t)\tilde F(x_{t+1})$.
Multiplying this condition across time and eliminate the term of intermediate $\tilde F(x_t)$ will lead to the objective derived in the relative GFlowNet work~\citep{venkatraman2024amortizing} as shown in Section~\ref{sec:relationship-tb}, which is a prior paper that proposes to work on reward finetuning GFlowNets with a given pretrained model.
\end{remark}

\begin{remark}
One may completely eliminate the need for any residual flow score function with the \resgraddb conditions of both directions: $ \nabla_{x_{t+1}} \log \tilde{P}_F(x_{t+1} | x_{t}) = -\nabla_{x_{t+1}} \log \tilde{P}_F(x_{t+2} | x_{t+1})$.
The bidirectional \resgraddb condition can be analogously understood as the balance condition of two forces from $x_t$ and $x_{t+2}$ acting on $x_{t+1}$: if not balanced, one can locally find some other $x_{t+1}$ that makes both transitions more probable.
\end{remark}

\vspace{-0.5cm}
\paragraph{Flow reparameterization through forward-looking (FL) trick.}
Though mathematically valid, the bidirectional pair of \graddb conditions suffers from inefficient credit assignment for long sequences, a problem commonly observed in RL settings \cite{Sutton1988LearningTP,van2018deep}. Instead, we may leverage the priors we have from the pretrained diffusion model to speed up the finetuning process and consider the individual conditions for the forward and reverse directions. Specifically, we employ the forward-looking (FL) technique for GFlowNets \cite{pan2023better,zhang2025improving} and parameterize the residual flow score function with a ``baseline'' of the ``one-step predicted reward gradient'':
\begin{align}
    \label{eqn:forward-looking}
    \nabla_{x_t} \log \tilde{F}(x_t) \triangleq \beta \gamma_t \nabla_{x_t} \log\underbrace{R(\hat{x}_\theta(x_t))}_{\small\text{predicted reward}} +\ g_\phi(x_t)
\end{align}
where $\gamma_t$ is the scalar to control the strength of forward looking with the constraint $\gamma_T=1$ and $g_\phi(x_t)$ is the \textit{actual} neural network with parameters  satisfying the terminal constraint $g_\phi(x_T) = 0$. Here $\hat{x}_\theta(\cdot)$ is the one-step clean data prediction defined in Equation~\ref{eqn:diffusion_objective}.
With the FL technique, one achieves faster convergences since it sets a better initialization for the residual flow score function than a na\"ive zero or random initialization~\cite{pan2023better}.

We therefore obtain the forward-looking version of \resgraddb losses of both directions:
\begin{align}
\label{eqn:grad-db-fl}
    L_{\overrightarrow{\nabla}\text{DB-FL-res}}(x_t, x_{t+1}) = 
    \Bignorm{ 
        \nabla_{x_{t+1}}\log \tilde{P}_F(x_{t+1} | x_t; \theta)
        - \Big[\beta \gamma_t \nabla_{x_{t+1}} \log R(\hat{x}_{\theta}(x_{t+1})) + g_\phi(x_{t+1})\Big]
    }^2.
\end{align}
\vspace{-0.5cm}
\begin{align}
\label{eqn:graddb-rev-res-FL}
     L_{\overleftarrow{\nabla}\text{DB-FL-res}}(x_{t}, x_{t+1}) = \Bignorm{
        \nabla_{x_{t}}\log \tilde{P}_F(x_{t+1} | x_t) + \Big[\beta \gamma_t \nabla_{x_{t}} \log R(\hat{x}_{\theta}(x_{t})) + g_\phi(x_{t})\Big]
    }^2.
\end{align}
Moreover, the corresponding terminal loss objective now becomes 
\begin{align}
     L_{\nabla\text{DB-FL-terminal}}(x_T) = \Bignorm{
        \nabla_{x_T} \log \tilde F(x_T) - \beta \gamma_t \nabla_{x_T} \log R(x_T)
    }^2=\Bignorm{
        g_\phi(x_T)
    }^2,
\end{align}
which indicates that the actual parameterized flow network $g_\phi$ should take a near-zero value for terminal states $x_T$. 
The total loss on a collected trajectory $\tau=(x_0, x_1, ..., x_T)$ is therefore
\begin{align}
    \sum_{t} \Big[ w_F(t) L_{\overrightarrow{\nabla}\text{DB-FL-res}}(x_t, x_{t+1}) + w_B(t) L_{\overleftarrow{\nabla}\text{DB-FL-res}}(x_{t}, x_{t+1}) \Big] + L_{\nabla\text{DB-FL-terminal}}(x_T)
\end{align}
where $w_F(t)$ and $w_B(t)$ are scalar weights to control the relative importance of each term.

We summarize the resulting algorithm in Algorithm~\ref{alg:resgraddb} in Appendix~\ref{alg:resgraddb}.

\textbf{Choice of FL scale.} 
Na\"ively setting $\gamma_t=1$ 
can be aggressive especially when the reward scale $\beta$ and the learning rate are set to a relatively high value. Inspired by the fact that in diffusion models $F_t(x_t)$ can be seen as $R(x)$ smoothed with a Gaussian kernel, we propose to set $\gamma_t = \alpha_{T-t}$. 

\section{Experiments and Results}
\vspace{-1mm}
\subsection{Baselines}
\vspace{-1mm}
For gradient-free methods, we consider DAG-DB \cite{zhang2025improving} (\textit{i.e.}, GFlowNet finetuning with the DB objective) and DDPO \cite{black2024training}. Since the original DB objective aims to finetune with $R^\beta(x)$ instead of $P_F^\#(x_T) R^\beta(x_T)$, we also consider the forward-looking residual DB loss, defined as
\begin{align}
\label{eqn:residual-db}
    L_\text{DB-FL-res}(x_{t}, x_{t+1}) = \Big( \log \tilde{P}_F(x_{t+1} | x_t) - & \beta \log R(\hat{x}_\theta(x_{t+1})) - g_\phi(x_{t+1})
    \notag \\
    + & \beta \log R(\hat{x}_\theta(x_{t})) + g_\phi(x_{t})
    \Big)^2.
\end{align}

For other gradient-aware finetuning methods, we consider ReFL \cite{xu2024imagereward} and DRaFT \cite{clark2024directly}. ReFL samples a trajectory and stops at some random time step $t$, with which it maximizes $R(\hat{x}_\theta(x_t)$ where $\hat{x}_\theta(\cdot)$ is the one-step sample prediction function and $x_t$ is the computed via denoising $\slashed{\nabla}(x_{t-1})$ with $\slashed{\nabla}$ being the stop-gradient operation. Differently, DRaFT samples time step $T-K$ (typically $K=1$) and expand the computational graph of DDPM from $\slashed{\nabla}(x_{T-K})$ to $x_T$ so that $R(x_T)$ can be backpropagated to $x_{T-K}$, with all $x_t$ in the previous time steps removed from this computational graph. A variant of DRaFT called DRaFT-LV performs a few more differentiable steps of ``noising-denoising'' on the sampled $x_T$ before feeding it into the reward function $R(\cdot)$. We follow the paper of DRaFT and use only one ``noising-denoising`` step: $x_{T-1}' \sim P_B(x_{T-1}' | x_T)$ and $x_T' \sim P_F(x_T' | x_{T-1}')$.

\subsection{Reward functions, prompt datasets and metrics}

For the main experiments, we consider two reward functions: Aesthetic Score \cite{laion_aesthetic_2024}, Human Preference Score (HPSv2) \cite{wu2023humanv2, wu2023human} and ImageReward~\cite{xu2024imagereward}, all of which trained on large-scale human preference datasets such as LAION-aesthetic~\cite{laion_aesthetic_2024} and predict the logarithm of reward values. For base experiments with Aesthetic Score, we use a set of 45 simple animal prompts as used in DDPO \cite{black2024training}; for those with HPSv2, we use photo+painting prompts from the human preference dataset (HPDv2) \cite{wu2023humanv2}. To measure the diversity of generated images, we follow Domingo-Enrich et al. \cite{domingoenrich2024adjoint} and compute the variance of latent features extracted from a batch of generated images (we use a batch of size $64$). Using the same set of examples, we evaluate the capability of prior preservation, we compute the per-prompt FID score between images generated from the pretrained model and from the finetuned model and take the average FID score over all evaluation prompts.

\begin{figure}[t]
    \centering
    \vspace{-5mm}
    \includegraphics[width=0.95\linewidth]{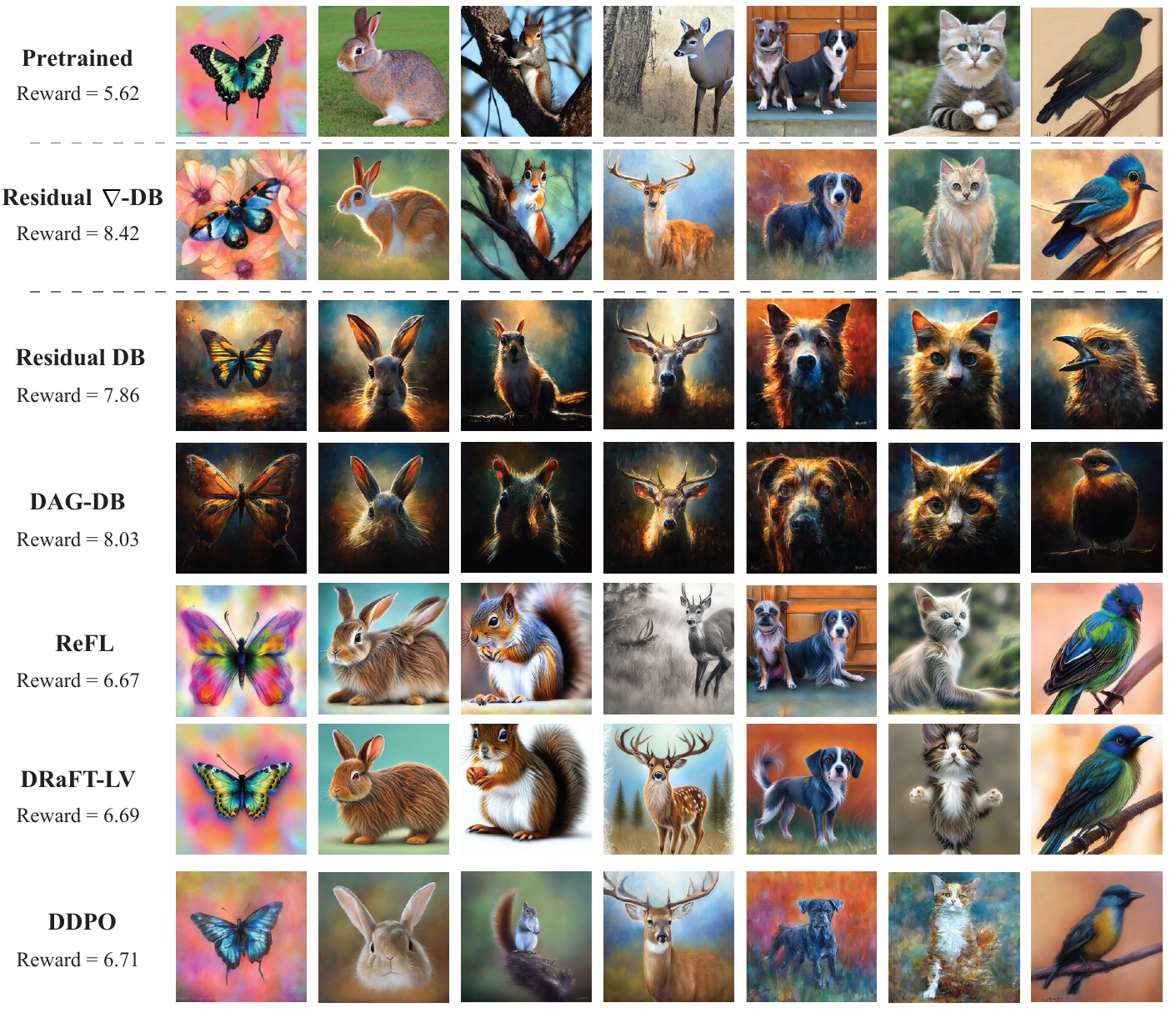}
    \vspace{-2mm}
    \caption{\footnotesize Comparison between images generated by models finetuned with different methods for a maximum of $200$ update steps. For each method, we pick the model trained that produces images with the highest rewards without semantic collapse among all model checkpoints, as methods like ReFL and DRaFT-LV easily collapses (as illustrated in Fig.~\ref{fig:robustness}). For each method, we show the average reward of the corresponding presented images.}
    \label{fig:main_qualitative_comparison}
    \vspace{-5mm}
\end{figure}

\subsection{Experiment settings}

For all methods, we use $50$-step DDPM sampler~\cite{ho2020denoising} to construct the MDP. StableDiffusion-v1.5 \cite{rombach2022high} is used as the base model. For finetuning diffusion model policies, we use LoRA~\cite{hu2022lora} with rank $8$.
The residual flow score function in \resgraddb is set to be a scaled-down version of the StableDiffusion U-Net, whereas the flow function (in DAG-DB and \resdb) is set to be a similar network but without the U-Net decoding structure (since the desired output is a scalar instead of an image vector). Both networks are initialized with tiny weights in the final output layers.

As the landscape of $R(x)$ can be highly non-smooth, we approximate $\nabla_{x_t} \log R(\hat{x}_\theta(x_t))$ with $\mathbb{E}_{\epsilon\sim\mathcal{N}(0, c)} \nabla_{x_t} \log R(\hat{x}_\theta(x_t) + \epsilon)$ where $c$ is a tiny constant. For StableDiffusion \cite{rombach2022high}, since the diffusion process runs in the latent space, the reward function is instead $\mathbb{E}_{\epsilon\sim\mathcal{N}(0, c)} \nabla_{x_t} \log R(\text{decode}(\hat{x}_\theta(x_t) + \epsilon))$ in which $\text{decode}(\cdot)$ is the pretrained (and frozen) VAE decoder and $c$ is set to $2 \times 10^{-3}$, slightly smaller than one pixel (i.e., $1 / 255$). We approximate this expectation with $3$ independent samples for each transition in each trajectory. For all experiments, we try $3$ random seeds. We set $w_F(t) = 1$ for all $t$'s and unless otherwise specified $w_B(t) = 1$.

To stabilize the training process of our method (\resgraddb), we follow the official repo of DAG-DB~\cite{zhang2025improving} and uses the following output regularization: $\lambda \lVert \epsilon_\theta(x_t) - \epsilon_{\theta^\dagger}(x_t) \rVert^2$ where $\theta^\dagger$ is the diffusion model parameters in the previous update step\footnote{Essentially a Fisher divergence between the pretrained and the finetuned distributions conditioned on $x_t$. Similar regularizations with KL divergence has been seen in RL algorithms like TRPO \cite{schulman2015trust} and PPO \cite{schulman2017proximal}.}. For \resgraddb, We set the output regularization strength $\lambda = 2000$ in Aesthetic Score experiments and $\lambda = 5000$ in HPSv2 and ImageReward experiments.  For all experiments with \resgraddb, we set the learning rate to $1 \times 10^{-3}$ and ablate over a set of choices of reward temperature $\beta$, in a range such that the reward gradients are more significant than the residual policy score function $\nabla_{x_t} \log P_F(x_{t+1} | x_t)$ of the pretrained model. For HPSv2 and ImageReward experiments, we set $\beta$ to be $500000$ and $10000$, respectively. For each epoch, we collect $64$ generation trajectories for each of which we randomly shuffle the orders of transitions. We use the number of gradient accumulation steps to $4$ and for each $32$ trajectories we update both the forward policy and the residual flow score function. For \resgraddb in most of the experiments, we for training sub-sample $10\%$ of the transitions in each collected trajectory by uniformly sample one transition in each of the uniformly split time-step intervals but ensuring that the final transition step always included.

\begin{wrapfigure}{r}{0.41\textwidth}
    \centering
    \vspace{-6mm}
    \includegraphics[width=\linewidth]{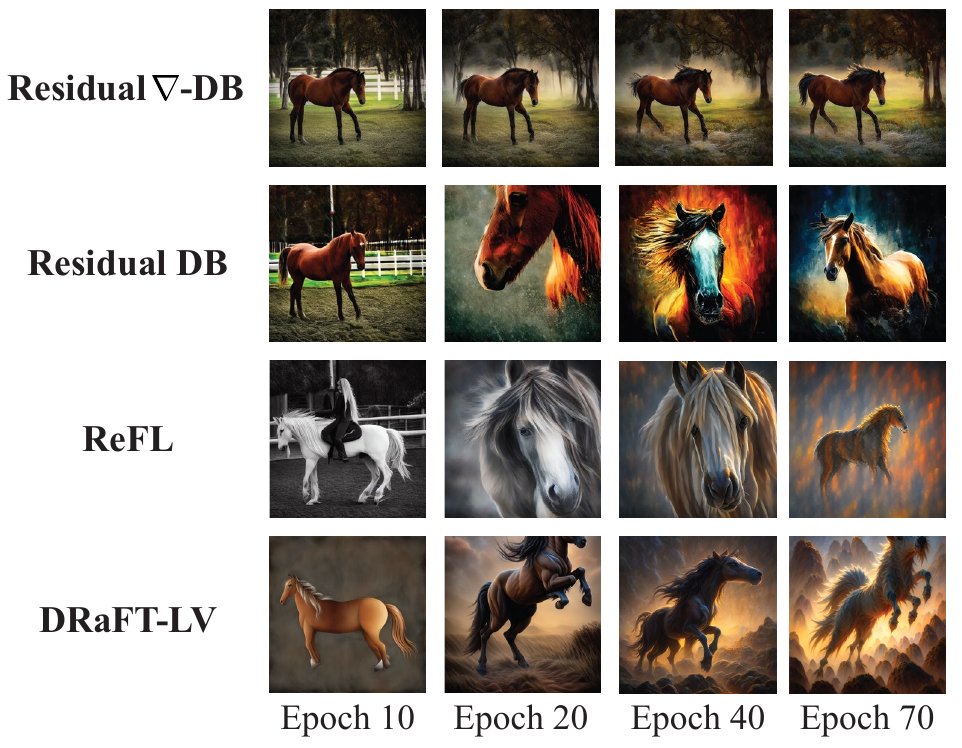}
    \vspace{-5.5mm}
    \caption{\footnotesize Our \methodname finetuning yields stable output compared to other baselines.}
    \vspace{-13mm}
    \label{fig:robustness}
\end{wrapfigure}

For \resdb and DAG-DB, we set the learning rate to $3 \times 10^{-4}$ with the output regularization strength $\lambda = 1$ The sampling and training procedures are similar to the \resgraddb experiments. For both ReFL and DRaFT, we use a learning rate of $10^{-4}$. For ReFL, we follow the official repo and similarly set the random stop time steps to between $35$ and $49$. For DRaFT, since the official code is not released, we follow the settings in AlignProp \cite{prabhudesai2023aligning}, a similar concurrent paper. We set the loss for both ReFL and DRaFT to $-\mathbb{E}_{x_T \sim P_F} \text{ReLU}(R(\text{decode}(x_T)))$ where the ReLU function is introduced for training stability in the case of the ImageReward reward function.

\vspace{-1mm}
\subsection{Experimental Results}
\vspace{-1mm}

\begin{figure}[t]
    \centering
    \includegraphics[width=.99\linewidth]{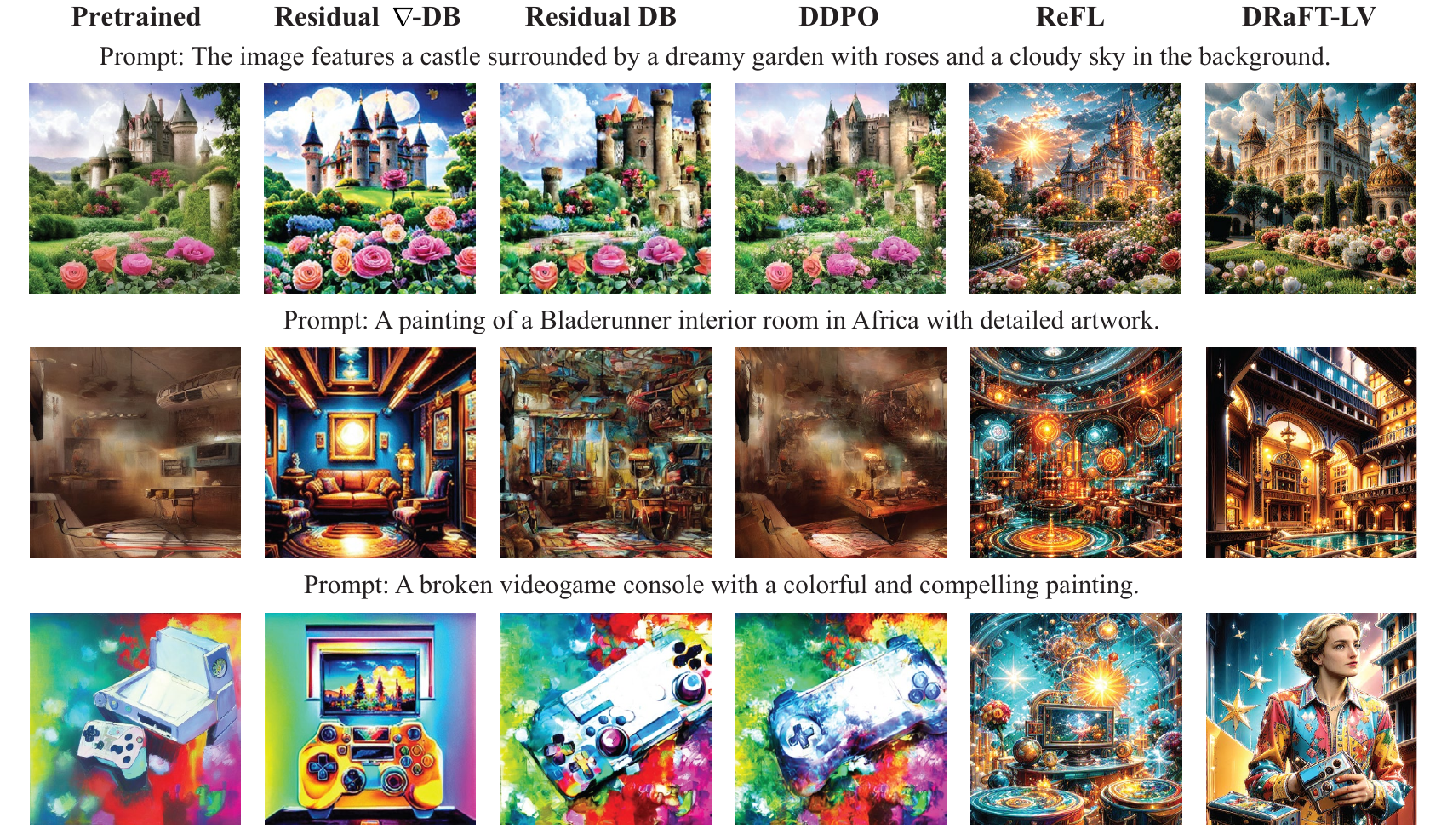}
    \vspace{-3mm}
    \caption{\footnotesize Qualitative results on HPSv2.}
    \label{fig:qualitative_hpsv2}
    \vspace{-1mm}
\end{figure}

\begin{figure}[t]
    \vspace{-1mm}
    \centering
    \includegraphics[width=.99\linewidth]{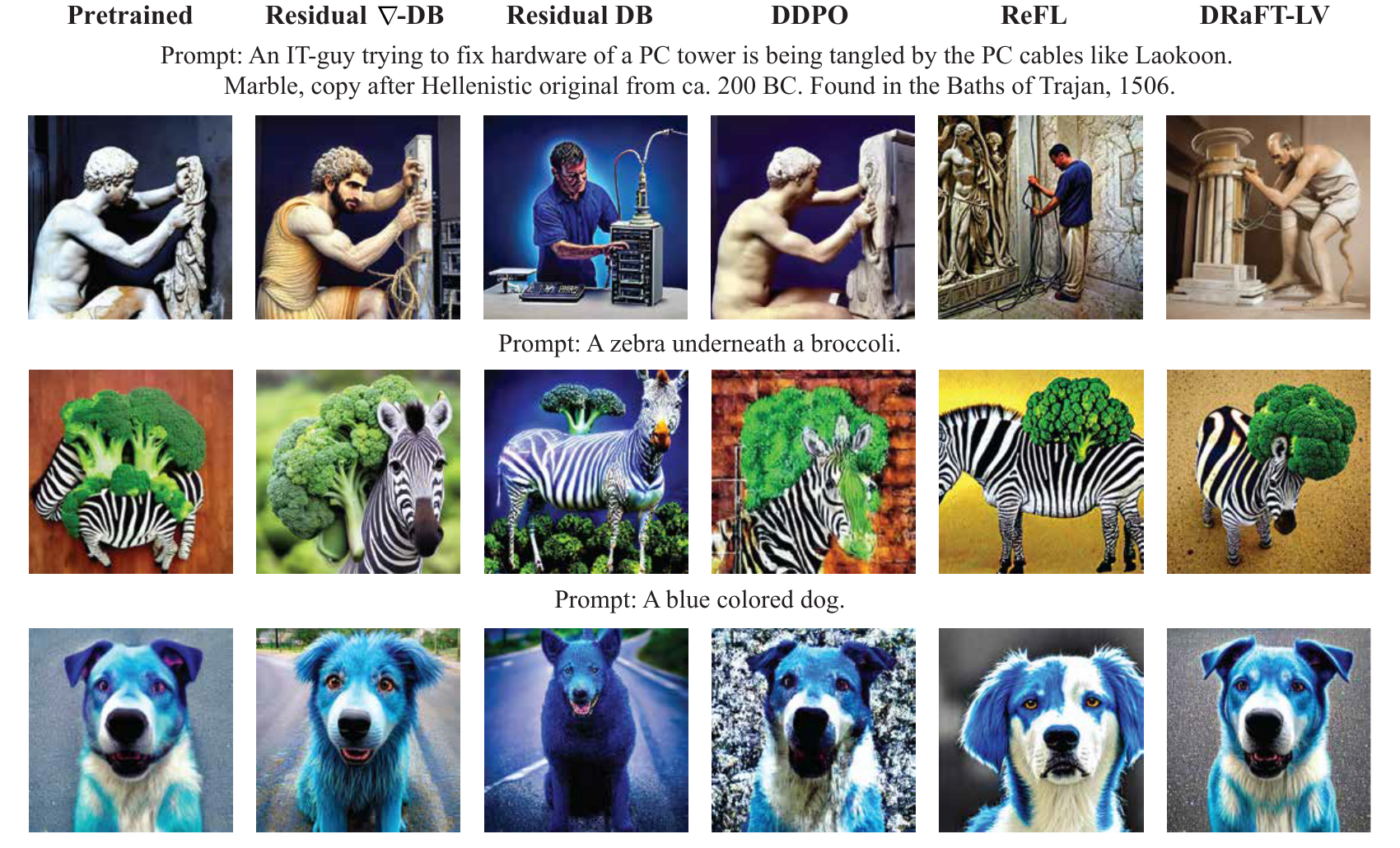}
    \vspace{-3mm}
    \caption{\footnotesize Qualitative results on ImageReward.}
    \label{fig:qualitative_imagereward}
    \vspace{-5mm}
\end{figure}

\begin{figure}[h]
    \centering
    \vspace{-17pt}
    \includegraphics[width=0.8\linewidth]{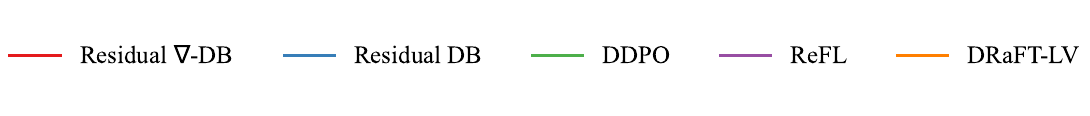}%
    \vspace{-1.5em}

    \includegraphics[width=0.33\linewidth]{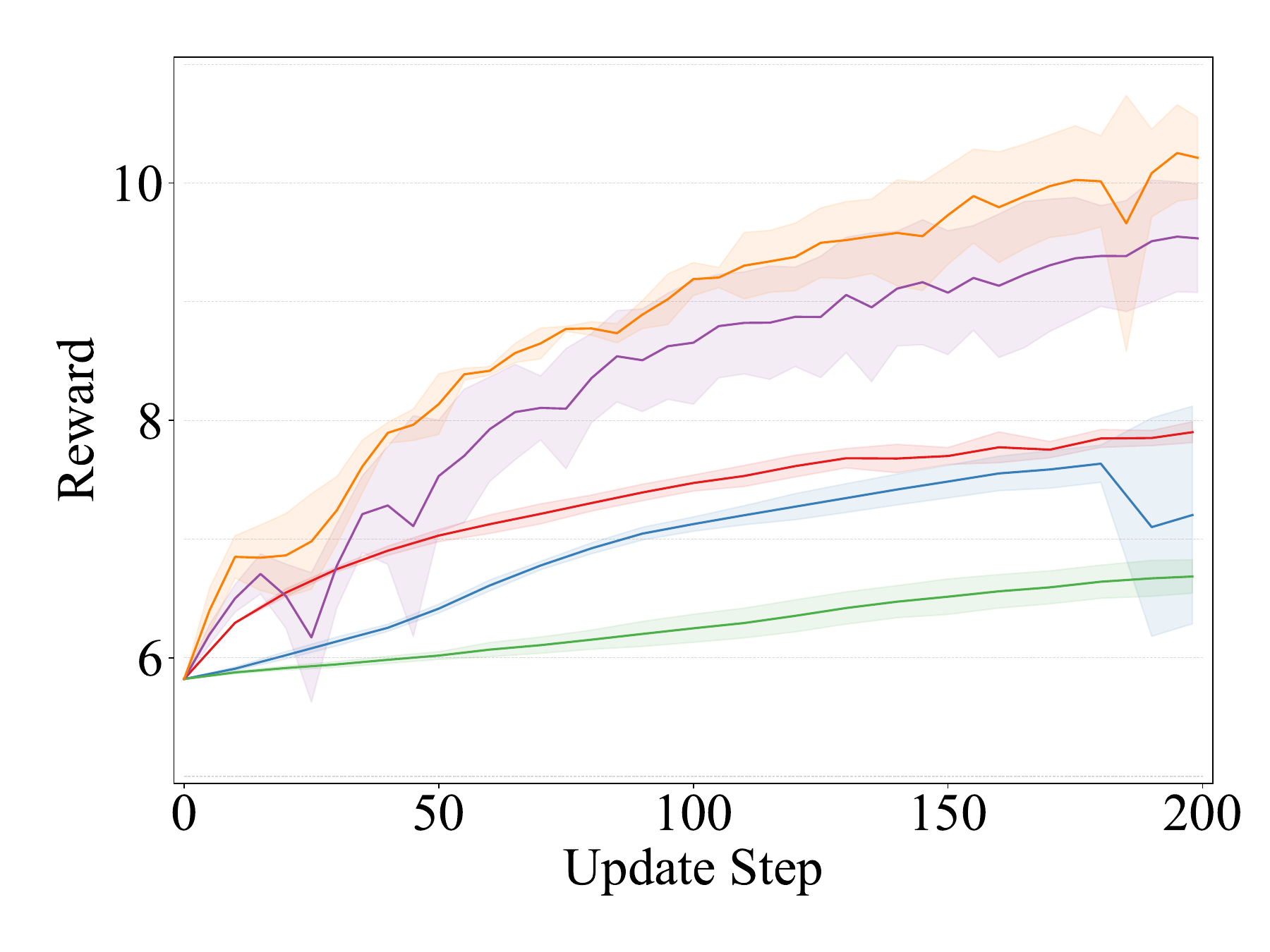}
    \hspace{-0.8em}
    \includegraphics[width=0.33\linewidth]{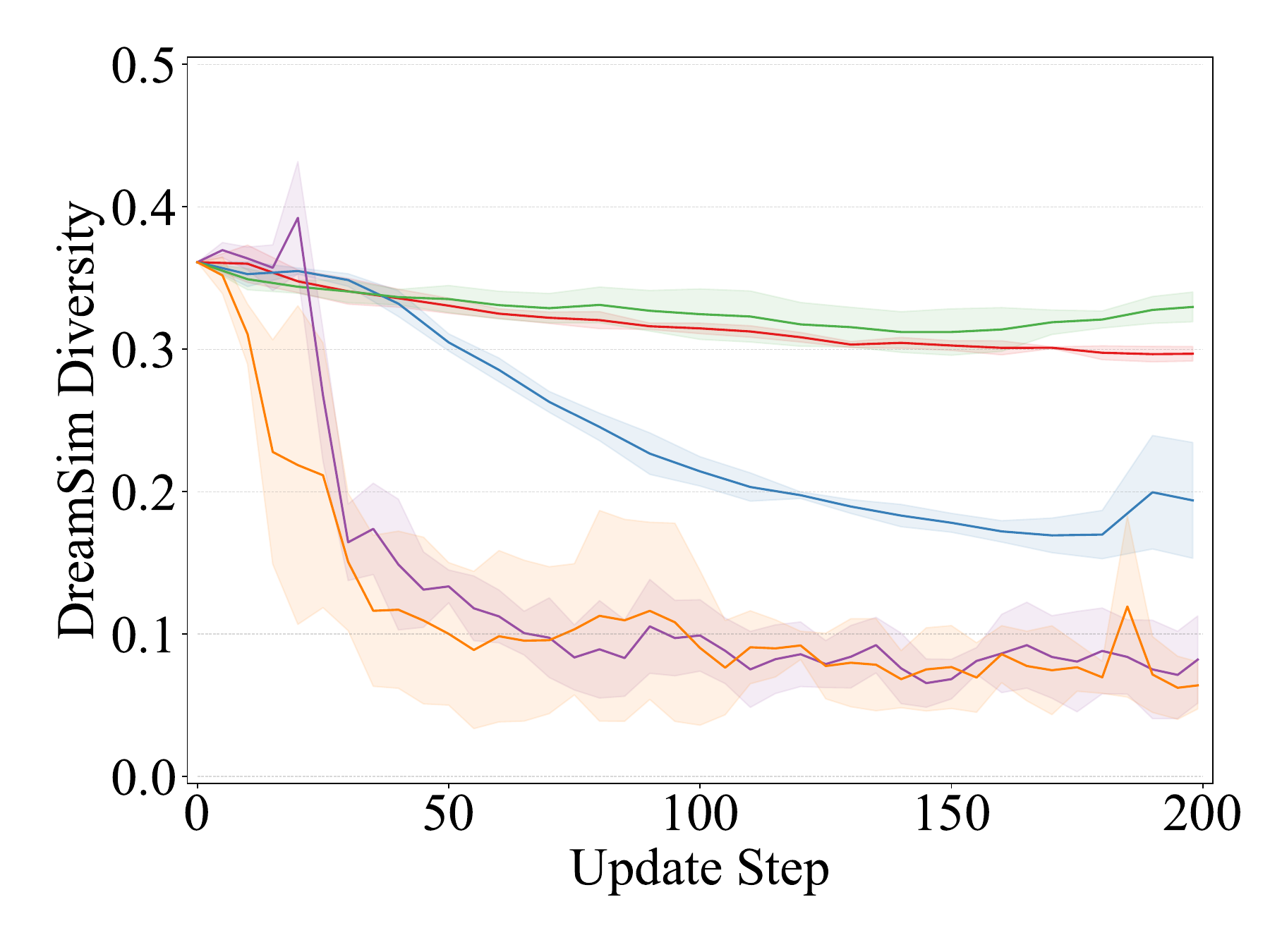}
    \hspace{-0.8em}
    \includegraphics[width=0.33\linewidth]{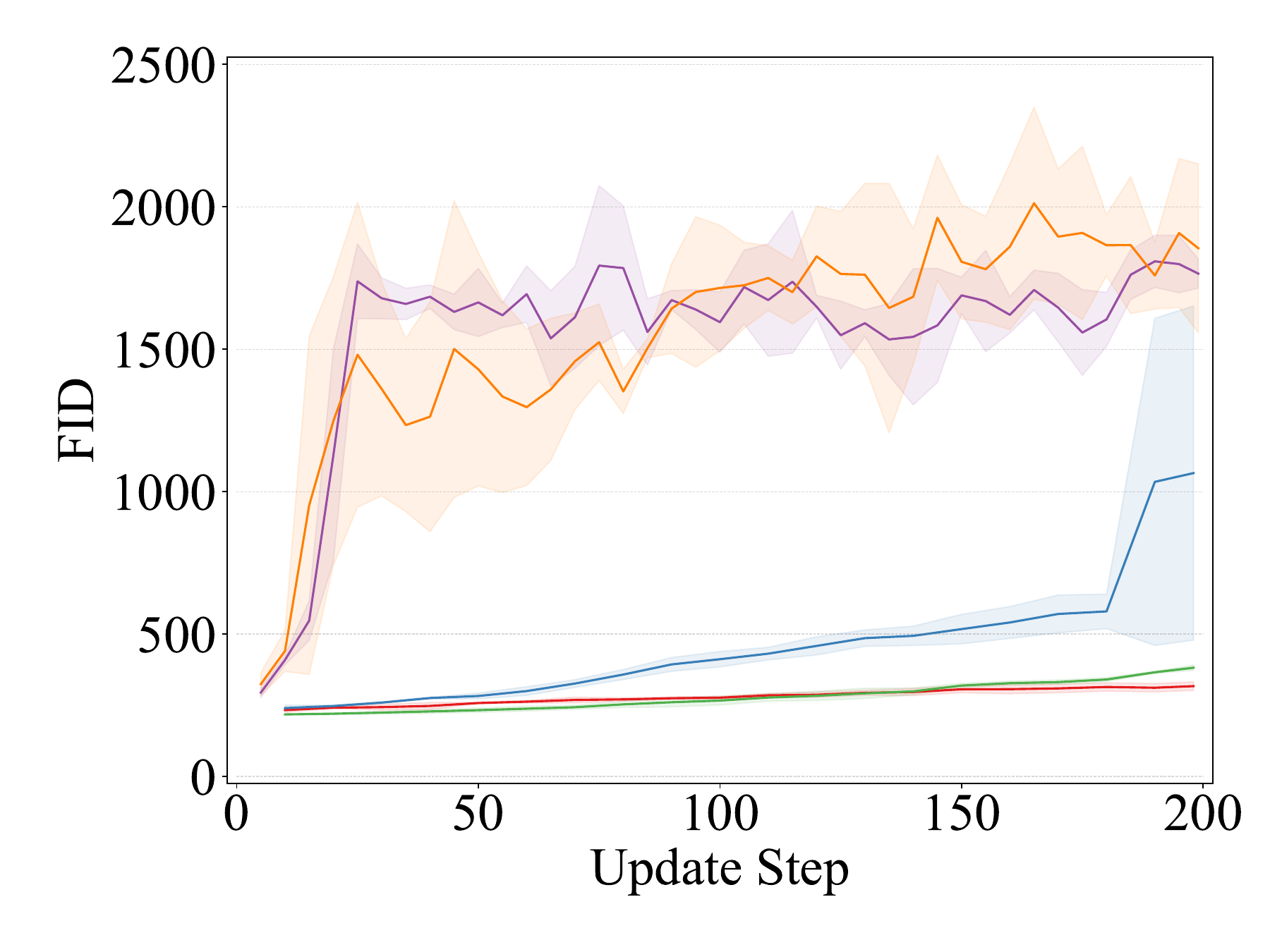}
    \vspace{-4mm}
    \caption{\footnotesize 
        Convergence curves of different metrics for different methods throughout the finetuning process on Aesthetic Score. Finetuning with our proposed \resgraddb converges faster than the non-gradient-informed methods and with better diversity- and prior-preserving capability.
    }
    \label{fig:general_results}
    \vspace{-5.5mm}
\end{figure}

\begin{figure}[t]
    \vspace{-6mm}
    \centering
    \includegraphics[width=0.48\linewidth]{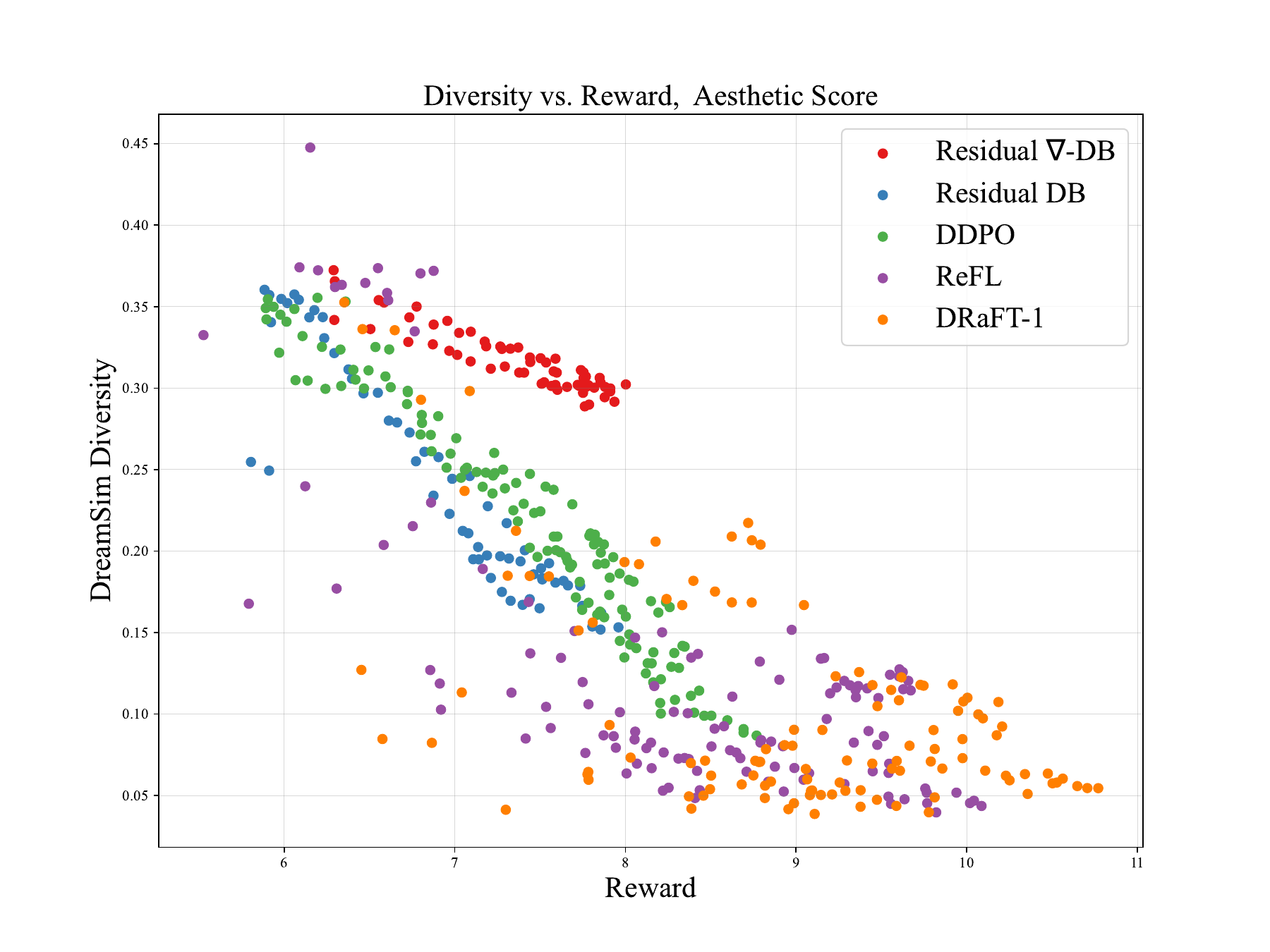}
    \hspace{-5mm}
    \includegraphics[width=0.48\linewidth]{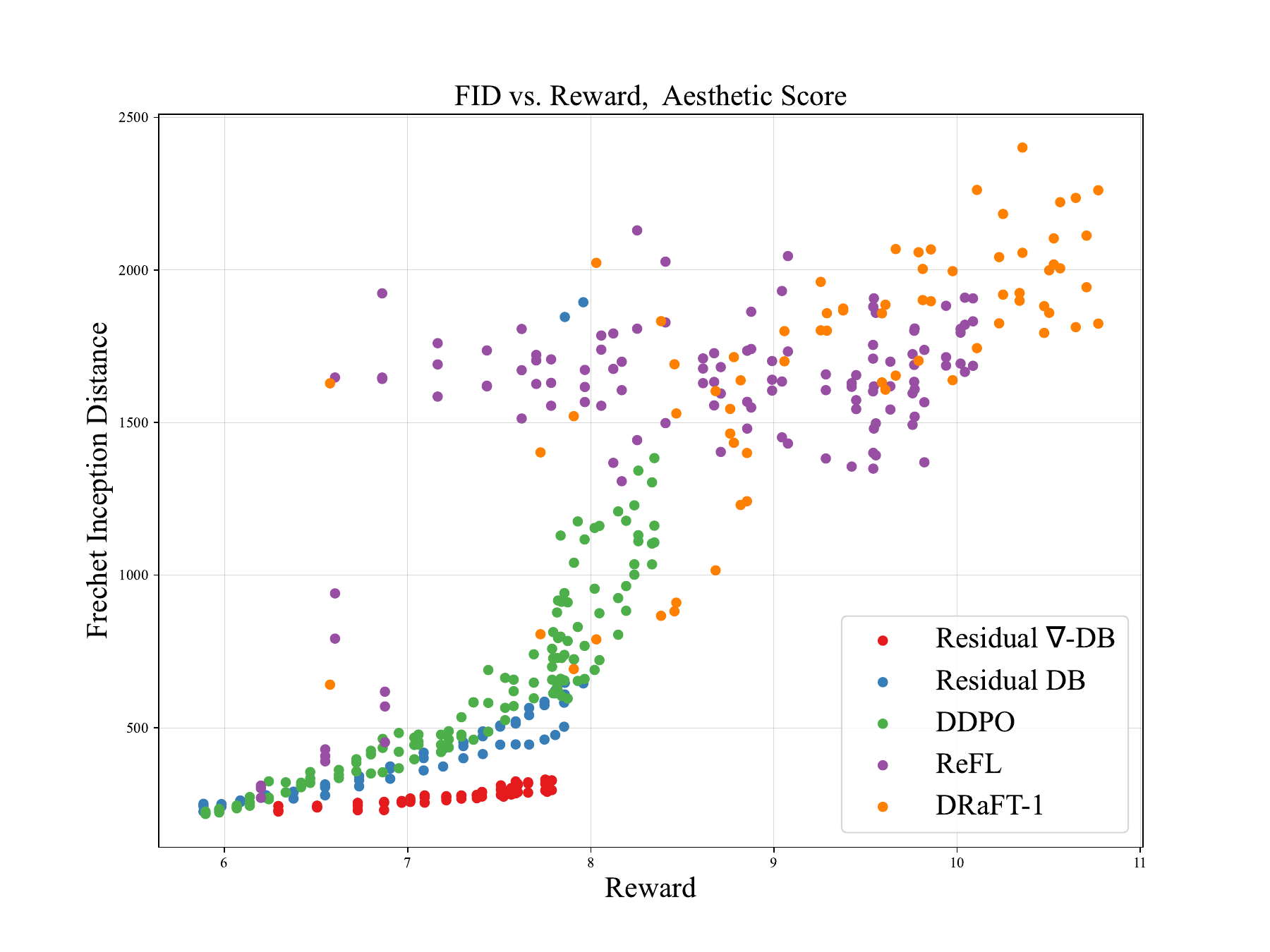}
    \vspace{-5mm}
    \caption{
        \footnotesize Trade-offs between reward, diversity preservation and prior preservation for different reward finetuning methods. Dots represent the evaluation results of models checkpoint saved after every 5 iterations of finetuning, where ones with larger reward, larger diversity scores and smaller FID scores are considered better.
    }
    \label{fig:general_tradeoff}
    \vspace{-2mm}
\end{figure}

\textbf{General experiments.} In Figure~\ref{fig:general_results} and Table~\ref{table:general_results}, we show the evolution of reward, DreamSim diversity and FID scores of all methods with the mean curves and the corresponding standard deviations (on $3$ random seeds). Our proposed \resgraddb is able to achieve comparable convergence speed, measured in update steps, to that of the gradient-free baselines while those diversity-aware baselines fail to do so. In Figure~\ref{fig:general_tradeoff}, we plot the diversity-reward tuples and FID-reward tuples for models evaluated at different checkpoints (every 5 update steps) and show that our method achieves Pareto improvements on diversity preservation, prior preservation and reward. The gradient-informed baselines, ReFL and the DRaFT variants, generally behave worse than \resgraddb due to their mode-seeking nature. 
Qualitatively, we show that the model finetuned with \resgraddb on Aesthetic Score generate more aesthetic and more diverse samples in both style and subject identity (Fig.~\ref{fig:main_qualitative_comparison} and Fig.~\ref{fig:additional_uncurated}), while the other baselines exhibit mode collapse or even catastrophic forgetting of pretrained image prior. We also demonstrate some images generated by the diffusion model finetuned with \resgraddb on HPSv2 in Figure~\ref{fig:qualitative_hpsv2} and on ImageReward in Figure~\ref{fig:qualitative_imagereward}. Furthermore, we qualitatively show that \resgraddb is robust while with the gradient-informed baseline methods are prone to training collapse (Fig.~\ref{fig:robustness}). Due to limited space, here we only show the most important ablation studies and leave the rest to Appendix~\ref{sec:more_ablation}. For the same reason, we leave plots for experiments on HPSv2 and ImageReward to Appendix~\ref{sec:appendix_ablation_figs} and qualitative comparisons on diversity to Appendix~\ref{sec:diversity_qualitative}.

\vspace{-0.3mm}

\begin{figure}[h]
    \centering
    \vspace{-4mm}
    \adjustbox{valign=t, max width=0.98\linewidth}{%
        \includegraphics[width=0.5\linewidth]{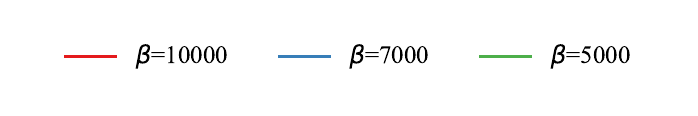}%
    }
    \vspace{-1.5em}

    \includegraphics[width=0.33\linewidth]{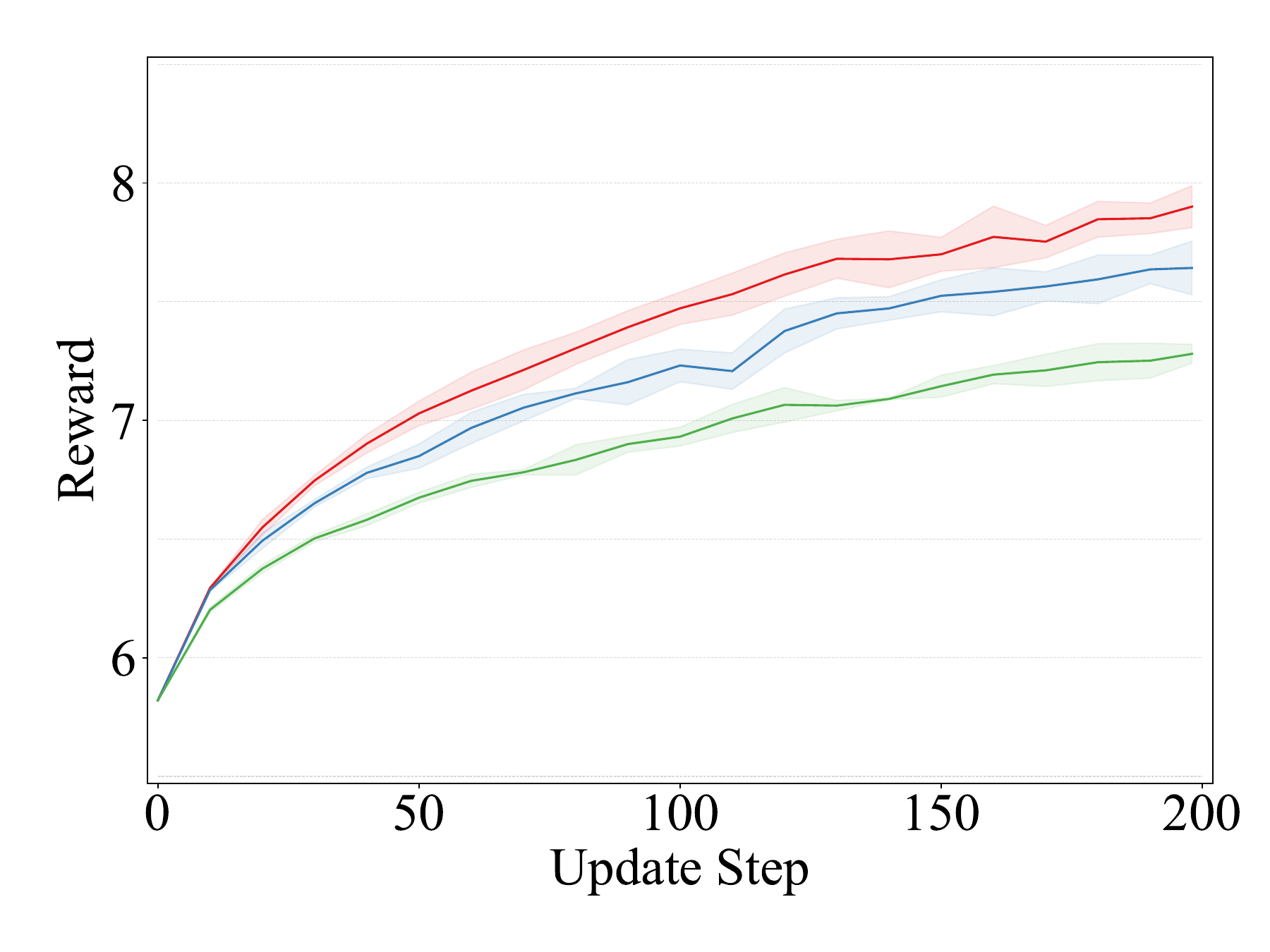}
    \hspace{-0.8em}
    \includegraphics[width=0.33\linewidth]{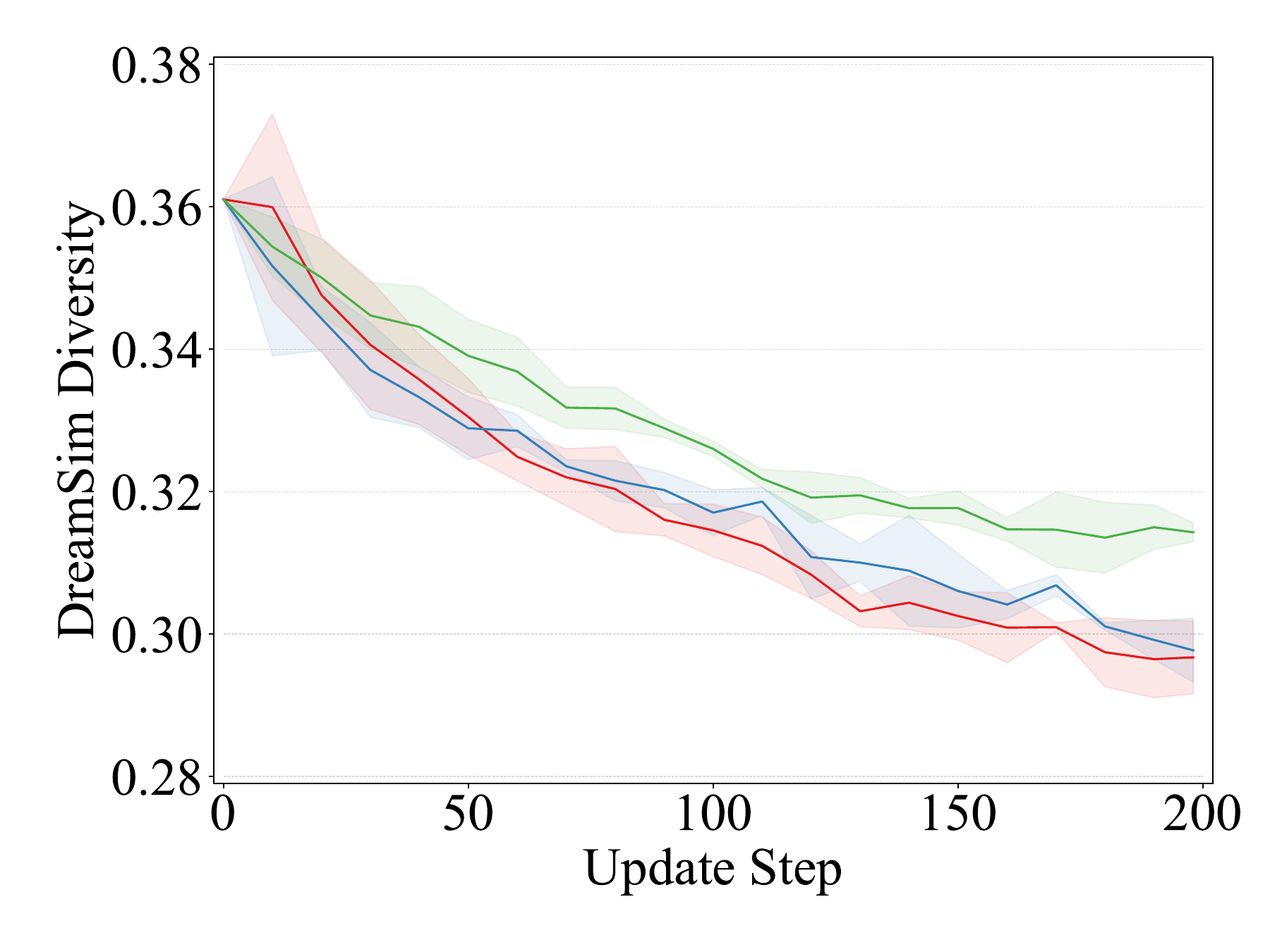}
    \hspace{-0.8em}
    \includegraphics[width=0.33\linewidth]{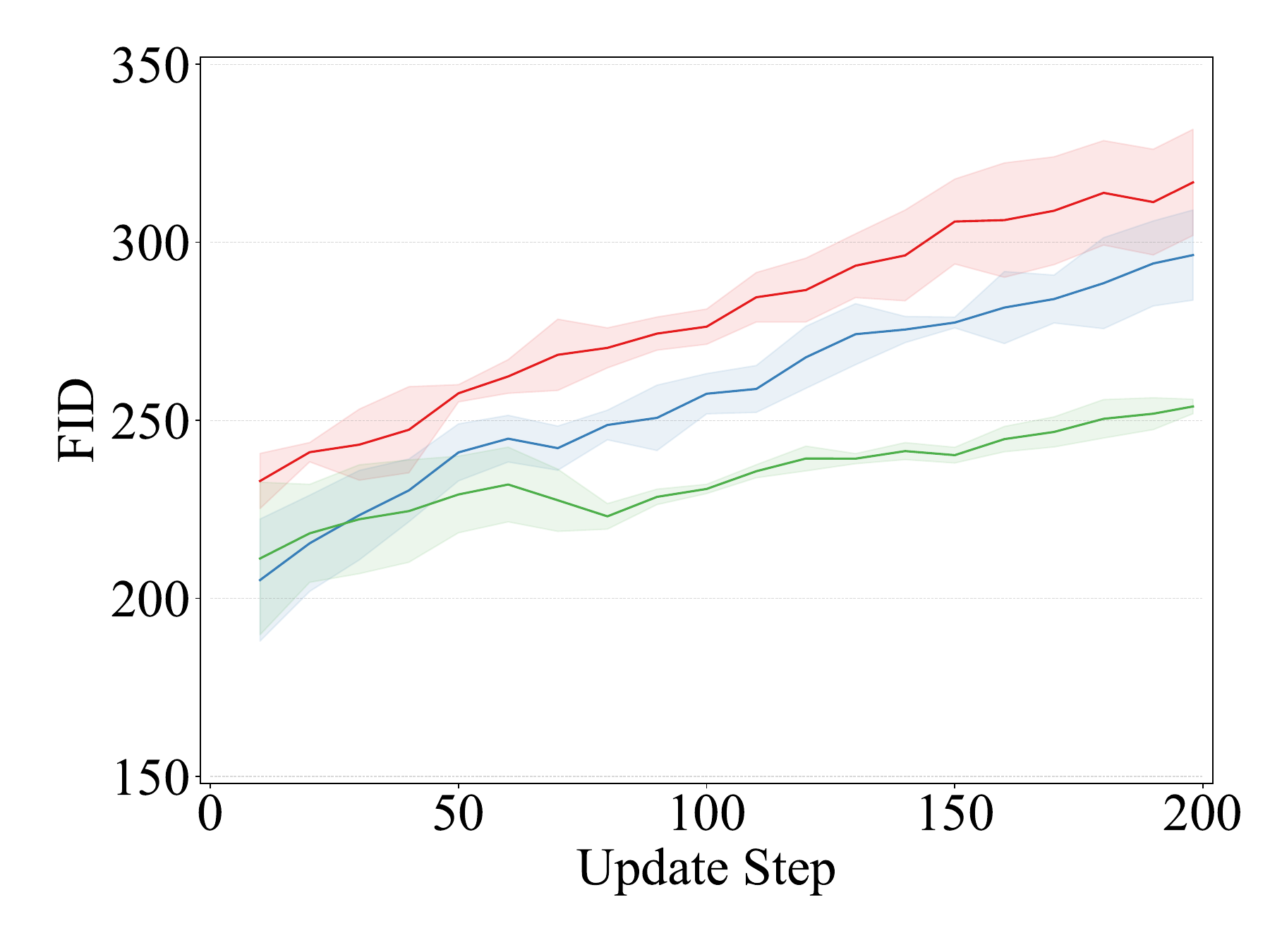}
    \vspace{-4mm}
    \caption{\footnotesize Higher temperature $\beta$ leads to faster convergence but with less diversity and less prior preservation.}
    \label{fig:ablation_temp}
    \vspace{-2.8mm}
\end{figure}

\textbf{Effect of reward temperature.} We perform ablation study on Aesthetic Score with $\beta \in \{5000, 7000, 10000\}$ in \resgraddb. Not surprisingly, a higher reward temperature leads to faster convergence at the cost of worse diversity- and worse prior-preservation, as observed in Figure~\ref{fig:ablation_temp}.

\vspace{-0.3mm}

\textbf{Effect of sub-sampling.} 
Typically, sub-sampling results in worse gradient estimates. We empirically study how sub-sampling may effect the performance and show the results in Figure~\ref{fig:ablation_subsampling} and \ref{fig:pareto_subsampling} in the appendix. We empirically do not observe huge performance drop due to the subsampling strategy, potentially because the rich gradient signals in both the reward and the flow are sufficient.

\vspace{-0.3mm}

\textbf{Effect of attenuating scaling on predicted reward.} In Fig.~\ref{fig:aesthetic_attenuation} and \ref{fig:pareto_attenuation}, we show that the model trained without attenuation converges faster, but at the cost of worse diversity and prior-preservation.

\vspace{-0.3mm}

\textbf{Reward finetuning with other sampling algorithms.} To show that our method generalizes to different sampling diffusion algorithms, we construct another MDP based on SDE-DPM-Solver++ \cite{lu2022dpm}, with 20 inference steps. In Fig.~\ref{fig:aesthetic_dpm} and \ref{fig:pareto_dpm}, we obverse that our \resgraddb can still achieve a good balance between reward convergence speed, diversity preservation and prior preservation.

\begin{table}[t]
\scriptsize
    \setlength{\tabcolsep}{3.1pt}
    \setlength\extrarowheight{3pt}
    \renewcommand{\arraystretch}{1.3}
    \centering
    \vspace{-2mm}
    \begin{tabular}{c|ccccccccc}
         \multirow{3}{*}{\hspace{1mm} Method \hspace{1mm}} & \multicolumn{3}{c}{Aesthetic Score} &  \multicolumn{3}{c}{HPSv2} & \multicolumn{3}{c}{ImageReward}
         \\
         & 
         \makecell{Reward \\ ($\uparrow$)} & \makecell{Diversity \\ DreamSim \\($\uparrow$, $10^{-2}$)} & \makecell{FID \\($\downarrow$)} & \makecell{Reward \\ ($\uparrow$, $10^{-1}$)} & \makecell{Diversity \\ DreamSim \\($\uparrow$, $10^{-2}$)} & \makecell{FID \\($\downarrow$)} & \makecell{Reward \\ ($\uparrow$, $10^{-1}$)} & \makecell{Diversity \\ DreamSim \\($\uparrow$, $10^{-2}$)} & \makecell{FID \\($\downarrow$)} \\\shline
         Base Model & 5.83 $\scriptscriptstyle\pm$ {\tiny 0.01} &35.91 $\scriptscriptstyle\pm$ {\tiny 0.00} &216 $\scriptscriptstyle\pm$ {\tiny 1} &2.38 $\scriptscriptstyle\pm$ {\tiny 0.13} &37.75 $\scriptscriptstyle\pm$ {\tiny 0.21} &563 $\scriptscriptstyle\pm$ {\tiny 5} &-0.38 $\scriptscriptstyle\pm$ {\tiny 0.12} &41.09 $\scriptscriptstyle\pm$ {\tiny 0.03} &468 $\scriptscriptstyle\pm$ {\tiny 1}
         \\\hline
         DDPO &6.68 $\scriptscriptstyle\pm$ {\tiny 0.14} &32.96 $\scriptscriptstyle\pm$ {\tiny 1.04} & \textbf{312} $\scriptscriptstyle\pm$ {\tiny 9} &2.52 $\scriptscriptstyle\pm$ {\tiny 0.04} &3.49 $\scriptscriptstyle\pm$ {\tiny 0.03} & \textbf{681} $\scriptscriptstyle\pm$ {\tiny 16} &0.27 $\scriptscriptstyle\pm$ {\tiny 0.38} &38.51 $\scriptscriptstyle\pm$ {\tiny 1.49} &714 $\scriptscriptstyle\pm$ {\tiny 25}
         \\
         ReFL &9.53 $\scriptscriptstyle\pm$ {\tiny 0.46} &8.20 $\scriptscriptstyle\pm$ {\tiny 3.06} &1765 $\scriptscriptstyle\pm$ {\tiny 51} &3.67 $\scriptscriptstyle\pm$ {\tiny 0.06} &19.84 $\scriptscriptstyle\pm$ {\tiny 1.70} &1191 $\scriptscriptstyle\pm$ {\tiny 46} &1.36 $\scriptscriptstyle\pm$ {\tiny 0.30} &36.50 $\scriptscriptstyle\pm$ {\tiny 0.52} &597 $\scriptscriptstyle\pm$ {\tiny 10}
         \\
         DRaFT-1 &10.16 $\scriptscriptstyle\pm$ {\tiny 0.13} &4.24 $\scriptscriptstyle\pm$ {\tiny 0.45} &1665 $\scriptscriptstyle\pm$ {\tiny 182} &3.70 $\scriptscriptstyle\pm$ {\tiny 0.06} &18.96 $\scriptscriptstyle\pm$ {\tiny 1.35} &1222 $\scriptscriptstyle\pm$ {\tiny 84} &1.59 $\scriptscriptstyle\pm$ {\tiny 0.25} &37.27 $\scriptscriptstyle\pm$ {\tiny 0.49} &531 $\scriptscriptstyle\pm$ {\tiny 13}
         \\
         DRaFT-LV & \textbf{10.21} $\scriptscriptstyle\pm$ {\tiny 0.34} &6.39 $\scriptscriptstyle\pm$ {\tiny 1.66} &1854 $\scriptscriptstyle\pm$ {\tiny 296} & \textbf{3.75} $\scriptscriptstyle\pm$ {\tiny 0.08} &21.13 $\scriptscriptstyle\pm$ {\tiny 1.19} &1164 $\scriptscriptstyle\pm$ {\tiny 43} &1.44 $\scriptscriptstyle\pm$ {\tiny 0.25} &37.56 $\scriptscriptstyle\pm$ {\tiny 0.09} &529 $\scriptscriptstyle\pm$ {\tiny 18}
         \\
         DAG-DB &7.73 $\scriptscriptstyle\pm$ {\tiny 0.07} &15.88 $\scriptscriptstyle\pm$ {\tiny 0.70} &595 $\scriptscriptstyle\pm$ {\tiny 87} &2.52 $\scriptscriptstyle\pm$ {\tiny 0.06} & \textbf{32.50} $\scriptscriptstyle\pm$ {\tiny 0.59} &866 $\scriptscriptstyle\pm$ {\tiny 41} &4.70 $\scriptscriptstyle\pm$ {\tiny 0.75} &26.78 $\scriptscriptstyle\pm$ {\tiny 0.64} &809 $\scriptscriptstyle\pm$ {\tiny 34}
         \\
         \makecell{\resdb} &7.20 $\scriptscriptstyle\pm$ {\tiny 0.92} &19.38 $\scriptscriptstyle\pm$ {\tiny 4.07} &1065 $\scriptscriptstyle\pm$ {\tiny 587} &2.55 $\scriptscriptstyle\pm$ {\tiny 0.06} &32.49 $\scriptscriptstyle\pm$ {\tiny 0.47} &840 $\scriptscriptstyle\pm$ {\tiny 107} & \textbf{6.47} $\scriptscriptstyle\pm$ {\tiny 0.54} &25.52 $\scriptscriptstyle\pm$ {\tiny 0.35} &772 $\scriptscriptstyle\pm$ {\tiny 19}
         \\\rowcolor{Gray}
         \makecell{\methodname \\ {\scriptsize($\scriptscriptstyle w_B = 0$)}} &7.86 $\scriptscriptstyle\pm$ {\tiny 0.06} &29.21 $\scriptscriptstyle\pm$ {\tiny 0.18} &318 $\scriptscriptstyle\pm$ {\tiny 13} &3.53 $\scriptscriptstyle\pm$ {\tiny 0.03} &24.40 $\scriptscriptstyle\pm$ {\tiny 0.56} &973 $\scriptscriptstyle\pm$ {\tiny 5} & 
         5.33 $\scriptscriptstyle\pm$ {\tiny 0.62} & 35.85 $\scriptscriptstyle\pm$ {\tiny 0.66} & 638 $\scriptscriptstyle\pm$ {\tiny 15}
         \\\rowcolor{Gray}
         \makecell{\methodname \\{\scriptsize($\scriptscriptstyle w_B = 1$)}} &7.90 $\scriptscriptstyle\pm$ {\tiny 0.09} & \textbf{29.67} $\scriptscriptstyle\pm$ {\tiny 0.51}  &317 $\scriptscriptstyle\pm$ {\tiny 15} &3.53 $\scriptscriptstyle\pm$ {\tiny 0.04} &24.39 $\scriptscriptstyle\pm$ {\tiny 0.87} &1000 $\scriptscriptstyle\pm$ {\tiny 39} &2.23 $\scriptscriptstyle\pm$ {\tiny 0.34} & \textbf{39.85} $\scriptscriptstyle\pm$ {\tiny 0.67} & \textbf{501} $\scriptscriptstyle\pm$ {\tiny 7}\\
    \end{tabular}
    \vspace{-1.75mm}
    \caption{\footnotesize Comparison between the models finetuned with our proposed method \methodname (with the objective \resgraddb) and the baselines. All models are finetuned with 200 update steps. For each method, the model with the best mean reward is used for evaluation. Note that while baselines like DDPO can achieve better scores on some metric, it often comes with the price of much worse performance on some other.}
    \label{table:general_results}
    \vspace{-3.25mm}
\end{table}

\section{Related Work}
\label{sec:related_work}
\vspace{-1.75mm}

\noindent \textbf{Reward finetuning of diffusion models.} The demand for reward finetuning is commonly seen in alignment, where one obtains utilize a human reference reward function to align the behavior of generative models~\citep{ibarz2018reward, bai2022training,wu2023humanv2} for better instruction following capability and better AI safety. 
With a reward function, typically obtained by learning from human preference datasets~\cite{ziegler2019fine,stiennon2022learningsummarizehumanfeedback}, one may use reinforcement learning (RL) algorithms, for instance PPO~\cite{schulman2017proximal}, to adapt not only autoregressive language models~\cite{ouyang2022training} which naturally admits a Markov decision process (MDP), but also diffusion models~\cite{black2024training,Fan2023DPOKRL}. Specifically, one can construct MDPs from some diffusion sampling algorithm~\cite{song2021denoising, lu2022dpm} by considering each noisy image at some inference step as a state and each denoising step as an action. Besides RL algorithms, there exist some other approaches, including stochastic optimal control~\cite{domingoenrich2024adjoint, uehara2024fine}, GFlowNets~\cite{zhang2025improving} and some other ones akin to RL methods~\cite{lee2023aligning,dong2023raft}. While most of the aforementioned approaches train with only black-box rewards, once we have access to a differentiable reward function we may accelerate the finetuning process with reward gradient signals. For instance, methods exist to construct a computational graph from sampled generation trajectories to directly optimize for rewards~\cite{clark2024directly,prabhudesai2023aligning,xu2024imagereward}, yet with these methods models are not trained to correctly sample according to the reward function. While one may also generate samples from the reward function without finetuning using plug-in guidance methods for diffusion models~\cite{dhariwal2021diffusion,Song2023LossGuidedDM,kong2024diffusion,guo2024gradient} as an alternative, but the generated distributions are often very biased. Besides, reward finetuning for diffusion models is typically memory consuming as many methods require a large computational graph rolled out from long generation trajectories, for which it is typical to employ parameter-efficient finetuning techniques \cite{hu2022lora, qiu2023controlling, liu2024boft}.
\vspace{-0.5mm}

\noindent \textbf{GFlowNets.}
Generative flow network~(GFlowNet)~\citep{bengio2023gflownet} is a high-level algorithmic framework that introduces sequential decision-making into generative modeling~\citep{zhang2022unifying}, bridging methodology between reinforcement learning~\citep{zhang2024distributional,pan2023gafn,pan2023stochastic,pan2023better,lau2024qgfn} and energy-based modeling~\citep{zhang2022generative}. 
GFlowNets are versatile since many models, such as diffusion models, can be easily treated as specifications of GFlowNets~\citep{zhang2022unifying,lahlou2023cgfn,zhang2024diffusion,sendera2024improved}.
GFlowNets perform amortized variational inference~\cite{malkin2022gfnhvi} and generate samples with probability proportional to a given density or reward. 
GFlowNets are used in various applications including but not limited to drug discovery~\citep{jain2022biological,jain2023gflownetsfa,shen2023towardsua}, structure learning~\citep{deleu2022bayesian}, phylogenetic inference~\citep{zhou2024phylogfn}, combinatorial optimization~\citep{Zhang2023RobustSW,zhang2023let}, and prompt adaptation~\citep{yun2025learning}.

\vspace{-2mm}

\section{Concluding Remarks}

\vspace{-1.5mm}

We propose \methodname, a fast, diversity and prior-preserving diffusion alignment method, by leveraging gradient information in the probabilistic framework of GFlowNets that aims to sample according to a given unnormalized density function or reward. Specifically, we develop \graddb, the gradient-informed version of the Detailed Balance objective, and the variant of \resgraddb with which one may finetune a diffusion model with reward in a prior-preserving way. Our empirical results show that \methodname achieves a better trade-off between convergence speed, diversity in generated samples and prior preservation. We hope that our method sheds lights on future studies on more efficient post-training alignment strategies of diffusion models as well as related applications.

\clearpage
\bibliography{iclr2025_conference}

\begin{thebibliography}{84}
\providecommand{\natexlab}[1]{#1}
\providecommand{\url}[1]{\texttt{#1}}
\expandafter\ifx\csname urlstyle\endcsname\relax
  \providecommand{\doi}[1]{doi: #1}\else
  \providecommand{\doi}{doi: \begingroup \urlstyle{rm}\Url}\fi

\bibitem[Bai et~al.(2022)Bai, Jones, Ndousse, Askell, Chen, DasSarma, Drain,
  Fort, Ganguli, Henighan, et~al.]{bai2022training}
Yuntao Bai, Andy Jones, Kamal Ndousse, Amanda Askell, Anna Chen, Nova DasSarma,
  Dawn Drain, Stanislav Fort, Deep Ganguli, Tom Henighan, et~al.
\newblock Training a helpful and harmless assistant with reinforcement learning
  from human feedback.
\newblock \emph{arXiv preprint arXiv:2204.05862}, 2022.

\bibitem[Bengio et~al.(2023)Bengio, Lahlou, Deleu, Hu, Tiwari, and
  Bengio]{bengio2023gflownet}
Yoshua Bengio, Salem Lahlou, Tristan Deleu, Edward~J Hu, Mo~Tiwari, and
  Emmanuel Bengio.
\newblock Gflownet foundations.
\newblock \emph{Journal of Machine Learning Research}, 2023.

\bibitem[Black et~al.(2024)Black, Janner, Du, Kostrikov, and
  Levine]{black2024training}
Kevin Black, Michael Janner, Yilun Du, Ilya Kostrikov, and Sergey Levine.
\newblock Training diffusion models with reinforcement learning.
\newblock In \emph{ICLR}, 2024.

\bibitem[Blattmann et~al.(2023)Blattmann, Dockhorn, Kulal, Mendelevitch,
  Kilian, Lorenz, Levi, English, Voleti, Letts, et~al.]{blattmann2023stable}
Andreas Blattmann, Tim Dockhorn, Sumith Kulal, Daniel Mendelevitch, Maciej
  Kilian, Dominik Lorenz, Yam Levi, Zion English, Vikram Voleti, Adam Letts,
  et~al.
\newblock Stable video diffusion: Scaling latent video diffusion models to
  large datasets.
\newblock \emph{arXiv preprint arXiv:2311.15127}, 2023.

\bibitem[Christiano et~al.(2017)Christiano, Leike, Brown, Martic, Legg, and
  Amodei]{christiano2017deep}
Paul Christiano, Jan Leike, Tom Brown, Miljan Martic, Shane Legg, and Dario
  Amodei.
\newblock Deep reinforcement learning from human preferences.
\newblock In \emph{NeurIPS}, 2017.

\bibitem[Clark et~al.(2024)Clark, Vicol, Swersky, and Fleet]{clark2024directly}
Kevin Clark, Paul Vicol, Kevin Swersky, and David~J. Fleet.
\newblock Directly fine-tuning diffusion models on differentiable rewards.
\newblock In \emph{ICLR}, 2024.

\bibitem[Deleu(2025)]{deleu2025generative}
Tristan Deleu.
\newblock Generative flow networks: Theory and applications to structure
  learning.
\newblock \emph{arXiv preprint arXiv:2501.05498}, 2025.

\bibitem[Deleu et~al.(2022)Deleu, G\'{o}is, Emezue, Rankawat, Lacoste-Julien,
  Bauer, and Bengio]{deleu2022bayesian}
Tristan Deleu, Ant\'{o}nio G\'{o}is, Chris Emezue, Mansi Rankawat, Simon
  Lacoste-Julien, Stefan Bauer, and Yoshua Bengio.
\newblock Bayesian structure learning with generative flow networks.
\newblock In \emph{UAI}, 2022.

\bibitem[Dhariwal \& Nichol(2021)Dhariwal and Nichol]{dhariwal2021diffusion}
Prafulla Dhariwal and Alexander Nichol.
\newblock Diffusion models beat gans on image synthesis.
\newblock In \emph{NeurIPS}, 2021.

\bibitem[Domingo-Enrich et~al.(2024)Domingo-Enrich, Drozdzal, Karrer, and
  Chen]{domingoenrich2024adjoint}
Carles Domingo-Enrich, Michal Drozdzal, Brian Karrer, and Ricky T.~Q. Chen.
\newblock Adjoint matching: Fine-tuning flow and diffusion generative models
  with memoryless stochastic optimal control.
\newblock In \emph{ICLR}, 2024.

\bibitem[Dong et~al.(2023)Dong, Xiong, Goyal, Zhang, Chow, Pan, Diao, Zhang,
  SHUM, and Zhang]{dong2023raft}
Hanze Dong, Wei Xiong, Deepanshu Goyal, Yihan Zhang, Winnie Chow, Rui Pan,
  Shizhe Diao, Jipeng Zhang, KaShun SHUM, and Tong Zhang.
\newblock {RAFT}: Reward ranked finetuning for generative foundation model
  alignment.
\newblock \emph{Transactions on Machine Learning Research}, 2023.

\bibitem[Fan et~al.(2023)Fan, Watkins, Du, Liu, Ryu, Boutilier, Abbeel,
  Ghavamzadeh, Lee, and Lee]{Fan2023DPOKRL}
Ying Fan, Olivia Watkins, Yuqing Du, Hao Liu, Moonkyung Ryu, Craig Boutilier,
  P.~Abbeel, Mohammad Ghavamzadeh, Kangwook Lee, and Kimin Lee.
\newblock Dpok: Reinforcement learning for fine-tuning text-to-image diffusion
  models.
\newblock In \emph{NeurIPS}, 2023.

\bibitem[Gao et~al.(2024)Gao, Liu, Chen, Geiger, and
  Sch{\"o}lkopf]{gao2024graphdreamer}
Gege Gao, Weiyang Liu, Anpei Chen, Andreas Geiger, and Bernhard Sch{\"o}lkopf.
\newblock Graphdreamer: Compositional 3d scene synthesis from scene graphs.
\newblock In \emph{CVPR}, 2024.

\bibitem[Guo et~al.(2024)Guo, Yuan, Yang, Chen, and Wang]{guo2024gradient}
Yingqing Guo, Hui Yuan, Yukang Yang, Minshuo Chen, and Mengdi Wang.
\newblock Gradient guidance for diffusion models: An optimization perspective.
\newblock In \emph{NeurIPS}, 2024.

\bibitem[Haarnoja et~al.(2017)Haarnoja, Tang, Abbeel, and
  Levine]{haarnoja2017reinforcement}
Tuomas Haarnoja, Haoran Tang, Pieter Abbeel, and Sergey Levine.
\newblock Reinforcement learning with deep energy-based policies.
\newblock In \emph{ICML}, 2017.

\bibitem[Haarnoja et~al.(2018)Haarnoja, Zhou, Abbeel, and
  Levine]{haarnoja2018soft}
Tuomas Haarnoja, Aurick Zhou, Pieter Abbeel, and Sergey Levine.
\newblock Soft actor-critic: Off-policy maximum entropy deep reinforcement
  learning with a stochastic actor.
\newblock In \emph{ICML}, 2018.

\bibitem[Ho et~al.(2020)Ho, Jain, and Abbeel]{ho2020denoising}
Jonathan Ho, Ajay Jain, and Pieter Abbeel.
\newblock Denoising diffusion probabilistic models.
\newblock In \emph{NeurIPS}, 2020.

\bibitem[Ho et~al.(2022{\natexlab{a}})Ho, Chan, Saharia, Whang, Gao, Gritsenko,
  Kingma, Poole, Norouzi, Fleet, et~al.]{ho2022imagen}
Jonathan Ho, William Chan, Chitwan Saharia, Jay Whang, Ruiqi Gao, Alexey
  Gritsenko, Diederik~P Kingma, Ben Poole, Mohammad Norouzi, David~J Fleet,
  et~al.
\newblock Imagen video: High definition video generation with diffusion models.
\newblock \emph{arXiv preprint arXiv:2210.02303}, 2022{\natexlab{a}}.

\bibitem[Ho et~al.(2022{\natexlab{b}})Ho, Salimans, Gritsenko, Chan, Norouzi,
  and Fleet]{ho2022video}
Jonathan Ho, Tim Salimans, Alexey Gritsenko, William Chan, Mohammad Norouzi,
  and David~J Fleet.
\newblock Video diffusion models.
\newblock In \emph{NeurIPS}, 2022{\natexlab{b}}.

\bibitem[Hoogeboom et~al.(2022)Hoogeboom, Satorras, Vignac, and
  Welling]{hoogeboom2022equivariant}
Emiel Hoogeboom, V{\i}ctor~Garcia Satorras, Cl{\'e}ment Vignac, and Max
  Welling.
\newblock Equivariant diffusion for molecule generation in 3d.
\newblock In \emph{ICML}, 2022.

\bibitem[Hu et~al.(2022)Hu, Shen, Wallis, Allen-Zhu, Li, Wang, Wang, and
  Chen]{hu2022lora}
Edward~J. Hu, Yelong Shen, Phillip Wallis, Zeyuan Allen-Zhu, Yuanzhi Li, Shean
  Wang, Lu~Wang, and Weizhu Chen.
\newblock Lora: Low-rank adaptation of large language models.
\newblock In \emph{ICLR}, 2022.

\bibitem[Ibarz et~al.(2018)Ibarz, Leike, Pohlen, Irving, Legg, and
  Amodei]{ibarz2018reward}
Borja Ibarz, Jan Leike, Tobias Pohlen, Geoffrey Irving, Shane Legg, and Dario
  Amodei.
\newblock Reward learning from human preferences and demonstrations in atari.
\newblock In \emph{NeurIPS}, 2018.

\bibitem[Jain et~al.(2022)Jain, Bengio, Hernandez-Garcia, Rector-Brooks,
  Dossou, Ekbote, Fu, Zhang, Kilgour, Zhang, et~al.]{jain2022biological}
Moksh Jain, Emmanuel Bengio, Alex Hernandez-Garcia, Jarrid Rector-Brooks,
  Bonaventure~FP Dossou, Chanakya~Ajit Ekbote, Jie Fu, Tianyu Zhang, Michael
  Kilgour, Dinghuai Zhang, et~al.
\newblock Biological sequence design with gflownets.
\newblock In \emph{ICML}, 2022.

\bibitem[Jain et~al.(2023)Jain, Deleu, Hartford, Liu, Hern{\'a}ndez-Garc{\'i}a,
  and Bengio]{jain2023gflownetsfa}
Moksh Jain, Tristan Deleu, Jason~S. Hartford, Cheng-Hao Liu, Alex
  Hern{\'a}ndez-Garc{\'i}a, and Yoshua Bengio.
\newblock Gflownets for ai-driven scientific discovery.
\newblock \emph{ArXiv}, 2023.

\bibitem[Johnson(2004)]{johnson2004information}
Oliver Johnson.
\newblock \emph{Information theory and the central limit theorem}.
\newblock World Scientific, 2004.

\bibitem[Kong et~al.(2024)Kong, Du, Mu, Neklyudov, De~Bortol, Wang, Wu, Ferber,
  Ma, Gomes, et~al.]{kong2024diffusion}
Lingkai Kong, Yuanqi Du, Wenhao Mu, Kirill Neklyudov, Valentin De~Bortol,
  Haorui Wang, Dongxia Wu, Aaron Ferber, Yi-An Ma, Carla~P Gomes, et~al.
\newblock Diffusion models as constrained samplers for optimization with
  unknown constraints.
\newblock In \emph{AISTATS}, 2024.

\bibitem[Lahlou et~al.(2023)Lahlou, Deleu, Lemos, Zhang, Volokhova,
  Hern\'{a}ndez-Garc\'{i}a, Ezzine, Bengio, and Malkin]{lahlou2023cgfn}
Salem Lahlou, Tristan Deleu, Pablo Lemos, Dinghuai Zhang, Alexandra Volokhova,
  Alex Hern\'{a}ndez-Garc\'{i}a, L\'{e}na~N\'{e}hale Ezzine, Yoshua Bengio, and
  Nikolay Malkin.
\newblock A theory of continuous generative flow networks.
\newblock In \emph{ICML}, 2023.

\bibitem[LAION(2024)]{laion_aesthetic_2024}
LAION.
\newblock Laion aesthetic score predictor.
\newblock \url{https://laion.ai/blog/laion-aesthetics/}, 2024.
\newblock Accessed: 2024-09-27.

\bibitem[Lau et~al.(2024)Lau, Lu, Pan, Precup, and Bengio]{lau2024qgfn}
Elaine Lau, Stephen~Zhewen Lu, Ling Pan, Doina Precup, and Emmanuel Bengio.
\newblock Qgfn: Controllable greediness with action values.
\newblock In \emph{NeurIPS}, 2024.

\bibitem[Lee et~al.(2023)Lee, Liu, Ryu, Watkins, Du, Boutilier, Abbeel,
  Ghavamzadeh, and Gu]{lee2023aligning}
Kimin Lee, Hao Liu, Moonkyung Ryu, Olivia Watkins, Yuqing Du, Craig Boutilier,
  Pieter Abbeel, Mohammad Ghavamzadeh, and Shixiang~Shane Gu.
\newblock Aligning text-to-image models using human feedback.
\newblock \emph{arXiv preprint arXiv:2302.12192}, 2023.

\bibitem[Lipman et~al.(2023)Lipman, Chen, Ben-Hamu, Nickel, and
  Le]{lipman2023flow}
Yaron Lipman, Ricky~TQ Chen, Heli Ben-Hamu, Maximilian Nickel, and Matthew Le.
\newblock Flow matching for generative modeling.
\newblock In \emph{ICLR}, 2023.

\bibitem[Liu et~al.(2024{\natexlab{a}})Liu, Yan, Du, Liu, Li, Guo, Borgs,
  Chayes, and Anandkumar]{liu2024manifold}
Shengchao Liu, Divin Yan, Weitao Du, Weiyang Liu, Zhuoxinran Li, Hongyu Guo,
  Christian Borgs, Jennifer Chayes, and Anima Anandkumar.
\newblock Manifold-constrained nucleus-level denoising diffusion model for
  structure-based drug design.
\newblock \emph{arXiv preprint arXiv:2409.10584}, 2024{\natexlab{a}}.

\bibitem[Liu et~al.(2024{\natexlab{b}})Liu, Qiu, Feng, Xiu, Xue, Yu, Feng, Liu,
  Heo, Peng, et~al.]{liu2024boft}
Weiyang Liu, Zeju Qiu, Yao Feng, Yuliang Xiu, Yuxuan Xue, Longhui Yu, Haiwen
  Feng, Zhen Liu, Juyeon Heo, Songyou Peng, et~al.
\newblock Parameter-efficient orthogonal finetuning via butterfly
  factorization.
\newblock In \emph{ICLR}, 2024{\natexlab{b}}.

\bibitem[Liu et~al.(2023)Liu, Feng, Black, Nowrouzezahrai, Paull, and
  Liu]{liu2023meshdiffusion}
Zhen Liu, Yao Feng, Michael~J Black, Derek Nowrouzezahrai, Liam Paull, and
  Weiyang Liu.
\newblock Meshdiffusion: Score-based generative 3d mesh modeling.
\newblock In \emph{ICLR}, 2023.

\bibitem[Liu et~al.(2024{\natexlab{c}})Liu, Feng, Xiu, Liu, Paull, Black, and
  Sch{\"o}lkopf]{liu2024gshell}
Zhen Liu, Yao Feng, Yuliang Xiu, Weiyang Liu, Liam Paull, Michael~J Black, and
  Bernhard Sch{\"o}lkopf.
\newblock Ghost on the shell: An expressive representation of general 3d
  shapes.
\newblock In \emph{ICLR}, 2024{\natexlab{c}}.

\bibitem[Lou et~al.(2024)Lou, Meng, and Ermon]{lou2024discrete}
Aaron Lou, Chenlin Meng, and Stefano Ermon.
\newblock Discrete diffusion modeling by estimating the ratios of the data
  distribution.
\newblock In \emph{ICML}, 2024.

\bibitem[Lovelace et~al.(2024)Lovelace, Kishore, Wan, Shekhtman, and
  Weinberger]{lovelace2024latent}
Justin Lovelace, Varsha Kishore, Chao Wan, Eliot Shekhtman, and Kilian~Q
  Weinberger.
\newblock Latent diffusion for language generation.
\newblock In \emph{NeurIPS}, 2024.

\bibitem[Lu et~al.(2023)Lu, Zhou, Bao, Chen, Li, and Zhu]{lu2022dpm}
Cheng Lu, Yuhao Zhou, Fan Bao, Jianfei Chen, Chongxuan Li, and Jun Zhu.
\newblock Dpm-solver++: Fast solver for guided sampling of diffusion
  probabilistic models.
\newblock In \emph{ICLR}, 2023.

\bibitem[Malkin et~al.(2022)Malkin, Jain, Bengio, Sun, and
  Bengio]{malkin2022trajectory}
Nikolay Malkin, Moksh Jain, Emmanuel Bengio, Chen Sun, and Yoshua Bengio.
\newblock Trajectory balance: Improved credit assignment in gflownets.
\newblock In \emph{NeurIPS}, 2022.

\bibitem[Malkin et~al.(2023)Malkin, Lahlou, Deleu, Ji, Hu, Everett, Zhang, and
  Bengio]{malkin2022gfnhvi}
Nikolay Malkin, Salem Lahlou, Tristan Deleu, Xu~Ji, Edward Hu, Katie Everett,
  Dinghuai Zhang, and Yoshua Bengio.
\newblock {GFlowNets} and variational inference.
\newblock In \emph{ICLR}, 2023.

\bibitem[Mirhoseini et~al.(2020)Mirhoseini, Goldie, Yazgan, Jiang, Songhori,
  Wang, Lee, Johnson, Pathak, Bae, et~al.]{mirhoseini2020chip}
Azalia Mirhoseini, Anna Goldie, Mustafa Yazgan, Joe Jiang, Ebrahim Songhori,
  Shen Wang, Young-Joon Lee, Eric Johnson, Omkar Pathak, Sungmin Bae, et~al.
\newblock Chip placement with deep reinforcement learning.
\newblock \emph{arXiv preprint arXiv:2004.10746}, 2020.

\bibitem[Nachum et~al.(2017)Nachum, Norouzi, Xu, and
  Schuurmans]{nachum2017bridging}
Ofir Nachum, Mohammad Norouzi, Kelvin Xu, and Dale Schuurmans.
\newblock Bridging the gap between value and policy based reinforcement
  learning.
\newblock In \emph{NeurIPS}, 2017.

\bibitem[Ouyang et~al.(2022)Ouyang, Wu, Jiang, Almeida, Wainwright, Mishkin,
  Zhang, Agarwal, Slama, Ray, et~al.]{ouyang2022training}
Long Ouyang, Jeffrey Wu, Xu~Jiang, Diogo Almeida, Carroll Wainwright, Pamela
  Mishkin, Chong Zhang, Sandhini Agarwal, Katarina Slama, Alex Ray, et~al.
\newblock Training language models to follow instructions with human feedback.
\newblock In \emph{NeurIPS}, 2022.

\bibitem[Pan et~al.(2023{\natexlab{a}})Pan, Malkin, Zhang, and
  Bengio]{pan2023better}
Ling Pan, Nikolay Malkin, Dinghuai Zhang, and Yoshua Bengio.
\newblock Better training of gflownets with local credit and incomplete
  trajectories.
\newblock In \emph{ICML}, 2023{\natexlab{a}}.

\bibitem[Pan et~al.(2023{\natexlab{b}})Pan, Zhang, Courville, Huang, and
  Bengio]{pan2023gafn}
Ling Pan, Dinghuai Zhang, Aaron Courville, Longbo Huang, and Yoshua Bengio.
\newblock Generative augmented flow networks.
\newblock In \emph{ICLR}, 2023{\natexlab{b}}.

\bibitem[Pan et~al.(2023{\natexlab{c}})Pan, Zhang, Jain, Huang, and
  Bengio]{pan2023stochastic}
Ling Pan, Dinghuai Zhang, Moksh Jain, Longbo Huang, and Yoshua Bengio.
\newblock Stochastic generative flow networks.
\newblock In \emph{UAI}, 2023{\natexlab{c}}.

\bibitem[Poole et~al.(2023)Poole, Jain, Barron, and
  Mildenhall]{poole2023dreamfusion}
Ben Poole, Ajay Jain, Jonathan~T Barron, and Ben Mildenhall.
\newblock Dreamfusion: Text-to-3d using 2d diffusion.
\newblock In \emph{ICLR}, 2023.

\bibitem[Prabhudesai et~al.(2023)Prabhudesai, Goyal, Pathak, and
  Fragkiadaki]{prabhudesai2023aligning}
Mihir Prabhudesai, Anirudh Goyal, Deepak Pathak, and Katerina Fragkiadaki.
\newblock Aligning text-to-image diffusion models with reward backpropagation.
\newblock \emph{arXiv preprint arXiv:2310.03739}, 2023.

\bibitem[Qiu et~al.(2023)Qiu, Liu, Feng, Xue, Feng, Liu, Zhang, Weller, and
  Sch{\"o}lkopf]{qiu2023controlling}
Zeju Qiu, Weiyang Liu, Haiwen Feng, Yuxuan Xue, Yao Feng, Zhen Liu, Dan Zhang,
  Adrian Weller, and Bernhard Sch{\"o}lkopf.
\newblock Controlling text-to-image diffusion by orthogonal finetuning.
\newblock In \emph{NeurIPS}, 2023.

\bibitem[Rafailov et~al.(2023)Rafailov, Sharma, Mitchell, Manning, Ermon, and
  Finn]{rafailov2023direct}
Rafael Rafailov, Archit Sharma, Eric Mitchell, Christopher~D Manning, Stefano
  Ermon, and Chelsea Finn.
\newblock Direct preference optimization: Your language model is secretly a
  reward model.
\newblock In \emph{NeurIPS}, 2023.

\bibitem[Rombach et~al.(2022)Rombach, Blattmann, Lorenz, Esser, and
  Ommer]{rombach2022high}
Robin Rombach, Andreas Blattmann, Dominik Lorenz, Patrick Esser, and Bj{\"o}rn
  Ommer.
\newblock High-resolution image synthesis with latent diffusion models.
\newblock In \emph{CVPR}, 2022.

\bibitem[Ronneberger et~al.(2015)Ronneberger, Fischer, and
  Brox]{ronneberger2015unetcn}
Olaf Ronneberger, Philipp Fischer, and Thomas Brox.
\newblock U-net: Convolutional networks for biomedical image segmentation.
\newblock In \emph{MICCAI}, 2015.

\bibitem[Saharia et~al.(2022)Saharia, Chan, Saxena, Li, Whang, Denton,
  Ghasemipour, Gontijo~Lopes, Karagol~Ayan, Salimans,
  et~al.]{saharia2022photorealistic}
Chitwan Saharia, William Chan, Saurabh Saxena, Lala Li, Jay Whang, Emily~L
  Denton, Kamyar Ghasemipour, Raphael Gontijo~Lopes, Burcu Karagol~Ayan, Tim
  Salimans, et~al.
\newblock Photorealistic text-to-image diffusion models with deep language
  understanding.
\newblock In \emph{NeurIPS}, 2022.

\bibitem[Schulman et~al.(2015)Schulman, Levine, Abbeel, Jordan, and
  Moritz]{schulman2015trust}
John Schulman, Sergey Levine, Pieter Abbeel, Michael Jordan, and Philipp
  Moritz.
\newblock Trust region policy optimization.
\newblock In \emph{ICML}, 2015.

\bibitem[Schulman et~al.(2017)Schulman, Wolski, Dhariwal, Radford, and
  Klimov]{schulman2017proximal}
John Schulman, Filip Wolski, Prafulla Dhariwal, Alec Radford, and Oleg Klimov.
\newblock Proximal policy optimization algorithms.
\newblock \emph{arXiv preprint arXiv:1707.06347}, 2017.

\bibitem[Sendera et~al.(2024)Sendera, Kim, Mittal, Lemos, Scimeca,
  Rector-Brooks, Adam, Bengio, and Malkin]{sendera2024improved}
Marcin Sendera, Minsu Kim, Sarthak Mittal, Pablo Lemos, Luca Scimeca, Jarrid
  Rector-Brooks, Alexandre Adam, Yoshua Bengio, and Nikolay Malkin.
\newblock Improved off-policy training of diffusion samplers.
\newblock In \emph{NeurIPS}, 2024.

\bibitem[Shen et~al.(2023)Shen, Bengio, Hajiramezanali, Loukas, Cho, and
  Biancalani]{shen2023towardsua}
Max~W Shen, Emmanuel Bengio, Ehsan Hajiramezanali, Andreas Loukas, Kyunghyun
  Cho, and Tommaso Biancalani.
\newblock Towards understanding and improving gflownet training.
\newblock In \emph{ICML}, 2023.

\bibitem[Silver et~al.(2016)Silver, Huang, Maddison, Guez, Sifre, van~den
  Driessche, Schrittwieser, Antonoglou, Panneershelvam, Lanctot, Dieleman,
  Grewe, Nham, Kalchbrenner, Sutskever, Lillicrap, Leach, Kavukcuoglu, Graepel,
  and Hassabis]{Silver2016MasteringTG}
David Silver, Aja Huang, Chris~J. Maddison, Arthur Guez, L.~Sifre, George
  van~den Driessche, Julian Schrittwieser, Ioannis Antonoglou, Vedavyas
  Panneershelvam, Marc Lanctot, Sander Dieleman, Dominik Grewe, John Nham, Nal
  Kalchbrenner, Ilya Sutskever, Timothy~P. Lillicrap, Madeleine Leach, Koray
  Kavukcuoglu, Thore Graepel, and Demis Hassabis.
\newblock Mastering the game of go with deep neural networks and tree search.
\newblock \emph{Nature}, 529:\penalty0 484--489, 2016.
\newblock URL \url{https://api.semanticscholar.org/CorpusID:515925}.

\bibitem[Song et~al.(2021{\natexlab{a}})Song, Meng, and
  Ermon]{song2021denoising}
Jiaming Song, Chenlin Meng, and Stefano Ermon.
\newblock Denoising diffusion implicit models.
\newblock In \emph{ICLR}, 2021{\natexlab{a}}.

\bibitem[Song et~al.(2023)Song, Zhang, Yin, Mardani, Liu, Kautz, Chen, and
  Vahdat]{Song2023LossGuidedDM}
Jiaming Song, Qinsheng Zhang, Hongxu Yin, Morteza Mardani, Ming-Yu Liu, Jan
  Kautz, Yongxin Chen, and Arash Vahdat.
\newblock Loss-guided diffusion models for plug-and-play controllable
  generation.
\newblock In \emph{ICML}, 2023.

\bibitem[Song et~al.(2021{\natexlab{b}})Song, Sohl-Dickstein, Kingma, Kumar,
  Ermon, and Poole]{song2021score}
Yang Song, Jascha Sohl-Dickstein, Diederik~P Kingma, Abhishek Kumar, Stefano
  Ermon, and Ben Poole.
\newblock Score-based generative modeling through stochastic differential
  equations.
\newblock In \emph{ICLR}, 2021{\natexlab{b}}.

\bibitem[Stiennon et~al.(2022)Stiennon, Ouyang, Wu, Ziegler, Lowe, Voss,
  Radford, Amodei, and Christiano]{stiennon2022learningsummarizehumanfeedback}
Nisan Stiennon, Long Ouyang, Jeff Wu, Daniel~M. Ziegler, Ryan Lowe, Chelsea
  Voss, Alec Radford, Dario Amodei, and Paul Christiano.
\newblock Learning to summarize from human feedback.
\newblock In \emph{NeurIPS}, 2022.

\bibitem[Sutton(1988)]{Sutton1988LearningTP}
Richard~S. Sutton.
\newblock Learning to predict by the methods of temporal differences.
\newblock \emph{Machine Learning}, 3:\penalty0 9--44, 1988.

\bibitem[Tiapkin et~al.(2024)Tiapkin, Morozov, Naumov, and
  Vetrov]{tiapkin2024generative}
Daniil Tiapkin, Nikita Morozov, Alexey Naumov, and Dmitry~P Vetrov.
\newblock Generative flow networks as entropy-regularized rl.
\newblock In \emph{AISTATS}, 2024.

\bibitem[Uehara et~al.(2024)Uehara, Zhao, Black, Hajiramezanali, Scalia,
  Diamant, Tseng, Biancalani, and Levine]{uehara2024fine}
Masatoshi Uehara, Yulai Zhao, Kevin Black, Ehsan Hajiramezanali, Gabriele
  Scalia, Nathaniel~Lee Diamant, Alex~M Tseng, Tommaso Biancalani, and Sergey
  Levine.
\newblock Fine-tuning of continuous-time diffusion models as
  entropy-regularized control.
\newblock \emph{arXiv preprint arXiv:2402.15194}, 2024.

\bibitem[van Hasselt et~al.(2018)van Hasselt, Doron, Strub, Hessel, Sonnerat,
  and Modayil]{van2018deep}
Hado van Hasselt, Yotam Doron, Florian Strub, Matteo Hessel, Nicolas Sonnerat,
  and Joseph Modayil.
\newblock Deep reinforcement learning and the deadly triad.
\newblock \emph{arXiv preprint arXiv:1812.02648}, 2018.

\bibitem[Venkatraman et~al.(2024)Venkatraman, Jain, Scimeca, Kim, Sendera,
  Hasan, Rowe, Mittal, Lemos, Bengio, et~al.]{venkatraman2024amortizing}
Siddarth Venkatraman, Moksh Jain, Luca Scimeca, Minsu Kim, Marcin Sendera,
  Mohsin Hasan, Luke Rowe, Sarthak Mittal, Pablo Lemos, Emmanuel Bengio, et~al.
\newblock Amortizing intractable inference in diffusion models for vision,
  language, and control.
\newblock In \emph{NeurIPS}, 2024.

\bibitem[Wu et~al.(2024)Wu, Trippe, Naesseth, Blei, and
  Cunningham]{wu2024practical}
Luhuan Wu, Brian Trippe, Christian Naesseth, David Blei, and John~P Cunningham.
\newblock Practical and asymptotically exact conditional sampling in diffusion
  models.
\newblock In \emph{NeurIPS}, 2024.

\bibitem[Wu et~al.(2023{\natexlab{a}})Wu, Hao, Sun, Chen, Zhu, Zhao, and
  Li]{wu2023humanv2}
Xiaoshi Wu, Yiming Hao, Keqiang Sun, Yixiong Chen, Feng Zhu, Rui Zhao, and
  Hongsheng Li.
\newblock Human preference score v2: A solid benchmark for evaluating human
  preferences of text-to-image synthesis.
\newblock \emph{arXiv preprint arXiv:2306.09341}, 2023{\natexlab{a}}.

\bibitem[Wu et~al.(2023{\natexlab{b}})Wu, Sun, Zhu, Zhao, and Li]{wu2023human}
Xiaoshi Wu, Keqiang Sun, Feng Zhu, Rui Zhao, and Hongsheng Li.
\newblock Human preference score: Better aligning text-to-image models with
  human preference.
\newblock In \emph{ICCV}, 2023{\natexlab{b}}.

\bibitem[Xu et~al.(2024)Xu, Liu, Wu, Tong, Li, Ding, Tang, and
  Dong]{xu2024imagereward}
Jiazheng Xu, Xiao Liu, Yuchen Wu, Yuxuan Tong, Qinkai Li, Ming Ding, Jie Tang,
  and Yuxiao Dong.
\newblock Imagereward: Learning and evaluating human preferences for
  text-to-image generation.
\newblock In \emph{NeurIPS}, 2024.

\bibitem[Xu et~al.(2022)Xu, Yu, Song, Shi, Ermon, and Tang]{xu2022geodiff}
Minkai Xu, Lantao Yu, Yang Song, Chence Shi, Stefano Ermon, and Jian Tang.
\newblock Geodiff: A geometric diffusion model for molecular conformation
  generation.
\newblock In \emph{ICLR}, 2022.

\bibitem[Yun et~al.(2025)Yun, Zhang, Park, and Pan]{yun2025learning}
Taeyoung Yun, Dinghuai Zhang, Jinkyoo Park, and Ling Pan.
\newblock Learning to sample effective and diverse prompts for text-to-image
  generation.
\newblock \emph{arXiv preprint arXiv:2502.11477}, 2025.

\bibitem[Zhang et~al.(2023{\natexlab{a}})Zhang, Rainone, Peschl, and
  Bondesan]{Zhang2023RobustSW}
David~W Zhang, Corrado Rainone, Markus Peschl, and Roberto Bondesan.
\newblock Robust scheduling with gflownets.
\newblock In \emph{ICLR}, 2023{\natexlab{a}}.

\bibitem[Zhang et~al.(2022{\natexlab{a}})Zhang, Chen, Malkin, and
  Bengio]{zhang2022unifying}
Dinghuai Zhang, Ricky T.~Q. Chen, Nikolay Malkin, and Yoshua Bengio.
\newblock Unifying generative models with {GFlowNets} and beyond.
\newblock \emph{arXiv preprint arXiv:2209.02606v2}, 2022{\natexlab{a}}.

\bibitem[Zhang et~al.(2022{\natexlab{b}})Zhang, Malkin, Liu, Volokhova,
  Courville, and Bengio]{zhang2022generative}
Dinghuai Zhang, Nikolay Malkin, Zhen Liu, Alexandra Volokhova, Aaron Courville,
  and Yoshua Bengio.
\newblock Generative flow networks for discrete probabilistic modeling.
\newblock In \emph{ICML}, 2022{\natexlab{b}}.

\bibitem[Zhang et~al.(2023{\natexlab{b}})Zhang, Dai, Malkin, Courville, Bengio,
  and Pan]{zhang2023let}
Dinghuai Zhang, Hanjun Dai, Nikolay Malkin, Aaron~C Courville, Yoshua Bengio,
  and Ling Pan.
\newblock Let the flows tell: Solving graph combinatorial problems with
  gflownets.
\newblock In \emph{NeurIPS}, 2023{\natexlab{b}}.

\bibitem[Zhang et~al.(2024{\natexlab{a}})Zhang, Chen, Liu, Courville, and
  Bengio]{zhang2024diffusion}
Dinghuai Zhang, Ricky T.~Q. Chen, Cheng-Hao Liu, Aaron Courville, and Yoshua
  Bengio.
\newblock Diffusion generative flow samplers: Improving learning signals
  through partial trajectory optimization.
\newblock In \emph{ICLR}, 2024{\natexlab{a}}.

\bibitem[Zhang et~al.(2024{\natexlab{b}})Zhang, Pan, Chen, Courville, and
  Bengio]{zhang2024distributional}
Dinghuai Zhang, Ling Pan, Ricky T.~Q. Chen, Aaron Courville, and Yoshua Bengio.
\newblock Distributional {GF}lownets with quantile flows.
\newblock \emph{Transactions on Machine Learning Research}, 2024{\natexlab{b}}.

\bibitem[Zhang et~al.(2025)Zhang, Zhang, Gu, ZHANG, Susskind, Jaitly, and
  Zhai]{zhang2025improving}
Dinghuai Zhang, Yizhe Zhang, Jiatao Gu, Ruixiang ZHANG, Joshua~M. Susskind,
  Navdeep Jaitly, and Shuangfei Zhai.
\newblock Improving {GF}lownets for text-to-image diffusion alignment.
\newblock \emph{Transactions on Machine Learning Research}, 2025.

\bibitem[Zhang et~al.(2024{\natexlab{c}})Zhang, Wang, Zhang, Qiu, Pang, Jiang,
  Yang, Xu, and Yu]{zhang2024clay}
Longwen Zhang, Ziyu Wang, Qixuan Zhang, Qiwei Qiu, Anqi Pang, Haoran Jiang, Wei
  Yang, Lan Xu, and Jingyi Yu.
\newblock Clay: A controllable large-scale generative model for creating
  high-quality 3d assets.
\newblock \emph{ACM Transactions on Graphics (TOG)}, 2024{\natexlab{c}}.

\bibitem[Zhao et~al.(2024)Zhao, Brekelmans, Makhzani, and
  Grosse]{zhao2024twist}
Stephen Zhao, Rob Brekelmans, Alireza Makhzani, and Roger~Baker Grosse.
\newblock Probabilistic inference in language models via twisted sequential
  monte carlo.
\newblock In \emph{ICML}, 2024.

\bibitem[Zhou et~al.(2024)Zhou, Yan, Layne, Malkin, Zhang, Jain, Blanchette,
  and Bengio]{zhou2024phylogfn}
Ming~Yang Zhou, Zichao Yan, Elliot Layne, Nikolay Malkin, Dinghuai Zhang, Moksh
  Jain, Mathieu Blanchette, and Yoshua Bengio.
\newblock Phylo{GFN}: Phylogenetic inference with generative flow networks.
\newblock In \emph{ICLR}, 2024.

\bibitem[Ziegler et~al.(2019)Ziegler, Stiennon, Wu, Brown, Radford, Amodei,
  Christiano, and Irving]{ziegler2019fine}
Daniel~M Ziegler, Nisan Stiennon, Jeffrey Wu, Tom~B Brown, Alec Radford, Dario
  Amodei, Paul Christiano, and Geoffrey Irving.
\newblock Fine-tuning language models from human preferences.
\newblock \emph{arXiv preprint arXiv:1909.08593}, 2019.

\end{thebibliography}
\bibliographystyle{iclr2025_conference}

\clearpage
\appendix

\addcontentsline{toc}{section}{Appendix} %
\renewcommand \thepart{} %
\renewcommand \partname{}
\part{\Large{\centerline{Appendix}}}
\parttoc
\newpage

\section{Overall algorithm}

\begin{algorithm}[H]
  \caption{\methodname Diffusion Finetuning with \resgraddb}
  \label{alg:resgraddb}
  \begin{algorithmic}[1]
    \State \textbf{Inputs:} Pretrained diffusion model $f_{\theta^\#}$ with the MDP constructed from some Markovian sampler, reward function $R(\cdot)$
    \State \textbf{Initialization:} Model to finetune $f_{\theta}$ with $\theta = \theta^\#$, residual flow score function $g_\phi(\cdot)$.
    \State Sample the initial batch of trajectories $\mathcal{D}_\text{prev} = \{(x_1, ..., x_T)_i\}_{i=1...N}$ with the current finetuned diffusion model $f_\theta$.
    \State Set $\theta^\dagger \leftarrow \theta$.
    \While{not converged}
        \State \parbox[t]{313pt}{Sample a batch of trajectories $\mathcal{D}_\text{curr} = \{(x_1, ..., x_T)_i\}_{i=1...N}$ with the finetuned diffusion model.\strut}
        \State \parbox[t]{313pt}{Subsample the time steps to train with: the full set $\mathcal{T}_i = \{1, ..., T\}$ or the sampled set $\mathcal{T}_i = \text{Sample-N}(\{1, ..., T\})$ where $\text{Sample-N}$ is some unbiased sampling algorithm to randomly pick $N$ samples.\strut}
        \State \parbox[t]{313pt}{Compute the loss \\
        $\sum_{t \in \mathcal{T}_i, (x_{1:T})_i \in \mathcal{D}_\text{prev}} L_{\nabla\text{DB-FL-res}}(x_t, x_{t+1}; \theta, \theta^\#, \phi, R) 
        \\ 
        + L_{\nabla\text{DB-FL-terminal}}(x_T; \phi) + \lambda \lVert f_\theta(x_t) - f_{\theta^\dagger}(x_t)\rVert^2$.\strut}
        \State Set $\theta^\dagger \leftarrow \theta$.
        \State Update the diffusion model and the residual flow score function.
        \State Set $\mathcal{D}_\text{prev} \gets \mathcal{D}_\text{curr}$.
    \EndWhile
    \State \Return finetuned model $f_\theta$.
  \end{algorithmic}
\end{algorithm}

\newpage
\section{Algorithmic details}

\subsection{\graddb objective as a statistical divergence}
\label{sec:fisher-div}
$L_{\nabla\text{DB}}$ (Equation~\ref{eqn:grad-db}) is analogous to a Fisher divergence (up to a constant scale) if we always use on-policy samples to update the diffusion model and the flow function:
\begin{align}
    & \ D_\text{Fisher}\Big(
        P_F(x_{t+1} | x_t) 
        \Big|\Big| 
        \frac{P_B(x_t | x_{t+1}) F(x_{t+1})}{F(x_{t})}
    \Big) \notag \\
    = & \ \frac{1}{2} \expt_{x_{t+1} \sim P_F(x_{t+1} | x_t)} \Big|\Big|
        \nabla_{x_{t+1}} \log P_B(x_t | x_{t+1}) - \nabla_{x_{t+1}} \log \frac{P_B(x_t | x_{t+1}) F(x_{t+1})}{F(x_{t})}
    \Big|\Big|^2 \notag \\
    = & \ \frac{1}{2} \expt_{x_{t+1} \sim P_F(x_{t+1} | x_t)} L_{\nabla\text{DB}}(x_t, x_{t+1}).
\end{align}

\subsection{Proof of Proposition~\ref{prop:valid_gfn}}
\label{sec:proof_prop_validgfn}
\begin{proof}
When the training objectives equal $0$ for all states, we would have
\begin{align}
    \nabla_{x_{t+1}}\log P_F(x_{t+1} | x_t) &= \nabla_{x_{t+1}} \log P_B(x_t | x_{t+1}) + \nabla_{x_{t+1}} \log F_{t+1}(x_{t+1}) \\
    \nabla_{x_{t}}\log P_F(x_{t+1} | x_t) &= \nabla_{x_{t}} \log P_B(x_t | x_{t+1}) - \nabla_{x_{t}} \log F_t(x_{t})\\
     \nabla_{x_T} \log F(x_T) &= \beta\nabla_{x_T} \log R(x_T)
\end{align}
for any trajectory $(x_0, \ldots, x_T)$. 

Through indefinite integral, these indicate that there exist a function $C_t(x_t)$ satisfies
\begin{align}
    C_t(x_t)P_F(x_{t+1} | x_t) &= F_{t+1}(x_{t+1})P_B(x_t | x_{t+1}) \\
    F_t(x_t)P_F(x_{t+1} | x_t) &= C_{t+1}(x_{t+1})P_B(x_t | x_{t+1}) \\
    F(x_T)&\propto R(x_T)^\beta.
\end{align}
Therefore, we have
\begin{align}
    \frac{C_t(x_t)}{F_t(x_t)} = \frac{F_{t+1}(x_{t+1})}{C_{t+1}(x_{t+1})}, \quad\forall (x_t, x_{t+1}).
\end{align}
The right hand side does not depend on $x_{t}$, therefore, the left hand side is a constant. So we have
\begin{align}
    C_t(x_t)\propto F_t(x_t),\quad\forall t.
\end{align}
The probability of generating a data $x_T$ then equals
\begin{align}
    P_F(x_T) &= \int P_0(x_0|\oslash) \prod_t P_F(x_{t+1}|x_t) d x_{0:{T-1}} \\
    &=  \int F_0(x_0) \prod_t\frac{F_{t+1}(x_{t+1})P_B(x_t | x_{t+1})}{C_t(x_t)} d x_{0:{T-1}} \\
    &\propto \int F_0(x_0) \prod_t\frac{F_{t+1}(x_{t+1})P_B(x_t | x_{t+1})}{F_t(x_t)} d x_{0:{T-1}} \\
    &\propto F(x_T) \int \prod_t P_B(x_t | x_{t+1})d x_{0:{T-1}} \\
    &\propto F(x_T)\propto R(x_T)^\beta,
\end{align}
which proves the validity of the \methodname algorithm.
\end{proof}

\subsection{Proof of Proposition~\ref{prop:valid-res-gfn}}

When the training objectives equal $0$ for all states, we have
\begin{align}
    \nabla_{x_{t+1}} \log \tilde{P}_F(x_{t+1} | x_t) &= \nabla_{x_{t+1}} \log \tilde{F}(x_{t+1}) \\
    \nabla_{x_{t}} \log \tilde{P}_F(x_{t+1} | x_t) &= - \nabla_{x_{t}} \log \tilde{F}(x_{t}) \\
    \nabla_{x_T} \log \tilde{F}(x_T) &= \beta \nabla_{x_T} \log R(x_T)
\end{align}
for any trajectory $(x_0, \ldots, x_T)$. 

Through indefinite integral, these indicate that there exist a function $C_t(x_t)$ satisfies
\begin{align}
    C_t(x_t) \tilde{P}_F(x_{t+1} | x_t) &= \tilde{F}(x_{t+1})\\
    \tilde{F}(x_{t})\tilde{P}_F(x_{t+1} | x_t) &= C_{t+1}(x_{t+1}) \\
    \tilde{F}(x_T) &\propto  R(x_T)^\beta. 
\end{align}
Thus we have
\begin{align}
    \frac{C_t(x_t)}{\tilde F_t(x_t)} = \frac{\tilde F_{t+1}(x_{t+1})}{C_{t+1}(x_{t+1})}, \quad\forall (x_t, x_{t+1}).
\end{align}
The right hand side does not depend on $x_{t}$, therefore, the left hand side is a constant. So we have
\begin{align}
    C_t(x_t)\propto \tilde F_t(x_t),\quad\forall t.
\end{align}
The probability of generating a data $x_T$ then equals
\begin{align}
    P_F(x_T) &=  \int P_0(x_0|\oslash) \prod_t P_F(x_{t+1}|x_t) d x_{0:{T-1}} \\
    &= \int P^\#_0(x_0|\oslash) \prod_t P^\#_F(x_{t+1}|x_t) \prod_t \tilde P_F(x_{t+1}|x_t) d x_{0:{T-1}} \\
    &= \int P^\#_0(x_0|\oslash) \prod_t P^\#_F(x_{t+1}|x_t) \frac{\tilde F_{t+1}(x_{t+1})}{C_t(x_t)} d x_{0:{T-1}} \\
    &\propto \tilde F_T(x_T) \int P^\#_0(x_0|\oslash) \prod_t P^\#_F(x_{t+1}|x_t) d x_{0:{T-1}} \\
    &\propto  R(x_T)^\beta P_F^\#(x_T).
\end{align}

Hence zero $\nabla$-DB losses in both forward and reverse direction for all $(x_t, x_{t+1})$ pairs yield the ideal tilted distribution $R(x_T)^\beta P_F^\#(x_T)$.

\subsection{Relationship between residual DB and Trajectory Balance}
\label{sec:relationship-tb}

Here we illustrate the Remark~\ref{remark:relative-tb}.
A different but equivalent condition for GFlowNets states:

\textbf{Trajectory Balance (TB) \cite{malkin2022trajectory}.} The following TB condition must hold for any transition sequence $(s_0, s_1, ..., s_N)$ where $x_0$ is the unique starting state in the MDP and $s_N$ is a terminal state, given a GFlowNet with the forward policy $P_F(s' | s)$ and the backward policy $P_B(s | s')$:

\begin{align}
    \log \frac{Z \prod P_F(s' | s)}{R(x_T) \prod P_B(s | s')} = 0
\end{align}

where $Z = F(x_0)$ is the total flow and $R(x_T) = F(x_T)$ the reward. The proof is immediate with a telescoping product of the DB condition.

With an (ideal) pretrained model $P_F^\#$ and the satisfication of the finetuning objective of Equation~\ref{eqn:finetuning-objective}, one can prove the conclusion in \cite{venkatraman2024amortizing}:

\begin{align}
\label{eqn:relative-tb}
    \log \frac{Z \prod P_F(s' | s)}{Z^\# \prod P_F^\#(s' | s)} = \beta \log R(x_T),
\end{align}

which is also an immediate result of a telescoping products of the \textit{residual} DB condition (which leads to Equation~\ref{eqn:residual-db}):

\begin{align}
    \log \tilde{P}_F(x_{t+1} | x_t) = \log \tilde{F}(x_{t+1}) - \log \tilde{F}(x_t).
\end{align}

While Equation~\ref{eqn:relative-tb} and residual DB are mathematically equivalent, implementation-wise TB in Equation~\ref{eqn:relative-tb} demands the whole sampling sequence be stored in the memory for gradient computation, or one has resort to the time-costly technique of gradient checkpointing. In comparison, with DB-based methods one may amortize the computational cost into flows at different time steps and therefore allow diffusion finetuning with flexible sampling sequence, of which the distribution approximation capacity and generation performance are generally greater.

\section{Generalization to the learnable $P_B$ setting}

If the backward process $P_B$ is learnable, the resulted $P_B$ does not cancel the backward propcess $P_B^\#$ in the pretrained model in the derivation of the \resgraddb condition. Let $\log \tilde{P}_B(x_{t} | x_{t+1}; \theta_B) = \log P_B(x_{t} | x_{t+1}; \theta_B) - \log P_B^\#(x_{t} | x_{t+1})$. We instead have the following \resgraddb losses:

\begin{align}
    \nabla_{x_{t+1}} \log \tilde{P}_F(x_{t+1} | x_t; \theta) - \nabla_{x_{t+1}} \log \tilde{P}_B(x_{t} | x_{t+1}; \theta_B) = \nabla_{x_{t+1}} \log \tilde{F}(x_{t+1}; \phi)
\end{align}

\begin{align}
    \nabla_{x_{t}} \log \tilde{P}_F(x_{t+1} | x_t; \theta) - \nabla_{x_{t}} \log \tilde{P}_B(x_{t} | x_{t+1}; \theta_B) = -\nabla_{x_{t}} \log \tilde{F}(x_{t}; \phi)
\end{align}

Therefore, we have the general \resgraddb losses:

\begin{align}
    & ~ L_\text{forward, general}(x_t, x_{t+1}) 
    \notag \\
    = & ~ \Bignorm{
        \nabla_{x_{t+1}} \log \tilde{P}_F(x_{t+1} | x_t; \theta) - \nabla_{x_{t+1}} \log \tilde{P}_B(x_{t} | x_{t+1}; \theta_B)
        - \nabla_{x_{t+1}} \log \tilde{F}(x_{t+1}; \phi)
    }^2,
\end{align}

and

\begin{align}
    & ~ L_\text{reverse, general}(x_t, x_{t+1})
    \notag \\
    = & ~ \Bignorm{
        \nabla_{x_{t}} \log \tilde{P}_F(x_{t+1} | x_t; \theta) - \nabla_{x_{t}} \log \tilde{P}_B(x_{t} | x_{t+1}; \theta_B)
        + \nabla_{x_{t}} \log \tilde{F}(x_{t}; \phi)
    }^2.
\end{align}

\section{Commons and differences between GFlowNet and soft RL}
\label{sec:soft-rl}

\subsection{GFlowNet in the lens of soft RL}
\label{sec:soft-rl-special-case}

We restate the finding~\cite{tiapkin2024generative, deleu2025generative} that, by constructing a special MDP, we may prove that GFlowNet and soft RL are equivalent when $P_B$ is fixed.

\begin{theorem}
    Let $\mathcal{M} = (\mathcal{G}, r)$ be an MDP define on a DAG $\mathcal{G}$ and a reward function $r(s,s')$ (defined for any transition $s \to s'$ on $\mathcal{G}$). Let $P_B$ be an arbitrary backward transition probability on $\mathcal{G}$. Suppose for any complete trajectory $\tau = (s_0, s_1, \ldots, s_T)$, we have the intermediate reward
    $$
        r(s_t, s_{t+1}) =
        \begin{cases}
            \frac{1}{\beta} \log P_B(s_t \mid s_{t+1}) &, t \neq T-1, \\
            \log R(s_T) + \frac{1}{\beta} \log P_B(s_t \mid s_{t+1}) &, t = T-1.
        \end{cases}
    $$
    Then the optimal policy with respect to the following objective (where $\mathcal{H}$ is the entropy function)
    $$
        \pi^\star = \arg\max_{\pi} \mathbb{E}_{\tau \sim \pi} 
        \left[ \sum_{t=0}^{T} r(s_t, s_{t+1}) + \frac{1}{\beta} \mathcal{H}\left( \pi(\cdot \mid s_t) \right) \right]
    $$
    samples terminating states $x$ with probability proportional to $R(x)^\beta$.
\end{theorem}

\begin{remark} The introduction of the intermediate reward terms $\log P_B(s_t | s_{t+1})$ (with fixed $P_B$) essentially anchors the optimal solution around the reference backward policy, which is rather natural from the generative model perspective and beneficial in creating better optimization landscapes.
\end{remark}

\begin{remark} Soft RL does not cover the case where $P_B$ can be jointly learned.
\end{remark}

\subsection{Derivation of \graddb with soft Q-learning}

With soft Q-learning~\cite{haarnoja2017reinforcement} (a special case of path consistency learning~\cite{nachum2017bridging}) in the MDP defined in Sec~\ref{sec:soft-rl-special-case} with a fixed $P_B$, one aims at

\begin{align}
    \Delta_{\mathrm{SQL}}(s \to s') \overset{\Delta}{=} Q(s, s') - \left( r(s, s') + V(s') \right) = 0
\end{align}

where the soft value function is $V(s') \overset{\Delta}{=} \frac{1}{\beta} \log \sum_{s': s \to s'} \exp\left( \beta Q(s', s'') \right)$ and $Q(s, s')$ is the soft Q-value function. One can show that

\begin{align}
    P_F(s' | s) = \frac{\exp(\beta Q(s, s'))}{\sum_{s'} \exp(\beta Q(s, s'))}.
\end{align}

Therefore, at optimality where $\Delta_{\mathrm{SQL}}(s \to s') = 0$ for all $s \to s'$, we can plug in the definitions above and obtain

\begin{align}
    \log P_F(s' | s) - \log P_B(s | s') = V(s') - V(s)
\end{align}

which is in the same form of the DB condition. Taking gradients of the above condition with respect to both $s$ and $s'$ leads to equivalent \graddb conditions.

\section{Corrections for non-ideal pretrained models}

For non-ideal pretrained models in which $P_F^\#$ does not match $P_B$ (\textit{i.e.}, the DB condition is violated), the original \resgraddb objective is apparently biased. We may introduce an additional learnable term $h(x_t, x_{t+1}; \psi)$ to compensate for this error, with which we have the corrected DB condition for the pretrained model:

\begin{align}
    \log P_F^\#(x_{t+1} | x_t) - \log P_B(x_t | x_{t+1}) = \log F^\#(x_{t+1}) - \log F^\#(x_{t}) - h(x_t, x_{t+1})
\end{align}

Hence, we have the following \resgraddb losses in the case of the imperfect pretrained model:

\begin{align}
    L_{\overrightarrow{\nabla}\text{DB-res-v2}}(x_{t}, x_{t+1})  
    =
    \Bignorm{ 
        \nabla_{x_{t+1}} \log \tilde{P}_F(x_{t+1} | x_t) - \nabla_{x_{t+1}} \log \tilde{F}(x_{t+1}) + \nabla_{x_{t+1}} h(x_t, x_{t+1}; \psi).
    }^2,
\end{align}

\begin{align}
    L_{\overleftarrow{\nabla}\text{DB-res-v2}}(x_{t}, x_{t+1})  
    =
    \Bignorm{ 
        \nabla_{x_{t}} \log \tilde{P}_F(x_{t+1} | x_t) + \nabla_{x_{t}} \log \tilde{F}(x_{t+1}) + \nabla_{x_{t}} h(x_t, x_{t+1}; \psi).
    }^2,
\end{align}

and the bidirectional \graddb loss for learning $h$ in the pretrained model:

\begin{align}
    L_{\overrightarrow{\nabla}\text{DB-pretrained}}(x_{t}, x_{t+1}, x_{t+2})  =
    \Bignorm{
        &
        \nabla_{x_{t+1}} \log P_F^\#(x_{t+1} | x_t) + \nabla_{x_{t+1}} \log P_F^\#(x_{t+2} | x_{t+1}) 
        \notag \\
        &
        - \nabla_{x_{t+1}} \log P_B(x_t | x_{t+1}) - \nabla_{x_{t+1}} \log P_B(x_{t+1} | x_{t+2})
        \notag \\
        &
        + \nabla_{x_{t+1}} h(x_t, x_{t+1}; \psi) + \nabla_{x_{t+1}} h(x_{t+1}, x_{t+2}; \psi)
    }^2.
\end{align}

\newpage
\section{MDP construction for diffusion models}
\label{sec:mdp_construction}

Many typical inference algorithms (or samplers) of diffusion models, including but not limited to DDPM~\cite{ho2020denoising}, DDIM~\cite{song2021denoising} and SDE-DPM-solver++~\cite{lu2022dpm}, can be abstracted into a loop with the following two steps (with the convention of GFlowNets on time indexing):

\begin{enumerate}
    \item \textbf{Computation of predicted clean data.} $\hat{x}_T = f(x_t, t)$.
    \item \textbf{Back-projection of the predicted clean data.} $x_{t+1} \sim \mathcal{N}(g(\hat{x}_T, t+1), \sigma_{t+1} I)$
\end{enumerate}

The MDP for diffusion models can therefore be simply constructed as:

\begin{itemize}
    \item State: $(x_t, t)$.
    \item Transition: $x_t \sim \mathcal{N}\Big(g(f(x_{t-1}, t-1), t), \sigma_t I\Big)$.
    \item Starting state: $x_0 \sim P(x_0) = \mathcal{N}(0, \sigma_0 I)$.
    \item Terminal state: $(x_T, T)$.
    \item Terminal reward: $R(x_T, T)$.
    \item Intermediate reward: $R(x_t, t) = 0$.
\end{itemize}

Since the intermediate rewards are always zero, for terminal rewards we simply write $R(x_T)$ without the second argument of $T$.

With this MDP defined, it is straightforward to apply policy gradient methods to optimize the return of the sampled on-policy trajectories. For instance, DDPO employs PPO~\cite{schulman2017proximal} to perform updates.

\section{Application to flow matching models}
\label{sec:flow_matching}

Flow matching models~\cite{lipman2023flow} are a popular and powerful class of generative models that sample points via simulation but of an ordinary differential equation (ODE) instead of an SDE. Specifically, a flow matching model defines a velocity field $v(x,t)$ and, by starting from a randomly initialized $x(0) = x_0 \sim \mathcal{N}(0, I)$, generate samples $x_1 = x(1)$ with $\dot x = v(x,t)$. Since this is a deterministic process, our \methodname does not na\"ively apply.

However, it is shown that a flow matching model above can be turned into an equivalent family of SDEs with the same probability flow, with an arbitrary diffusion term $\sigma(t)$~\cite{domingoenrich2024adjoint}:

\begin{align}
    \mathrm{d}X_t = \left( v(X_t, t) + \frac{\sigma(t)^2}{2 \beta_t \left( \frac{\dot{\alpha}_t}{\alpha_t} \beta_t - \dot{\beta}_t \right)} \left( v(X_t, t) - \frac{\dot{\alpha}_t}{\alpha_t} X_t \right) \right) \mathrm{d}t + \sigma(t) \, \mathrm{d}B_t, \quad X_0 \sim \mathcal{N}(0, I).
\end{align}

This is essentially a diffusion model but with a non-linear noising process. Since we can always fix this noising process (\textit{i.e.}, $P_B$ in our setting) during finetuning, we can use \methodname to obtain a finetuned diffusion model from this base diffusion model.

\newpage
\section{Finetuning convergence in wall time}

\begin{figure}[h]
    \centering
    \vspace{-3pt}
    \includegraphics[width=0.8\linewidth]{figs/aesthetic_evo_legend.pdf}%
    \vspace{-1.5em}

    \includegraphics[width=0.33\linewidth]{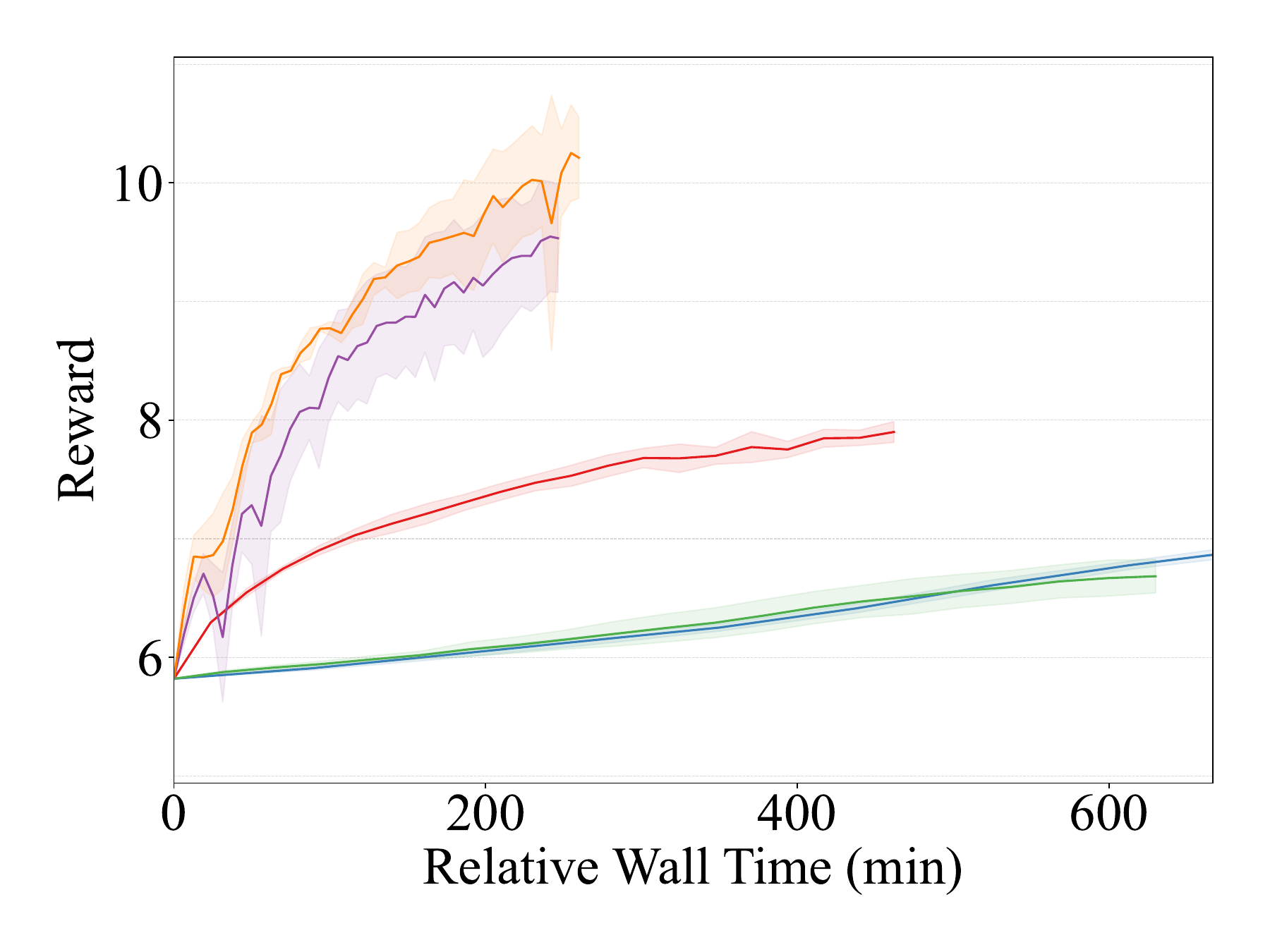}
    \hspace{-0.8em}
    \includegraphics[width=0.33\linewidth]{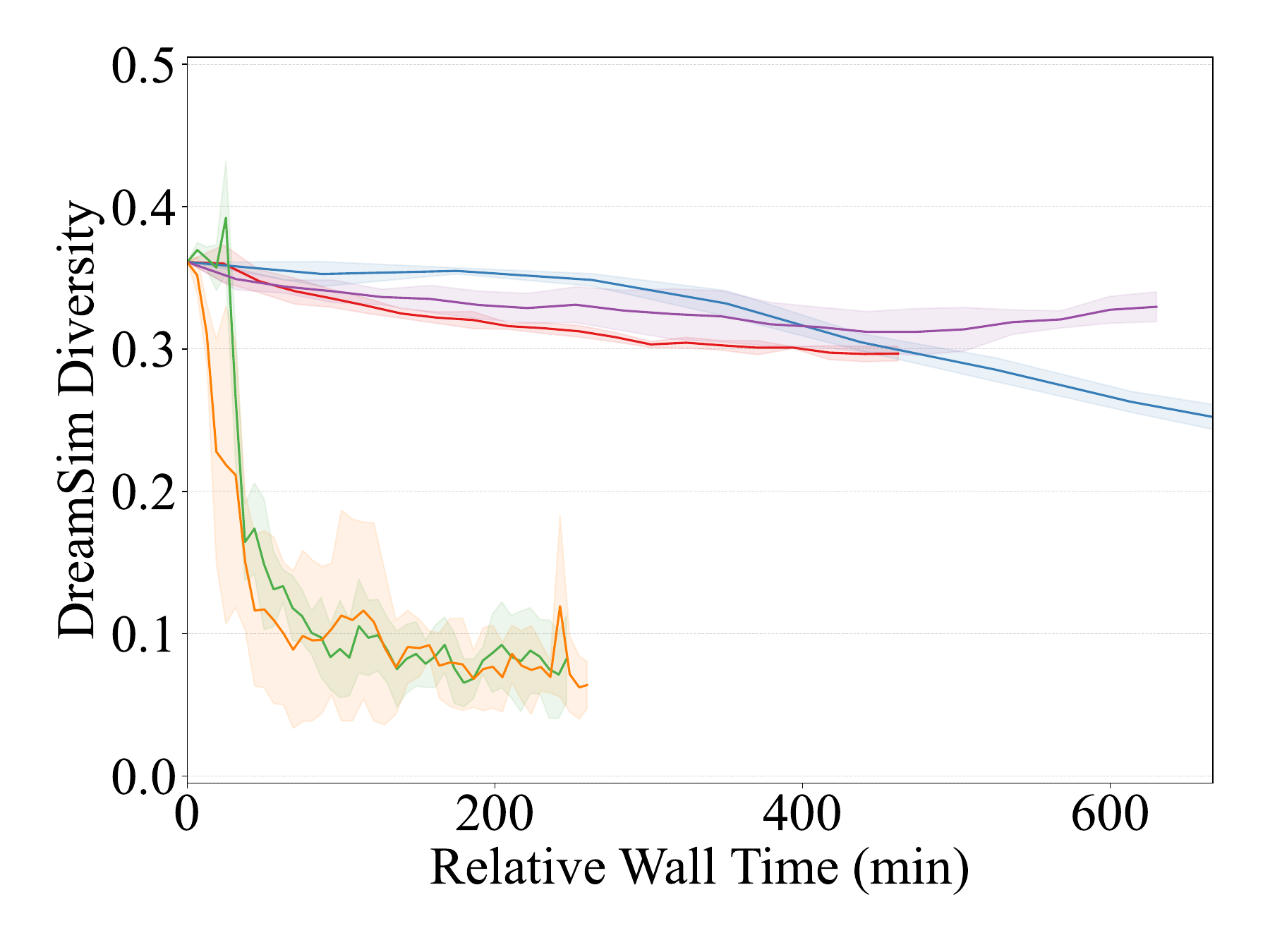}
    \hspace{-0.8em}
    \includegraphics[width=0.33\linewidth]{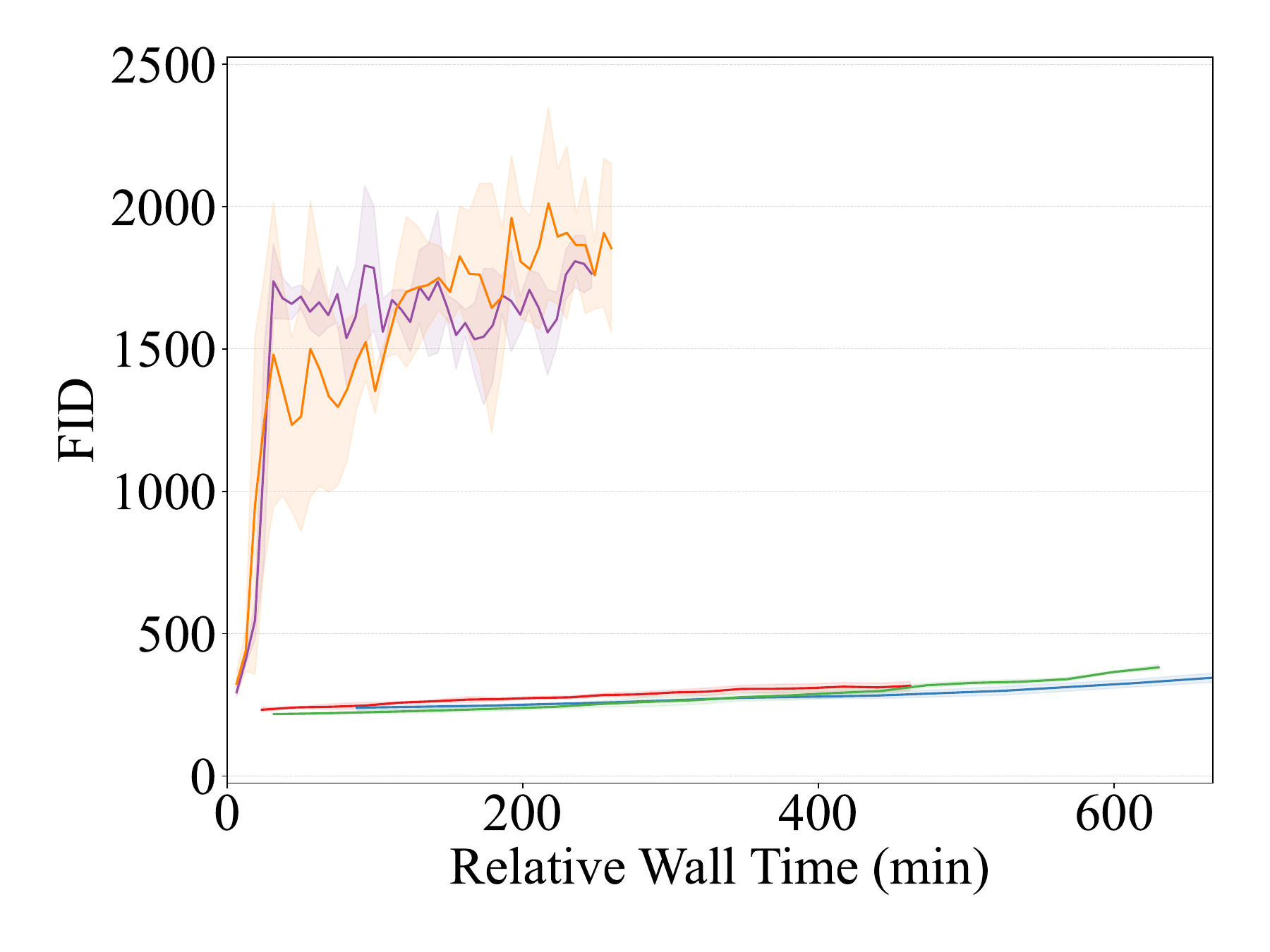}
    \caption{\footnotesize 
    Convergence curves of different metrics for different methods throughout the finetuning process on Aesthetic Score, with the $x$-axis being the relative wall time. All methods are benchmarked on a single node with 8 80GB-mem A100 GPUs.
    }
    \label{fig:evo_walltime}
    \vspace{-4mm}
\end{figure}

We further show the convergence speed measured in relative wall time on a single node with 8 80GB-mem A100 GPUs in Fig.~\ref{fig:evo_walltime}.

\newpage
\section{Results of ablation study}
\label{sec:appendix_ablation_figs}

\begin{figure}[h]
    \centering
    \vspace{-1pt}
    \adjustbox{valign=t, max width=0.98\linewidth}{%
        \includegraphics[width=0.5\linewidth]{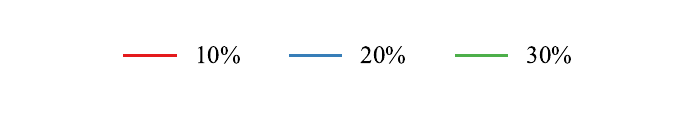}%
    }
    \vspace{-1.5em}

    \includegraphics[width=0.33\linewidth]{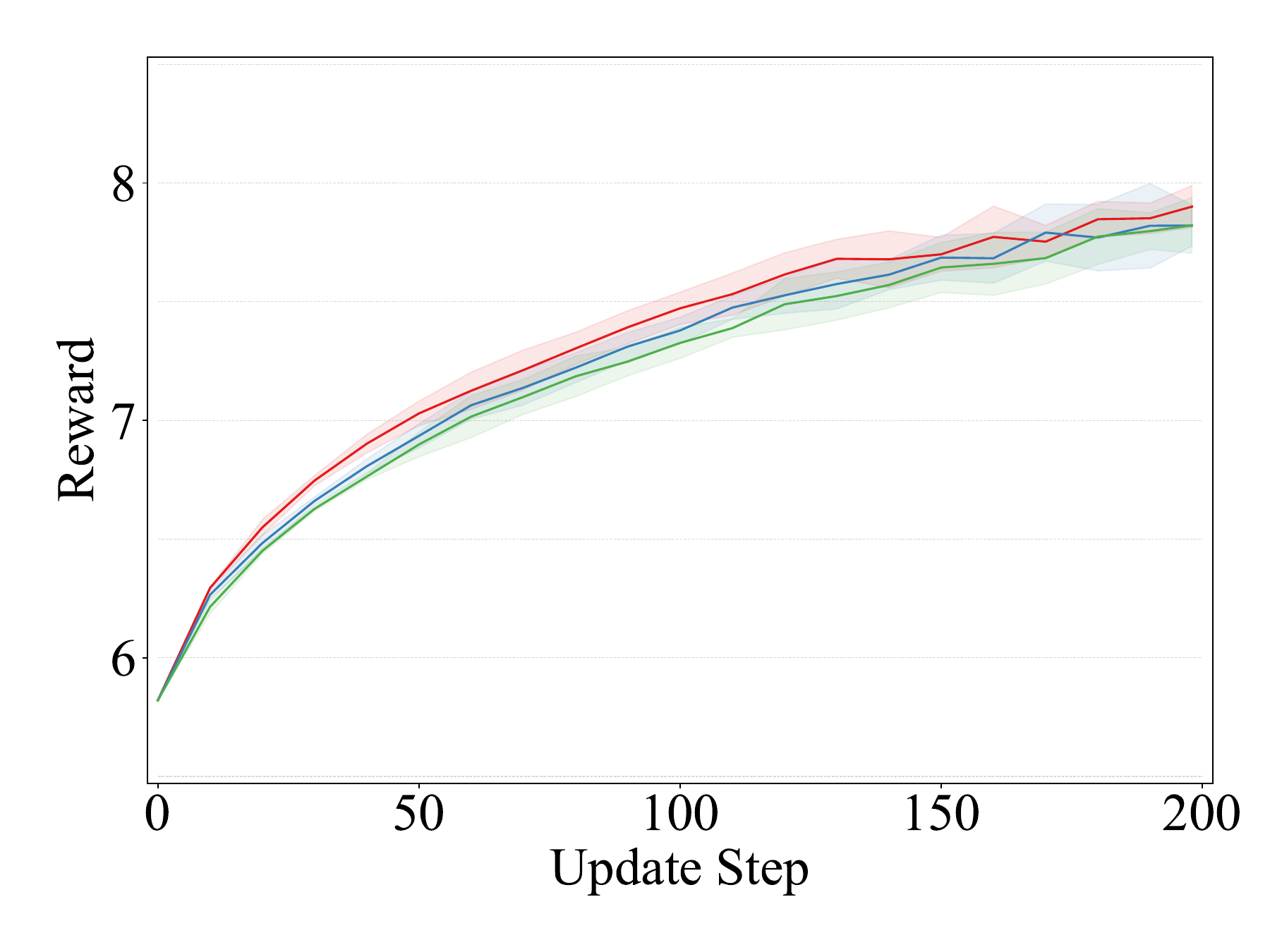}
    \hspace{-0.8em}
    \includegraphics[width=0.33\linewidth]{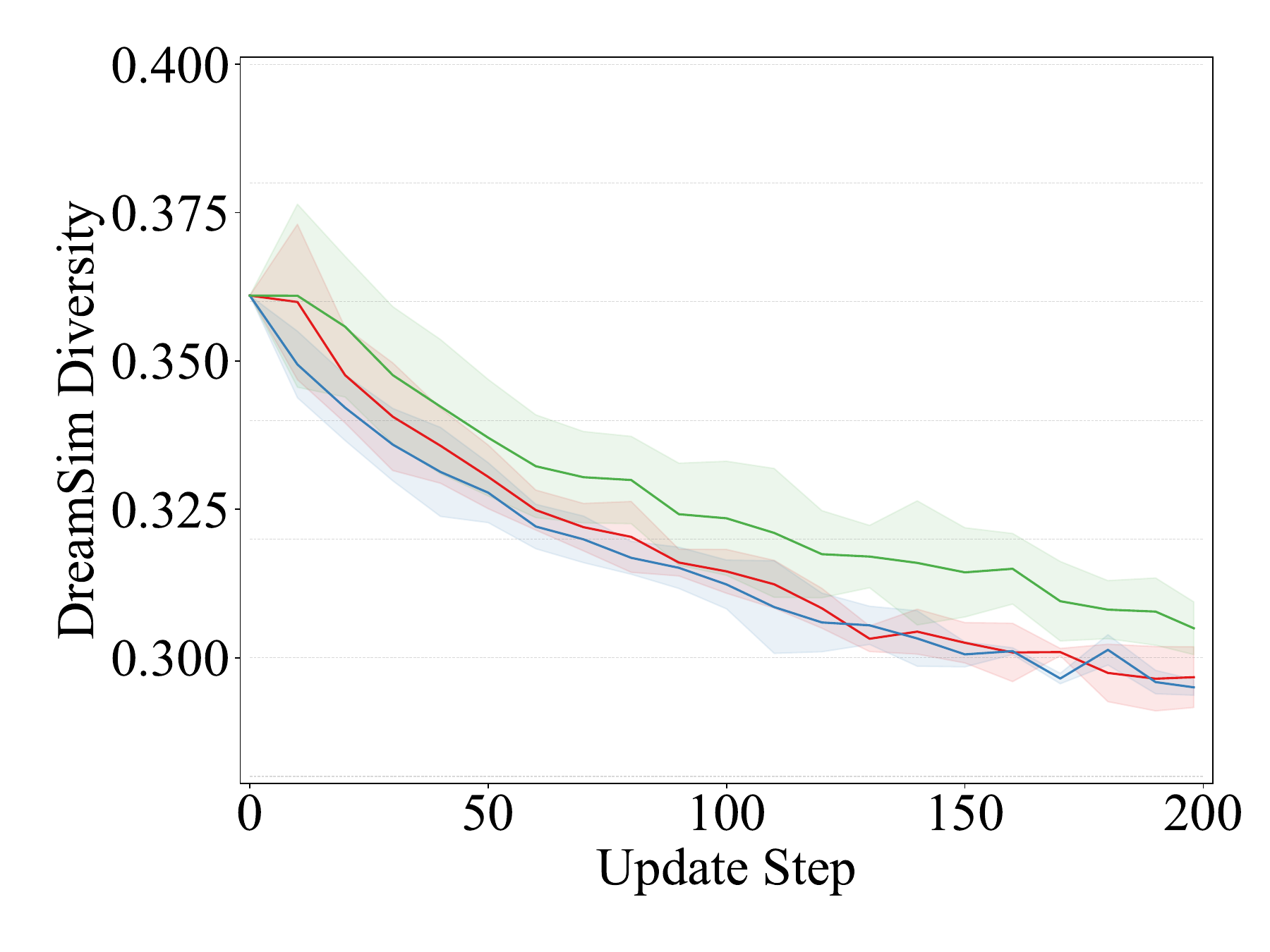}
    \hspace{-0.8em}
    \includegraphics[width=0.33\linewidth]{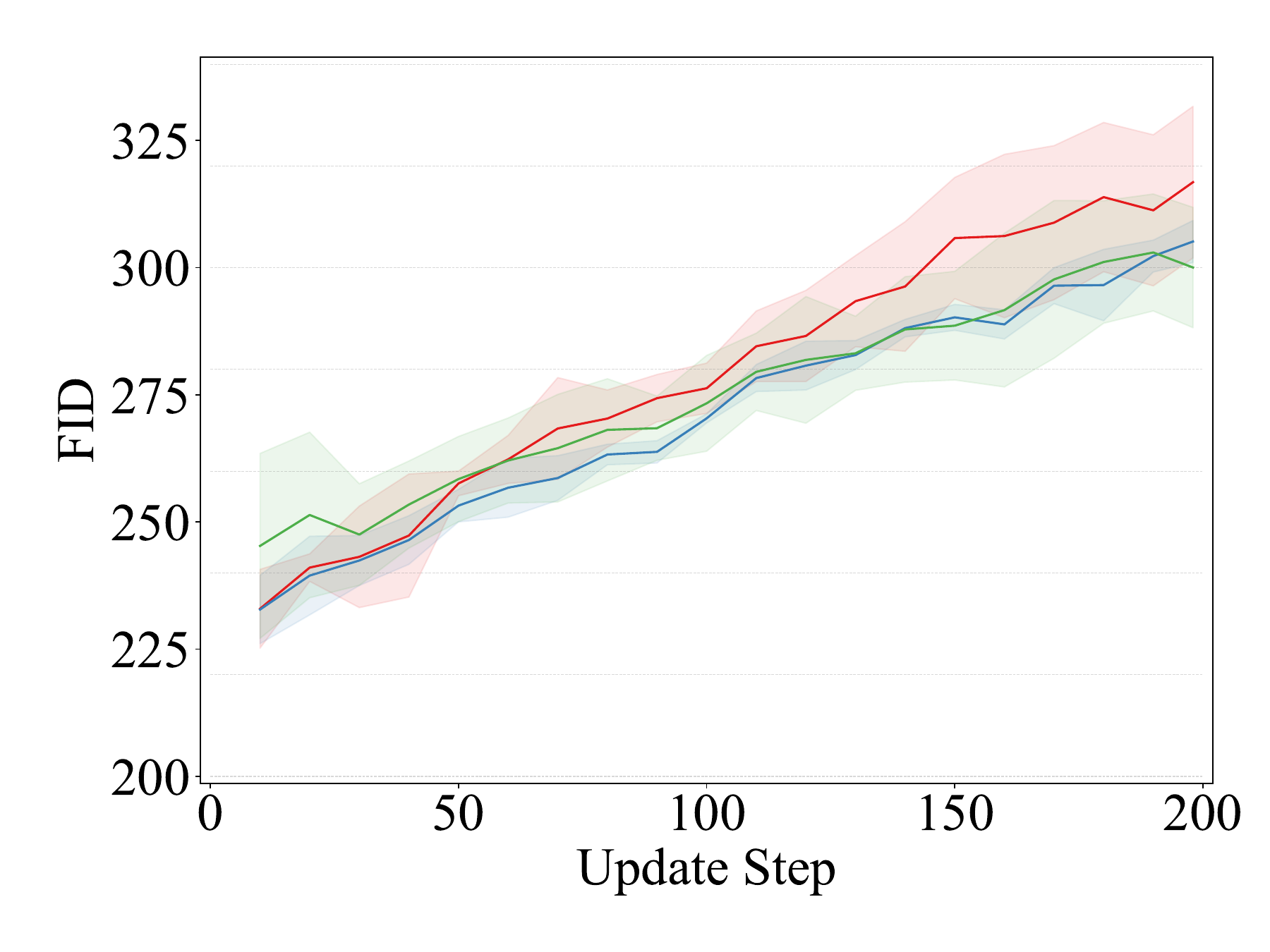}
    \caption{\footnotesize 
        Ablation study on the effect of subsampling rate on the collected trajectories for computing the \resgraddb loss.
    }
    \label{fig:ablation_subsampling}
\end{figure}

\begin{figure}[h]
    \vspace{-2mm}
    \centering
    \includegraphics[width=0.48\linewidth]{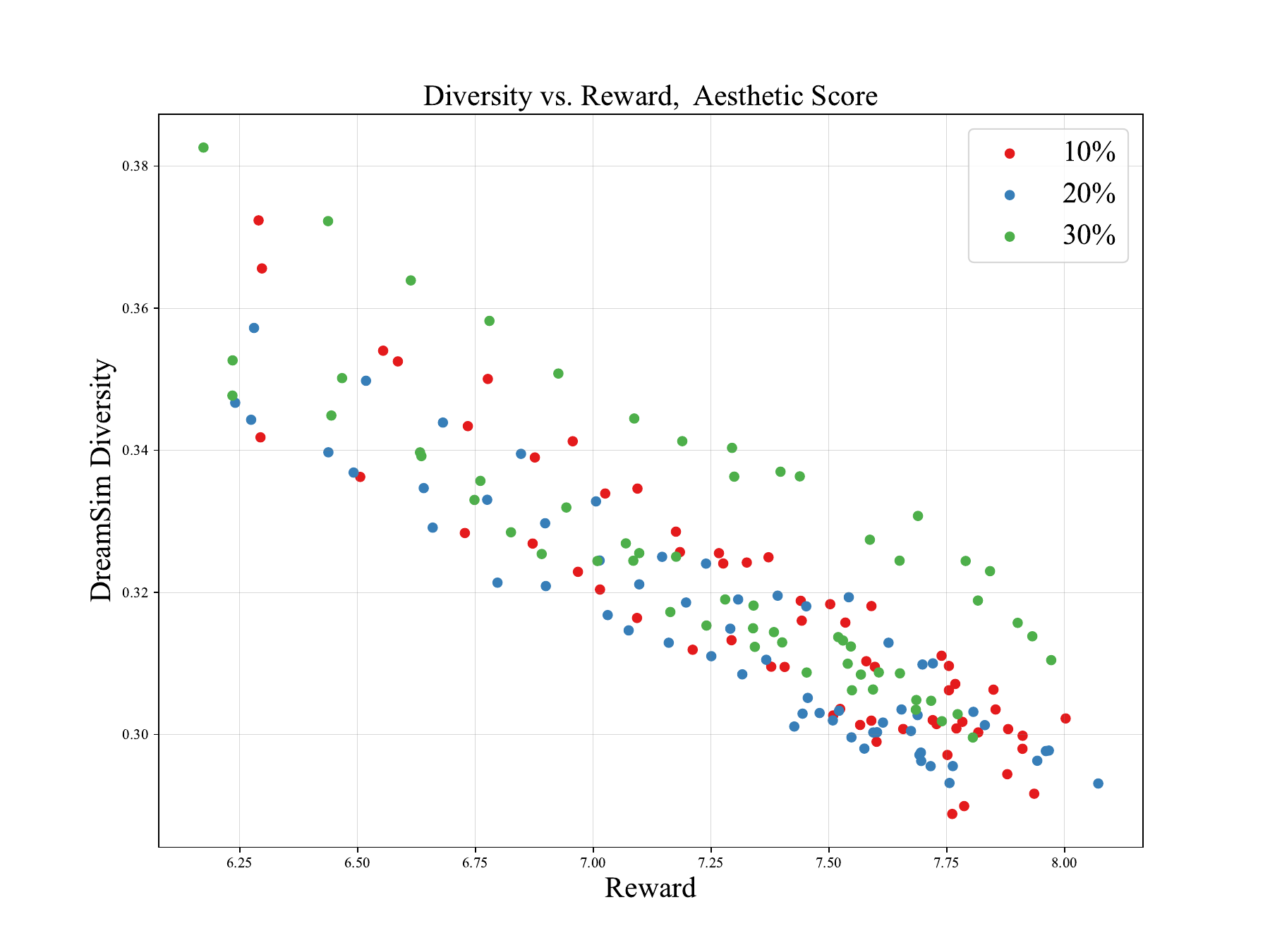}
    \hspace{-5mm}
    \includegraphics[width=0.48\linewidth]{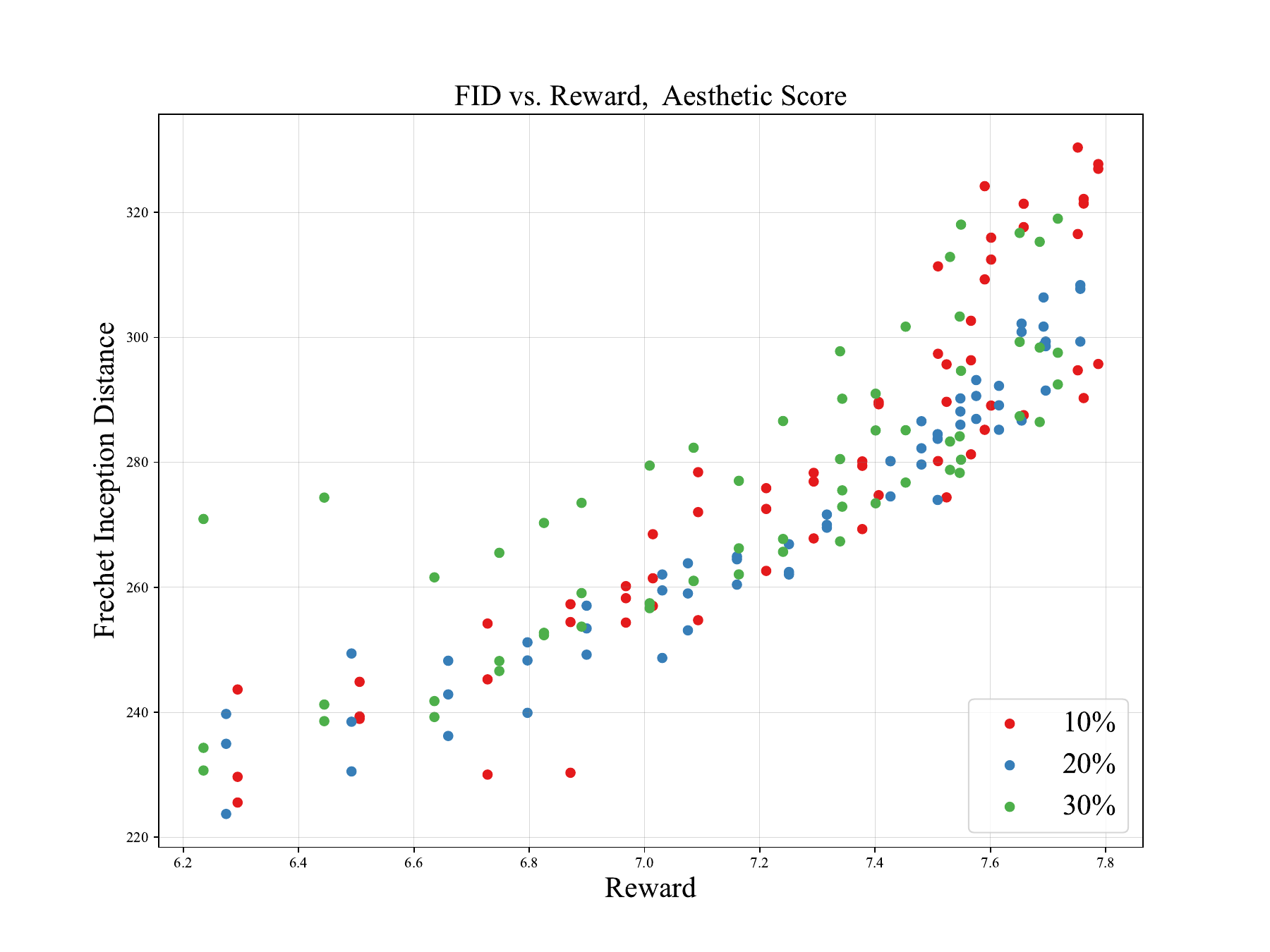}
    \vspace{-3mm}
    \caption{\footnotesize Pareto frontiers for reward, diversity and prior-preservation (measured by FID) of models trained with different subsampling rate. In expectation, higher subsampling rates seem to slightly help in increasing diversity.}
    \label{fig:pareto_subsampling}
    \vspace{-2mm}
\end{figure}

\begin{figure}[h]
    \centering
    \vspace{-1pt}
    \adjustbox{valign=t, max width=0.98\linewidth}{%
        \includegraphics{figs/aesthetic_evo_legend.pdf}%
    }
    \vspace{-1.5em}

    \includegraphics[width=0.33\linewidth]{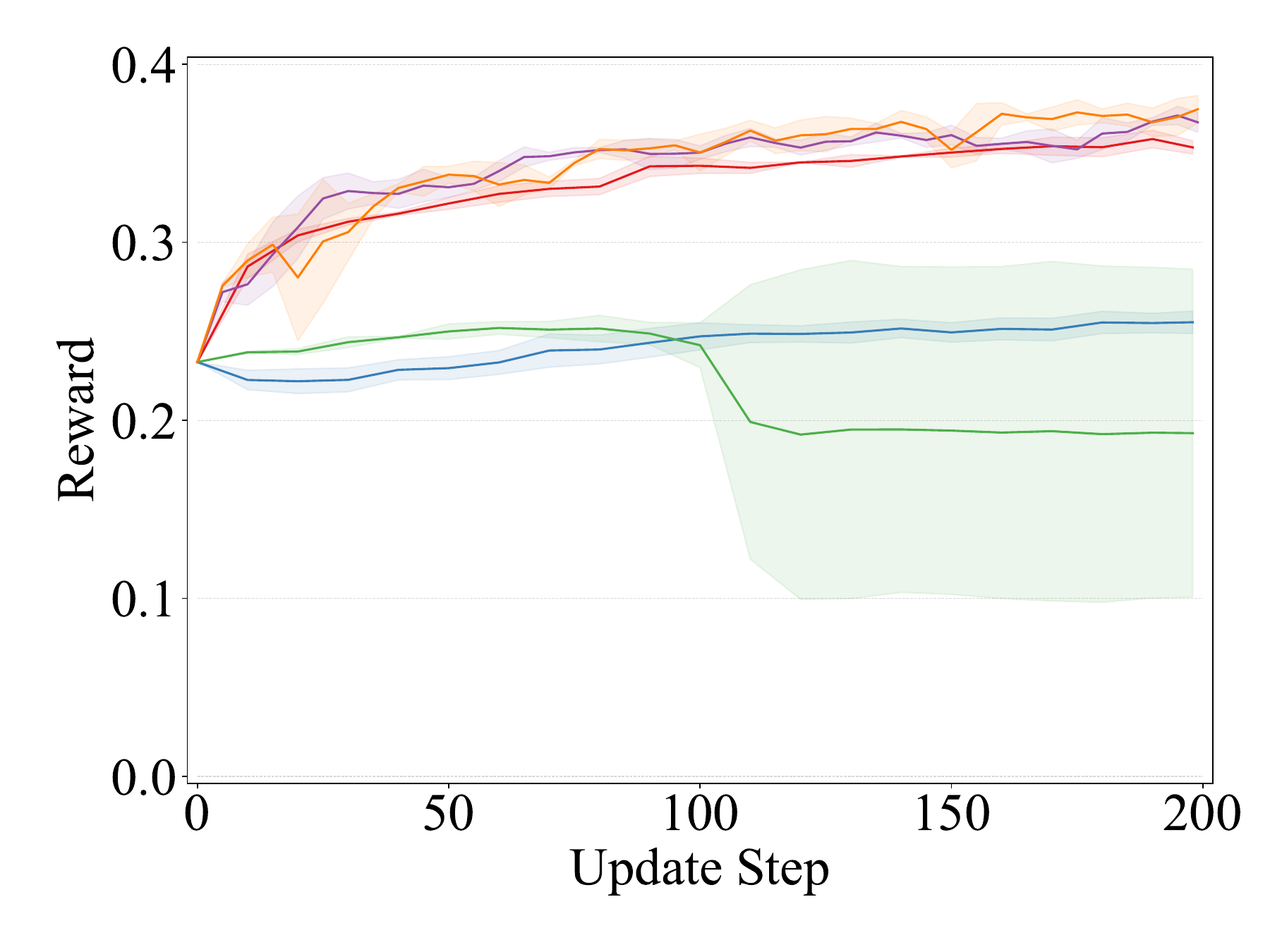}
    \hspace{-0.8em}
    \includegraphics[width=0.33\linewidth]{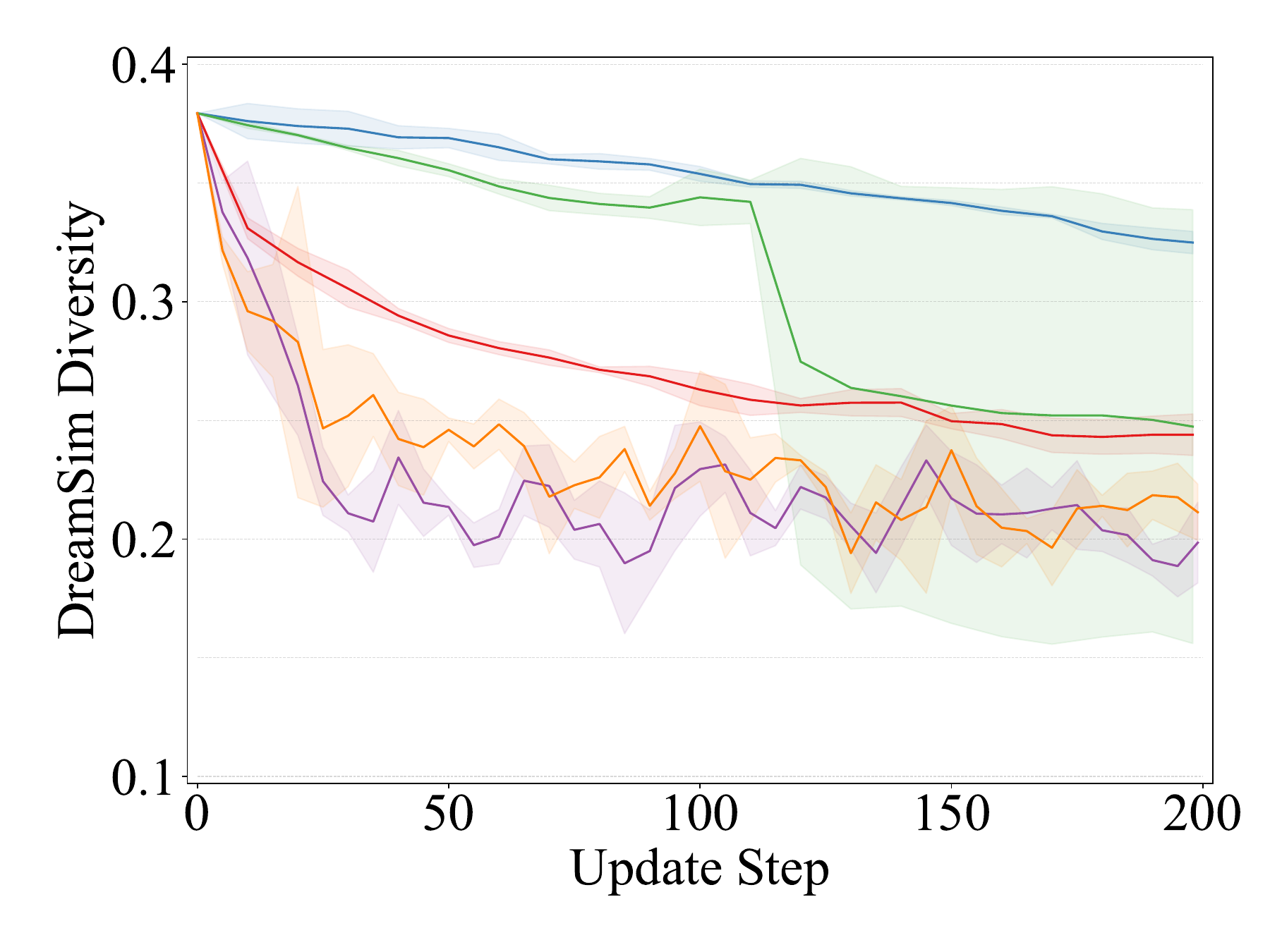}
    \hspace{-0.8em}
    \includegraphics[width=0.33\linewidth]{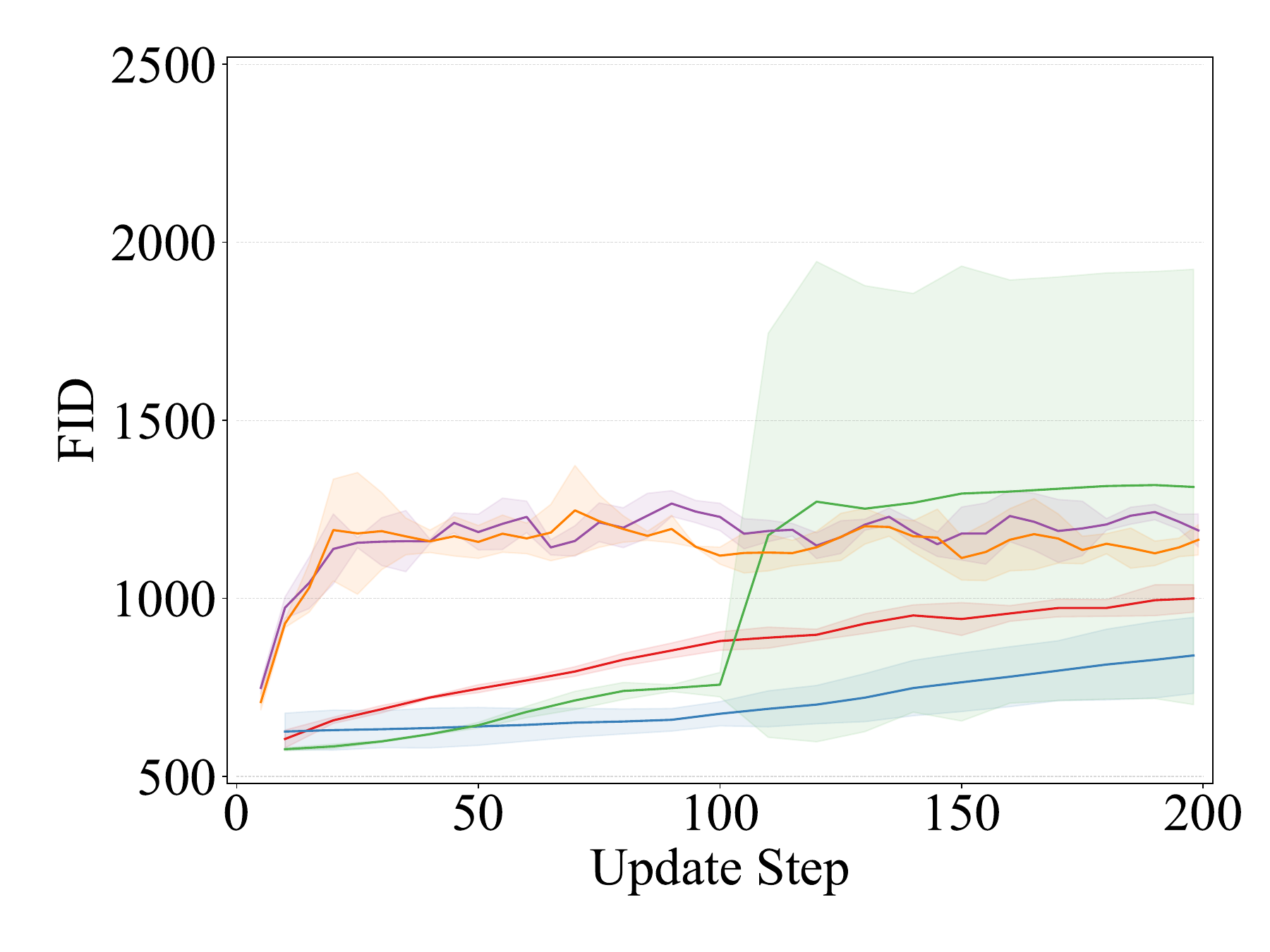}
    \caption{\footnotesize 
    Convergence curve of metrics of different methods throughout the finetuning process on the HPSv2 reward model.
    }
    \label{fig:hpsv2_evo}
\end{figure}

\begin{figure}[h]
    \vspace{-2mm}
    \centering
    \includegraphics[width=0.48\linewidth]{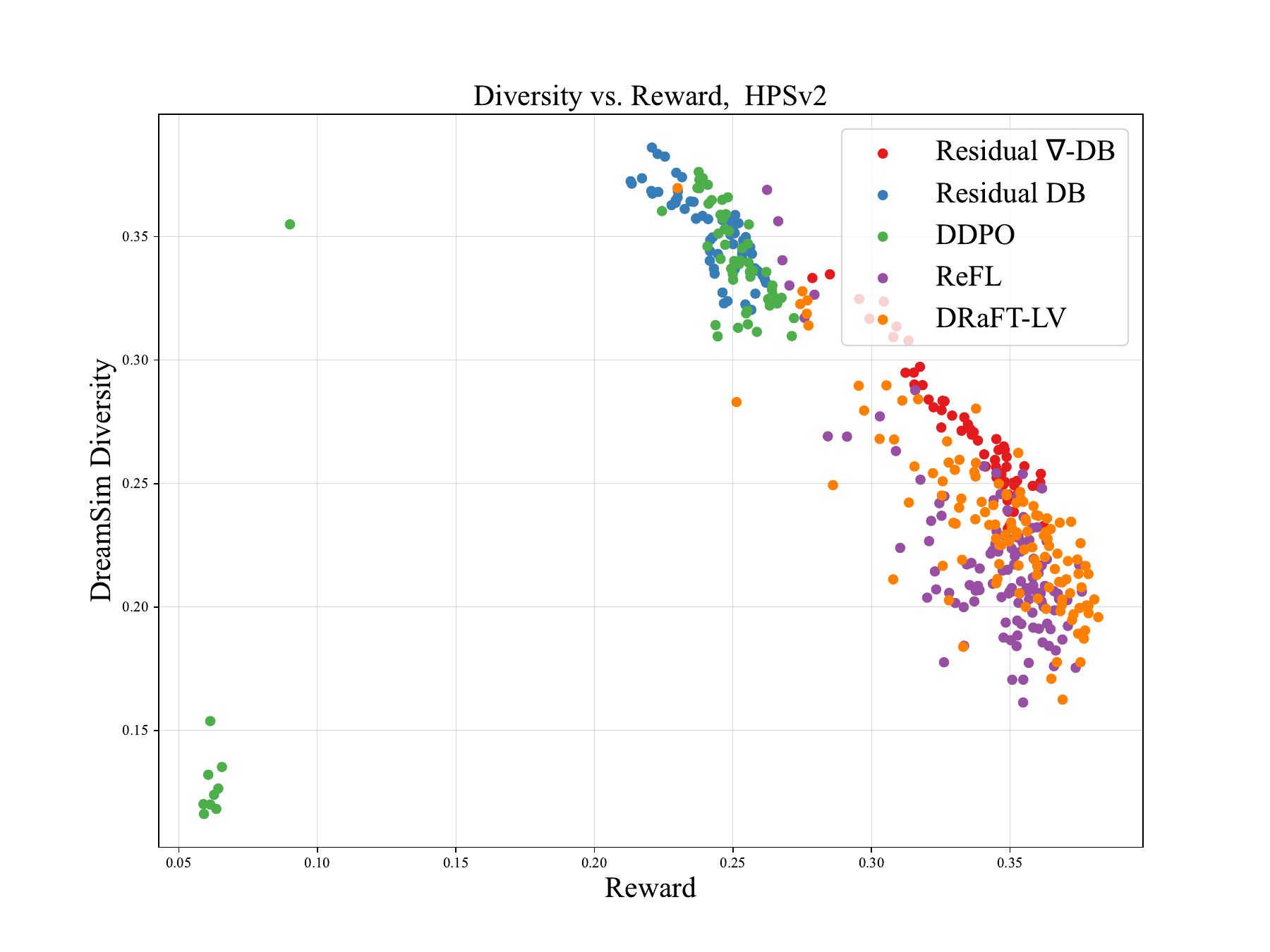}
    \hspace{-5mm}
    \includegraphics[width=0.48\linewidth]{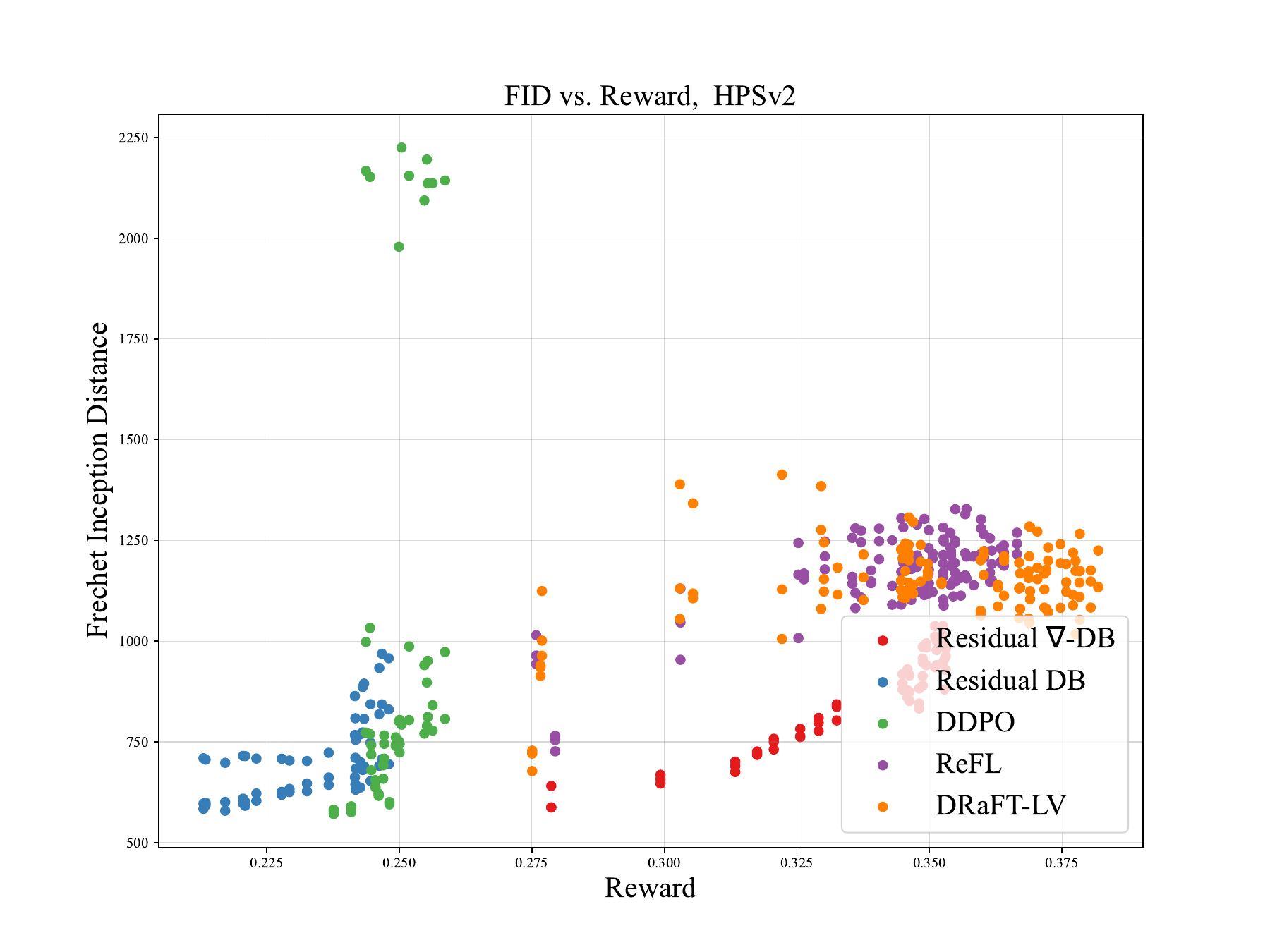}
    \vspace{-3mm}
    \caption{\footnotesize Pareto frontiers for reward, diversity and prior-preservation (measured by FID) on HPSv2.}
    \label{fig:pareto_hpsv2}
    \vspace{-2mm}
\end{figure}

\begin{figure}[h]
    \centering
    \vspace{-1pt}
    \adjustbox{valign=t, max width=0.98\linewidth}{%
        \includegraphics{figs/aesthetic_evo_legend.pdf}%
    }
    \vspace{-1.5em}

    \includegraphics[width=0.33\linewidth]{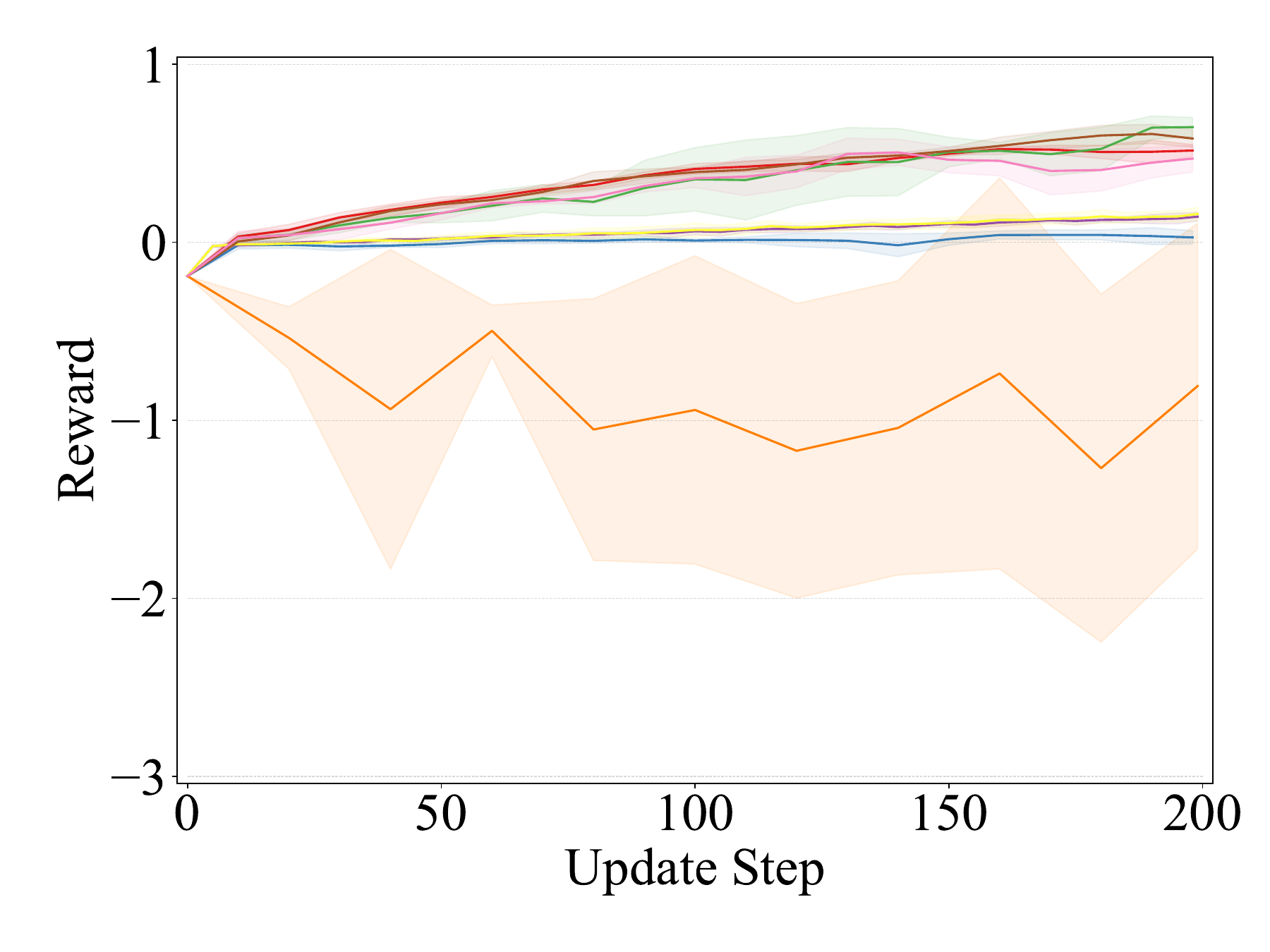}
    \hspace{-0.8em}
    \includegraphics[width=0.33\linewidth]{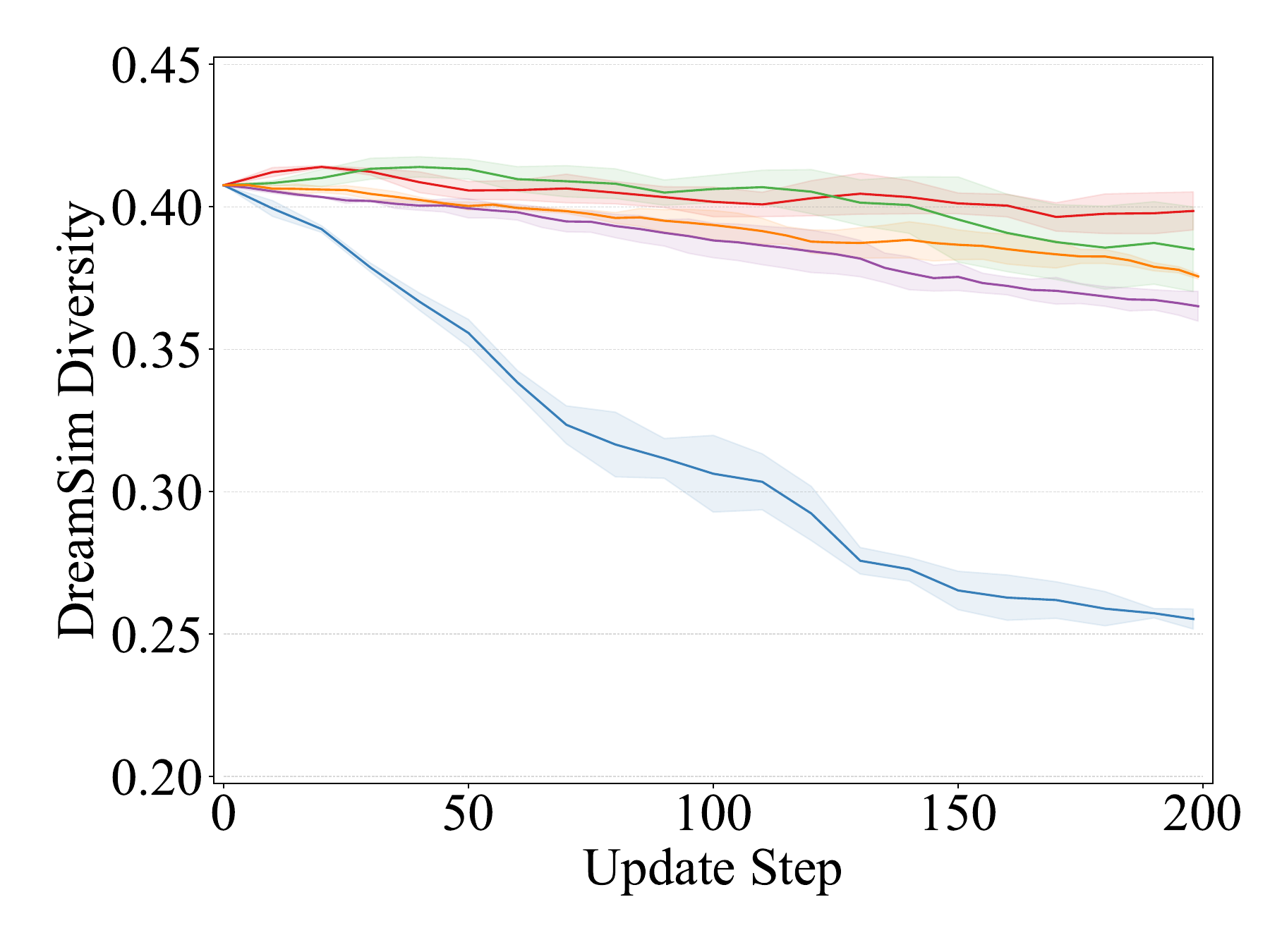}
    \hspace{-0.8em}
    \includegraphics[width=0.33\linewidth]{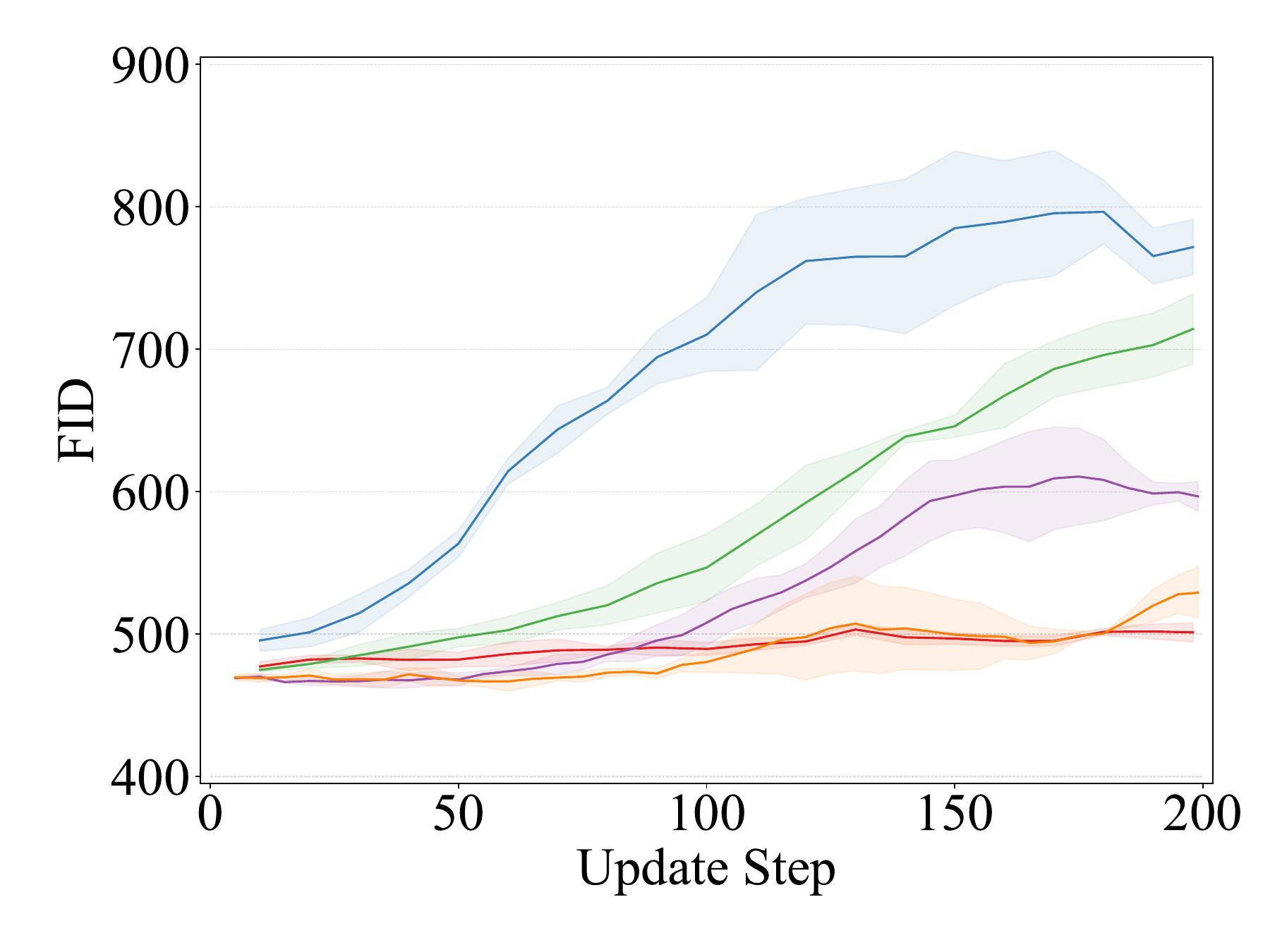}
    \caption{\footnotesize 
    Convergence curve of metrics of different methods throughout the finetuning process on the ImageReward reward model.
    }
    \label{fig:imagereward_evo}
\end{figure}

\begin{figure}[h]
    \vspace{-2mm}
    \centering
    \includegraphics[width=0.48\linewidth]{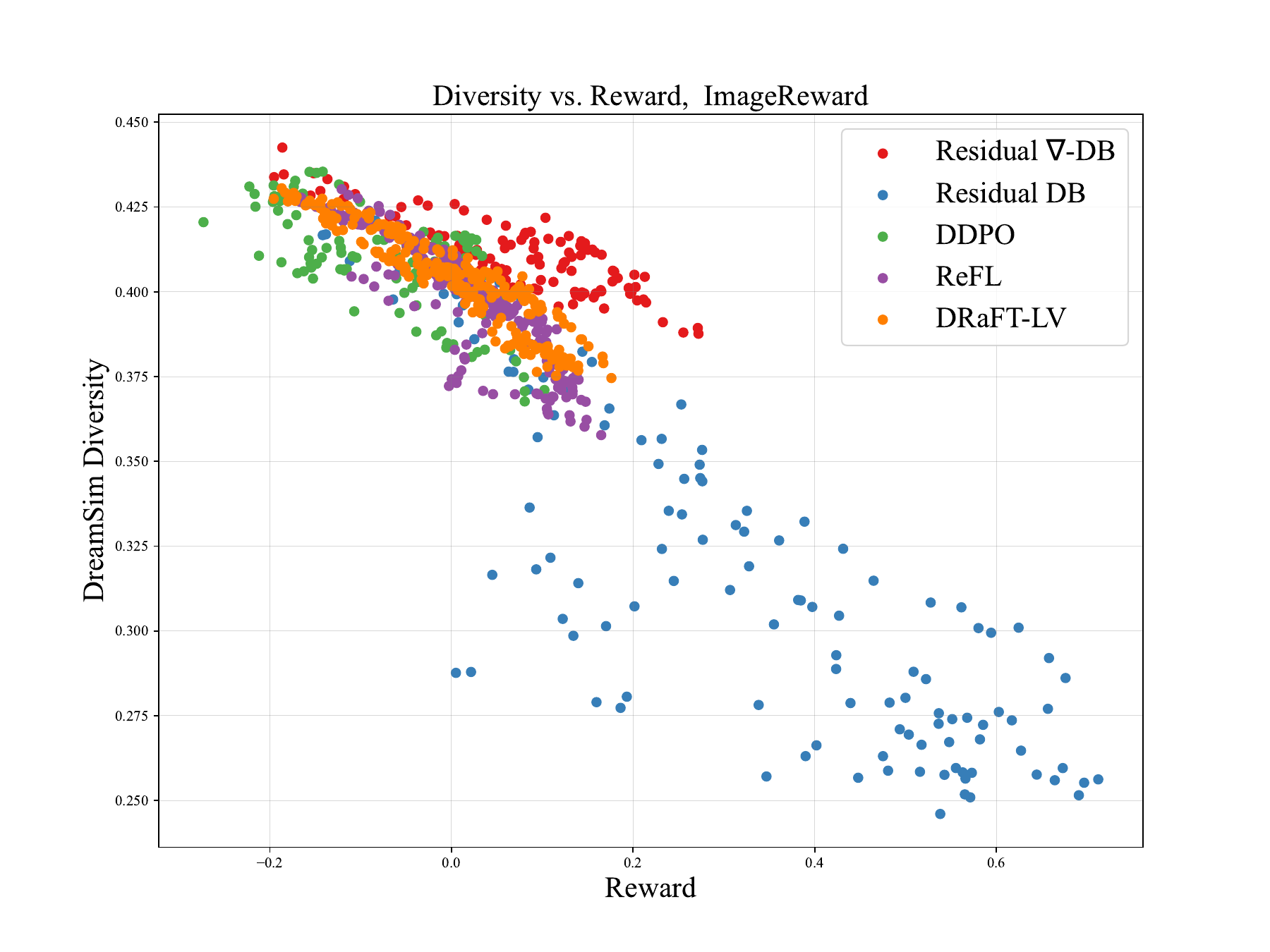}
    \hspace{-5mm}
    \includegraphics[width=0.48\linewidth]{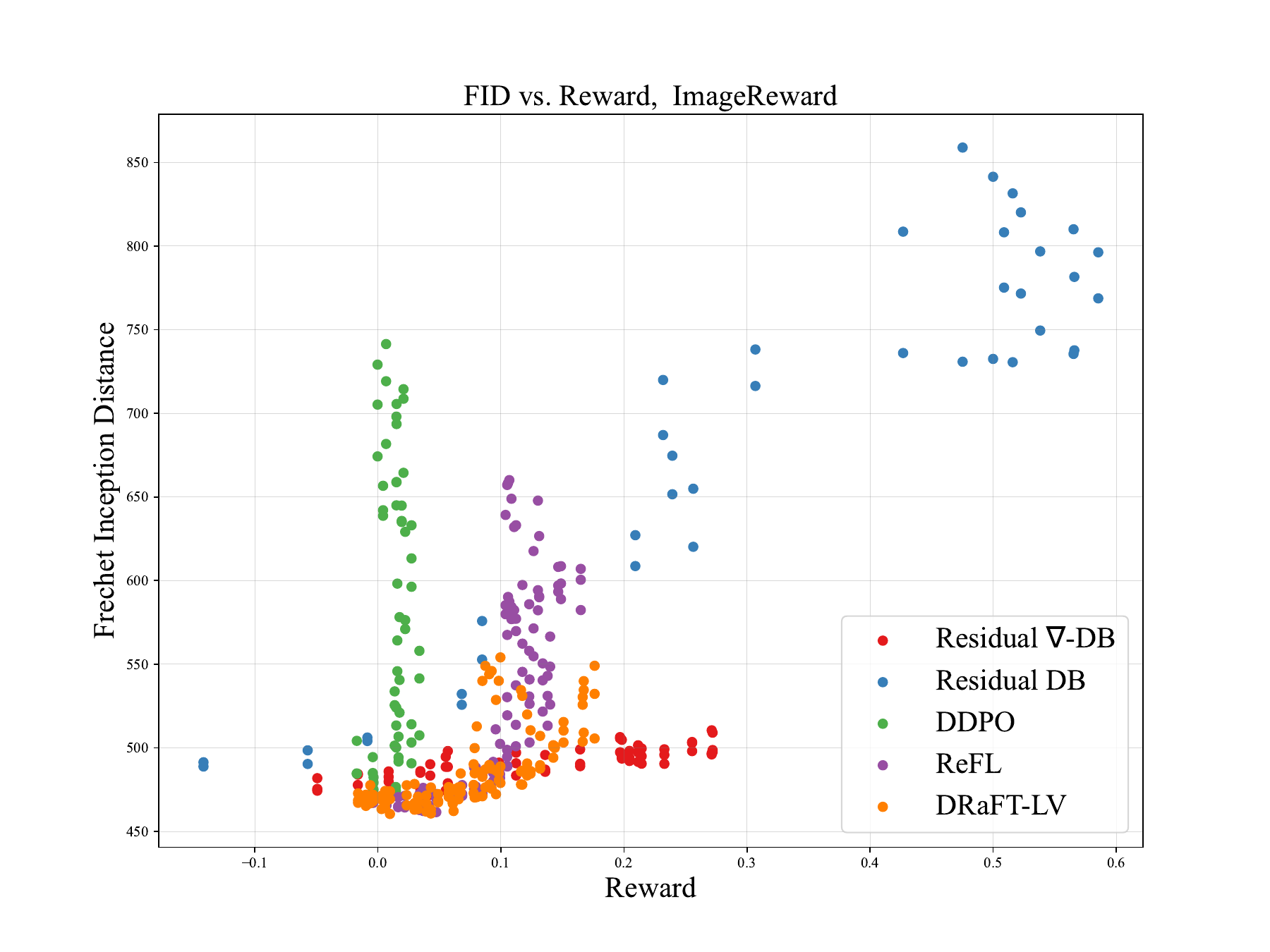}
    \vspace{-3mm}
    \caption{\footnotesize Pareto frontiers for reward, diversity and prior-preservation (measured by FID) on ImageReward.}
    \label{fig:pareto_imagereward}
    \vspace{-2mm}
\end{figure}

\begin{figure}[h]
    \centering
    \vspace{-1pt}
    \adjustbox{valign=t, max width=0.98\linewidth}{%
        \includegraphics[width=0.5\linewidth]{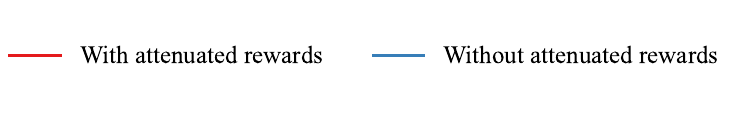}%
    }
    \vspace{-1.5em}

    \includegraphics[width=0.33\linewidth]{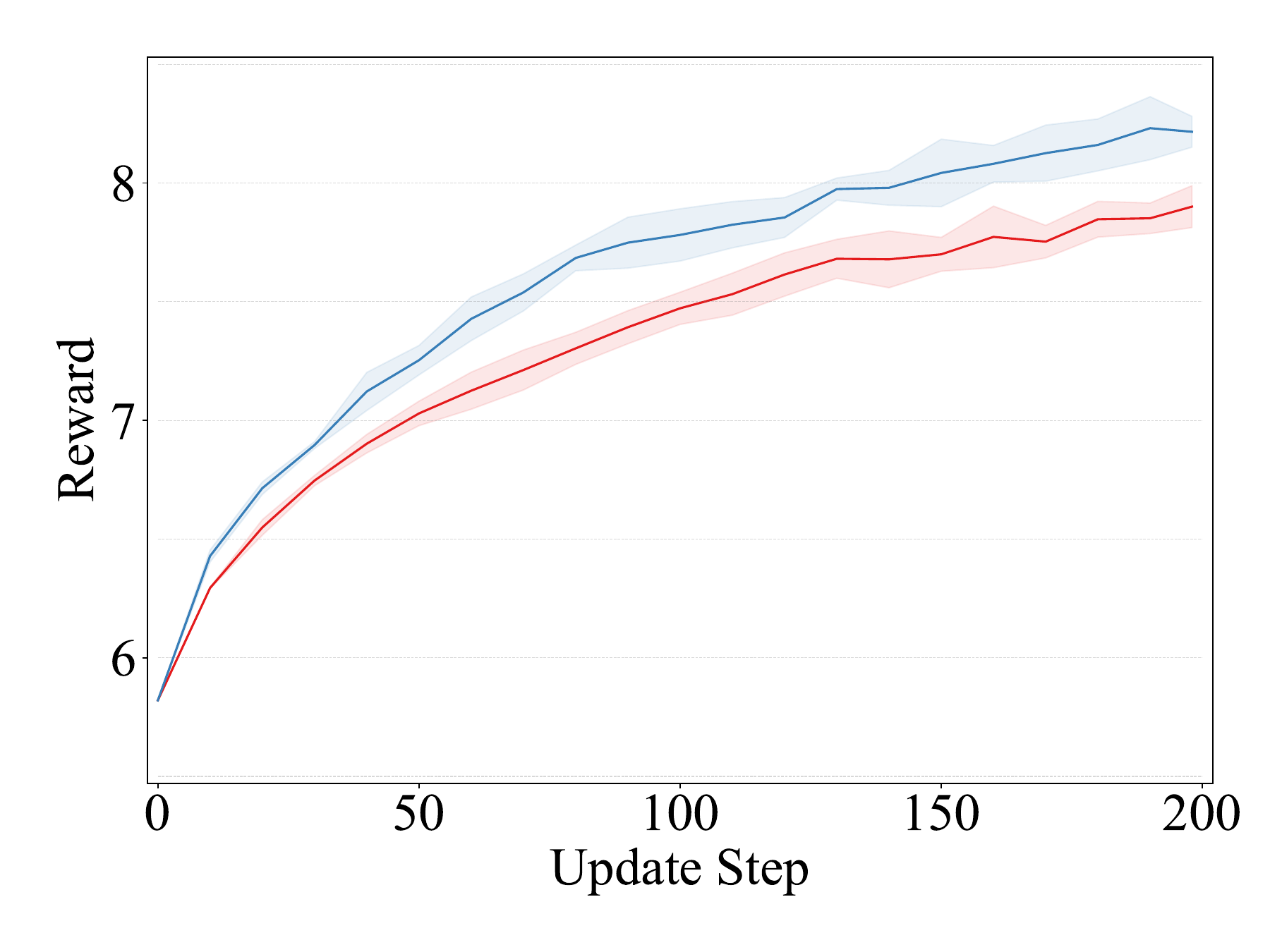}
    \hspace{-0.8em}
    \includegraphics[width=0.33\linewidth]{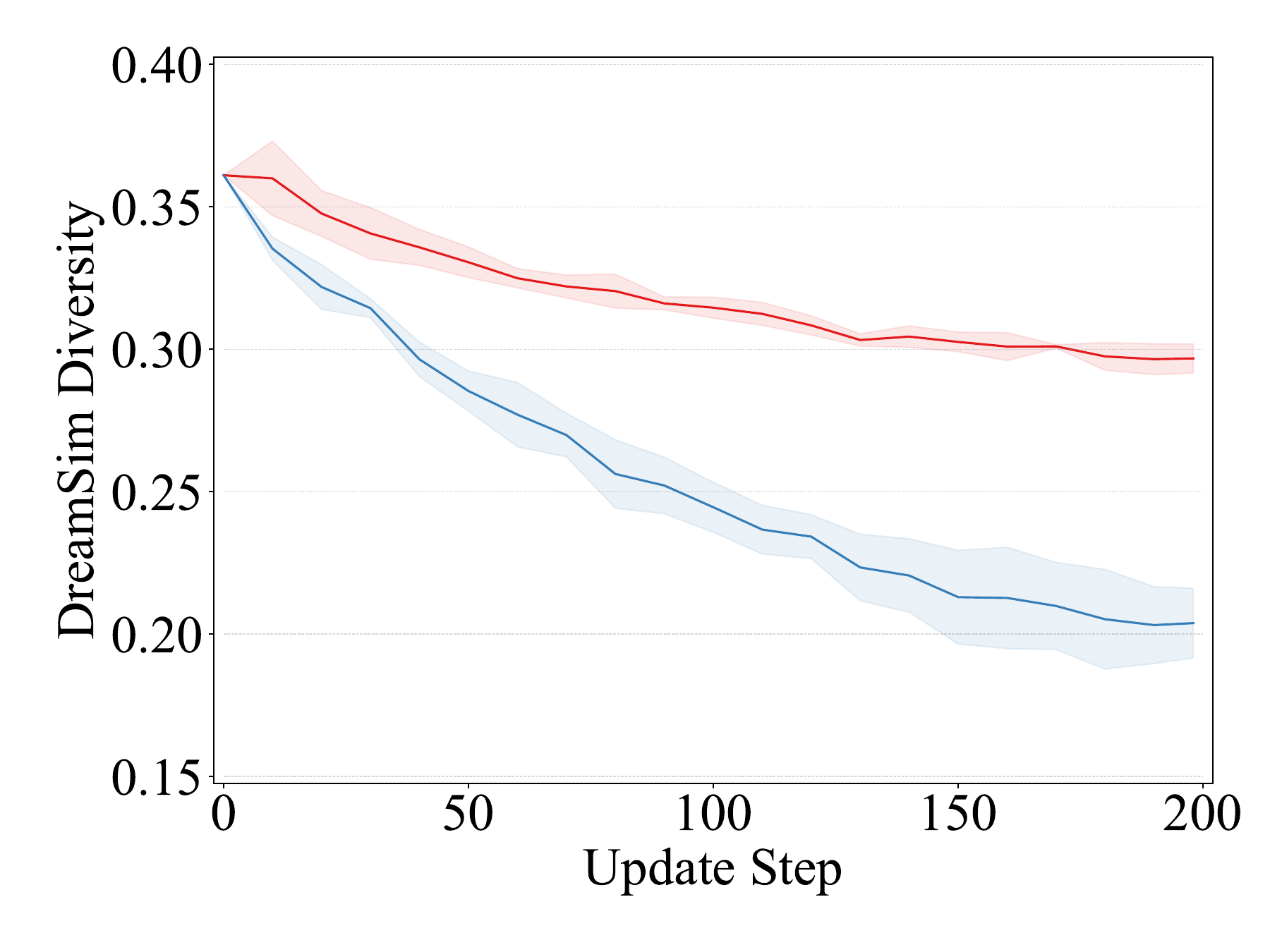}
    \hspace{-0.8em}
    \includegraphics[width=0.33\linewidth]{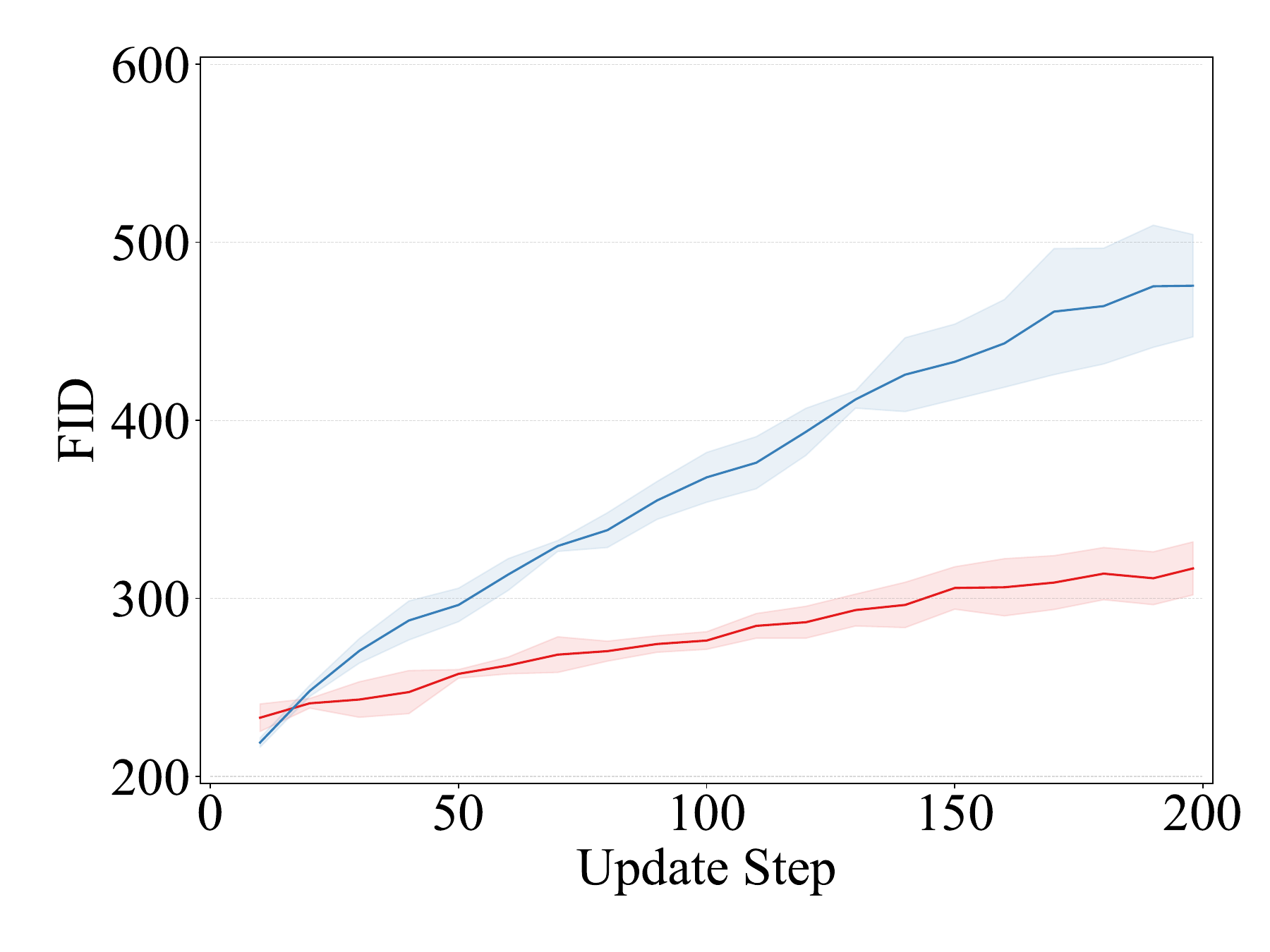}
    \caption{\footnotesize 
        Convergence curve of metrics of different methods throughout the finetuning process on Aesthetic Score with time-dependent attenuation of predicted rewards. Both models are trained with $\beta=10000$. With decayed predicted rewards, the convergence speed is slower but due to less aggressive prediction on reward signal, the model with reward attenuation achives better diversity and prior-preserving results.
    }
    \label{fig:aesthetic_attenuation}
\end{figure}

\begin{figure}[h]
    \vspace{-4mm}
    \centering
    \includegraphics[width=0.48\linewidth]{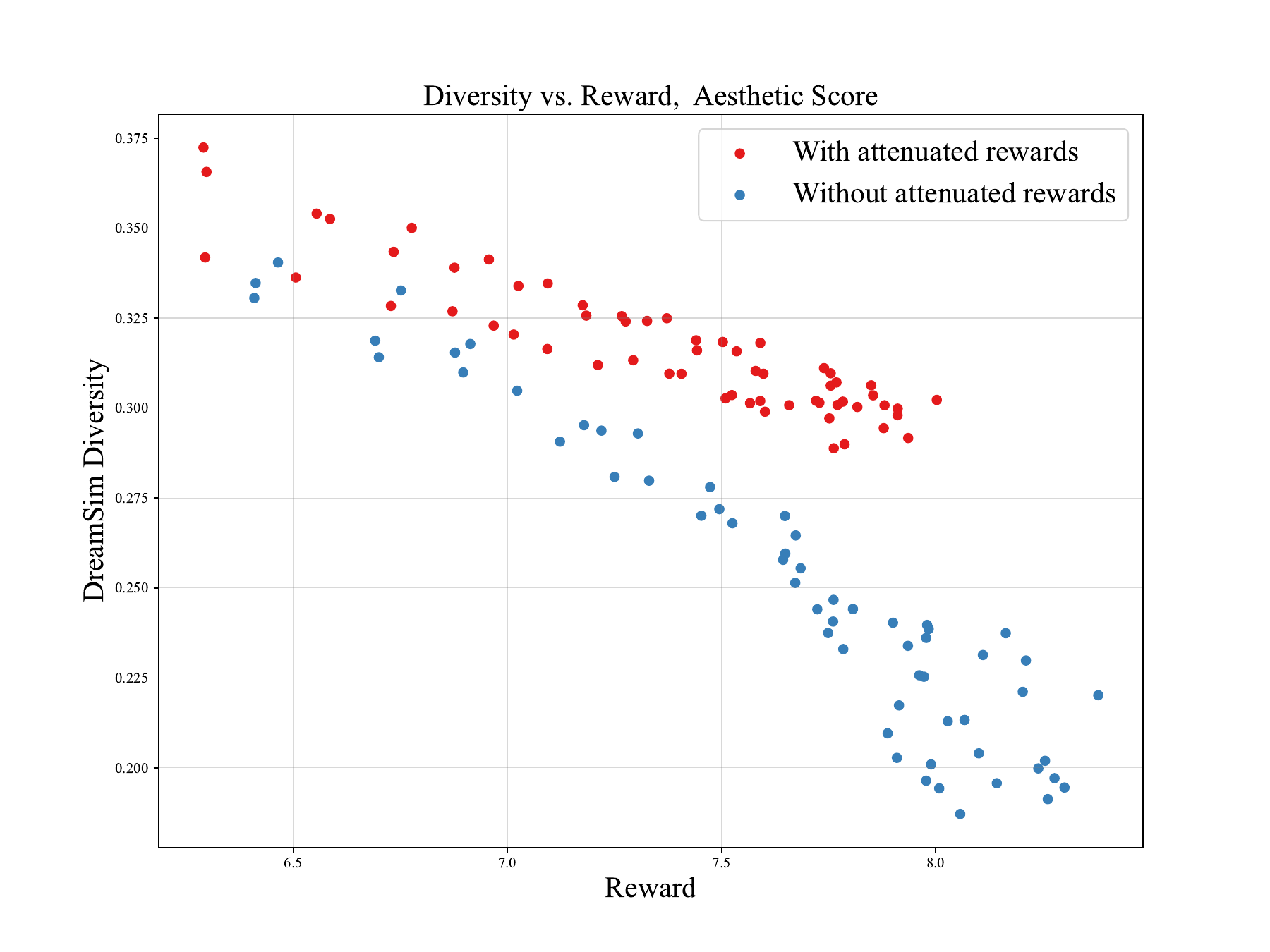}
    \hspace{-5mm}
    \includegraphics[width=0.48\linewidth]{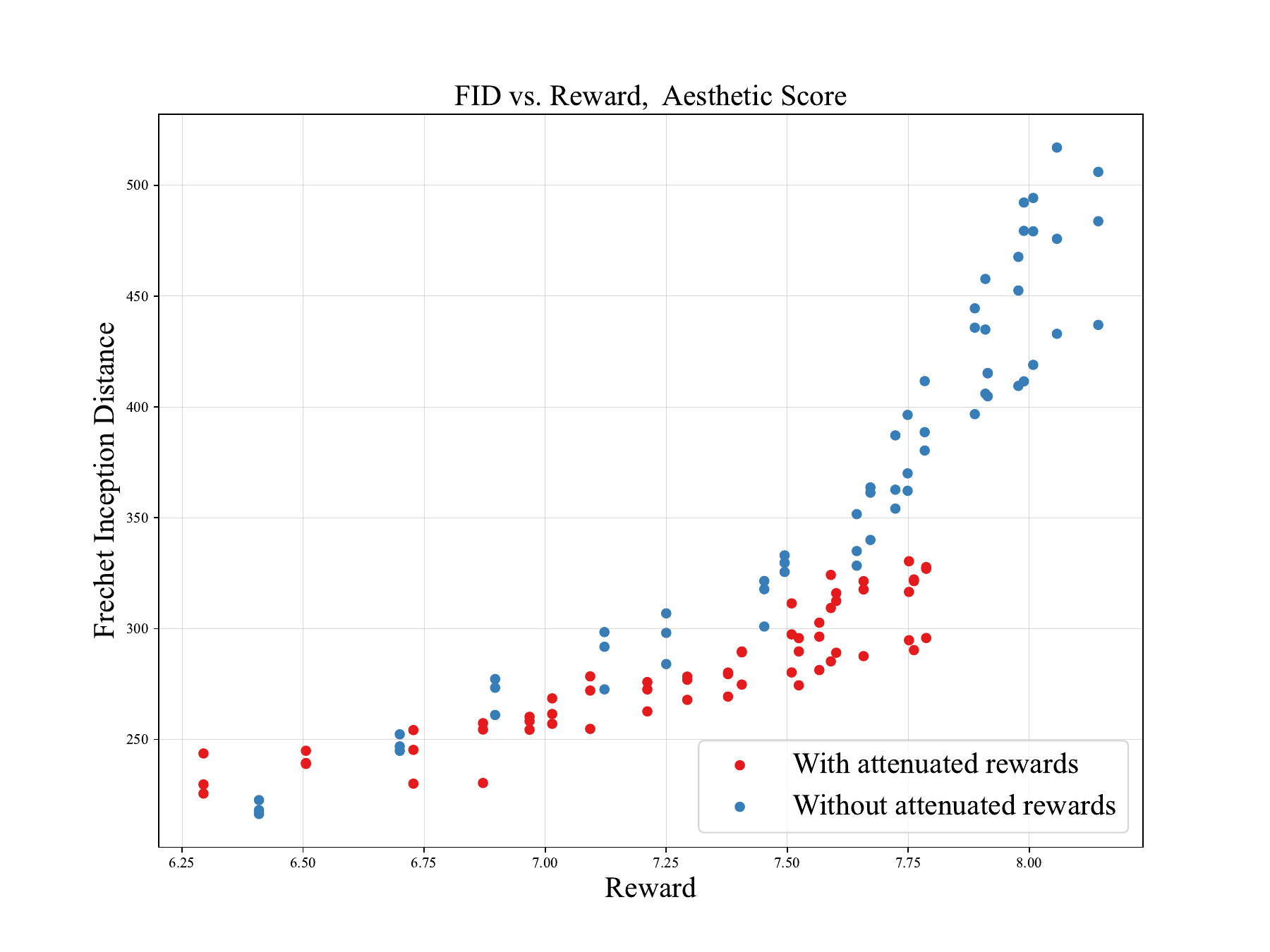}
    \vspace{-3mm}
    \caption{\footnotesize Pareto frontiers for reward, diversity and prior-preservation (measured by FID) on Aesthetic Score with time-dependent scaling of predicted rewards.}
    \label{fig:pareto_attenuation}
    \vspace{-2mm}
\end{figure}

\begin{figure}[h]
    \centering
    \vspace{-1pt}
    \adjustbox{valign=t, max width=0.98\linewidth}{%
        \includegraphics[width=0.5\linewidth]{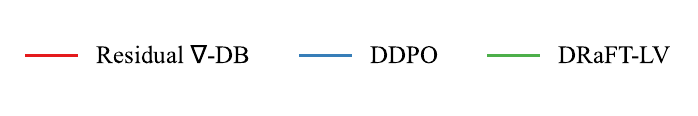}%
    }
    \vspace{-1.5em}

    \includegraphics[width=0.33\linewidth]{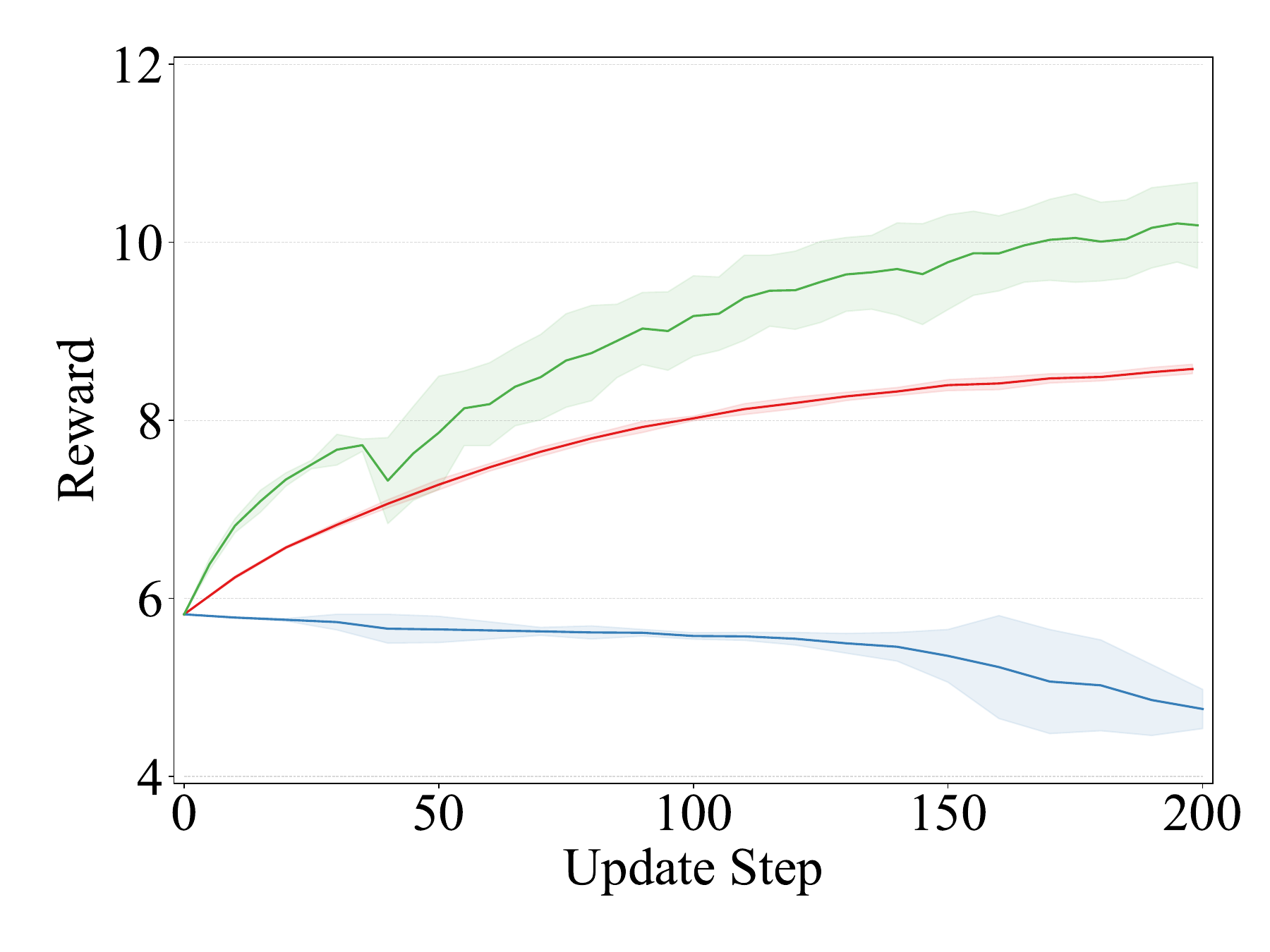}
    \hspace{-0.8em}
    \includegraphics[width=0.33\linewidth]{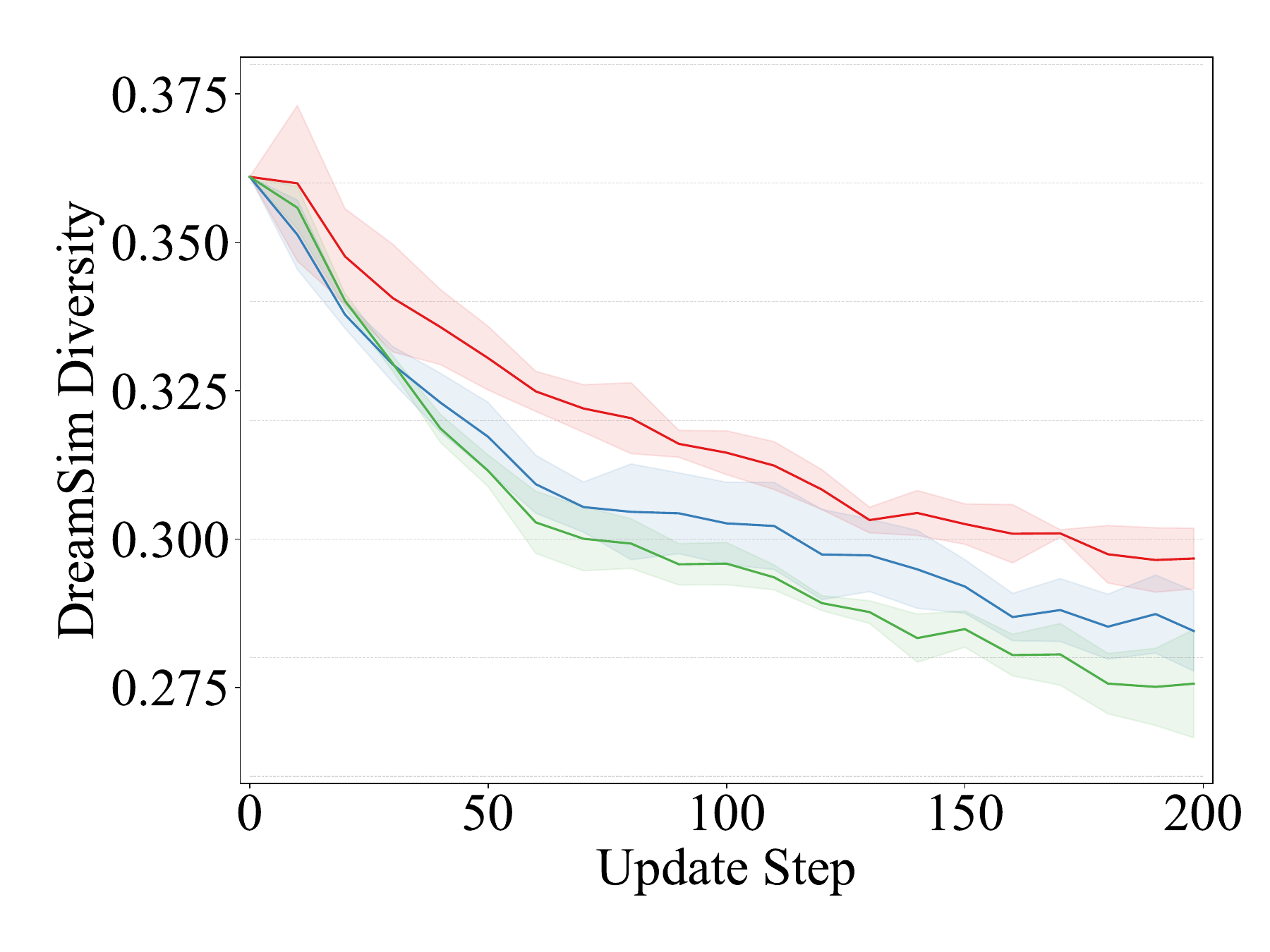}
    \hspace{-0.8em}
    \includegraphics[width=0.33\linewidth]{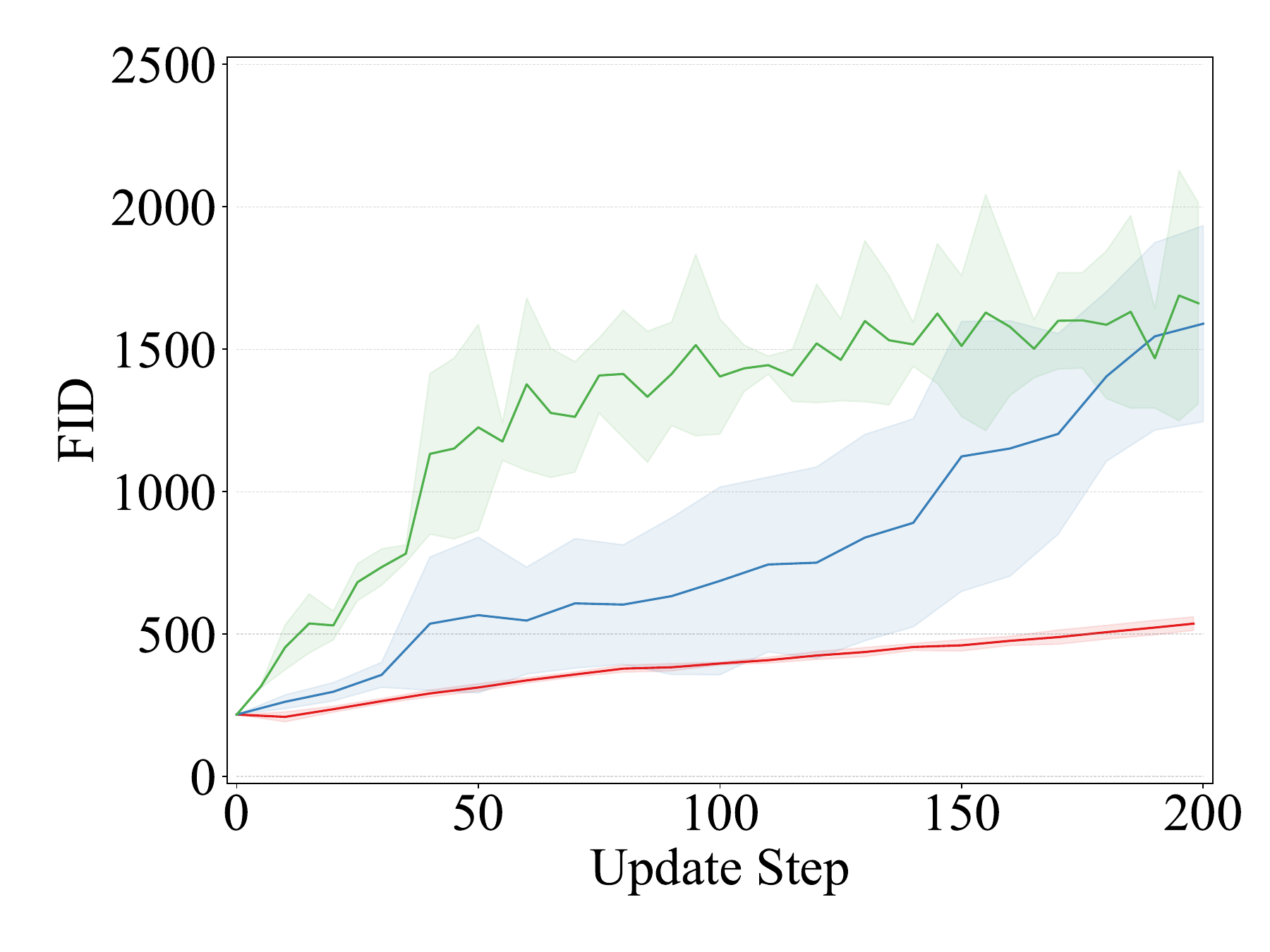}
    \caption{\footnotesize 
        Convergence curve of metrics of different methods throughout the finetuning process on Aesthetic Score with the MDP constructed by SDE-DPM-Solver++ (with 20 inference steps).
    }
    \label{fig:aesthetic_dpm}

\end{figure}

\begin{figure}[h]
    \vspace{-2mm}
    \centering
    \includegraphics[width=0.48\linewidth]{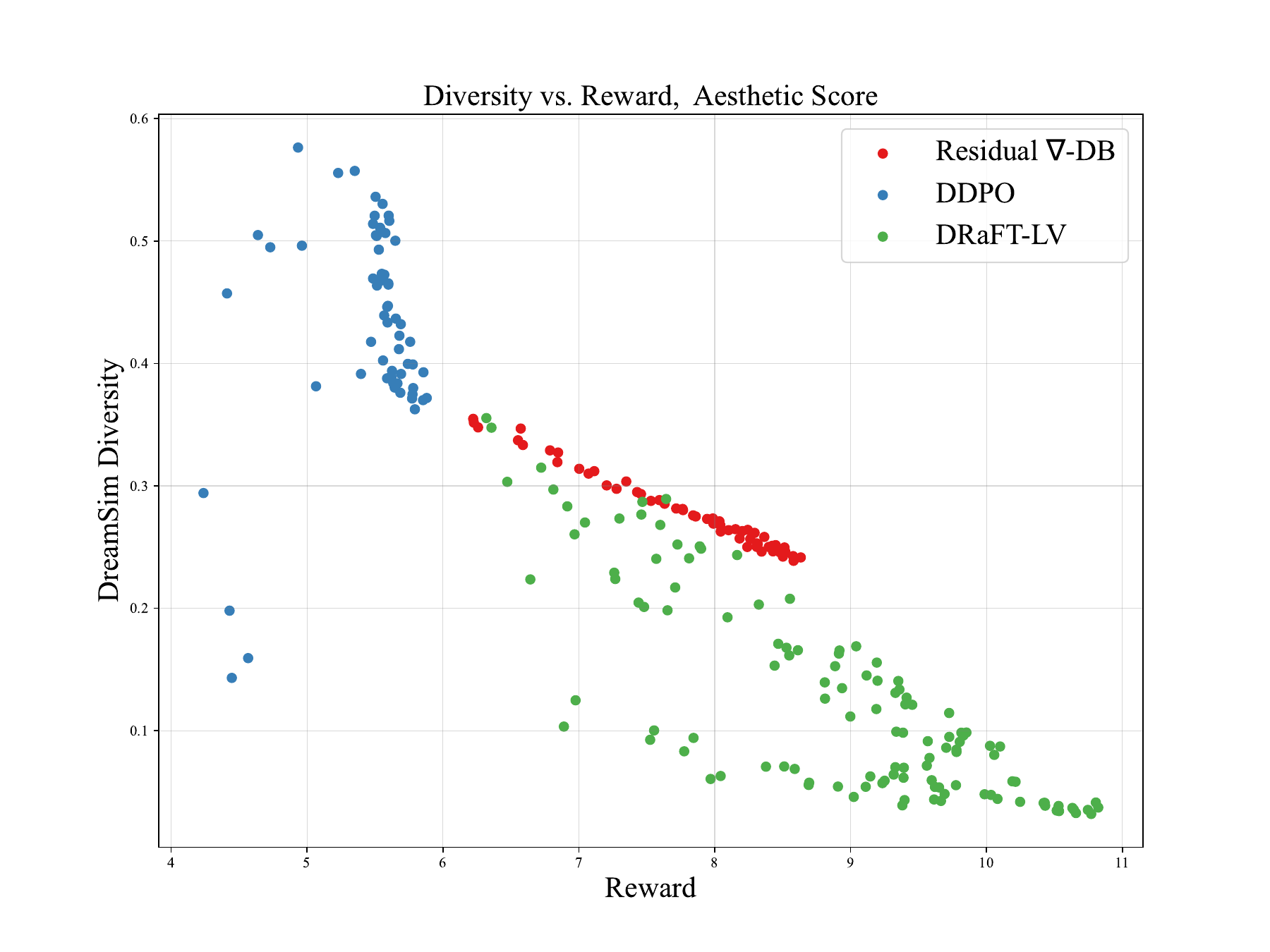}
    \hspace{-5mm}
    \includegraphics[width=0.48\linewidth]{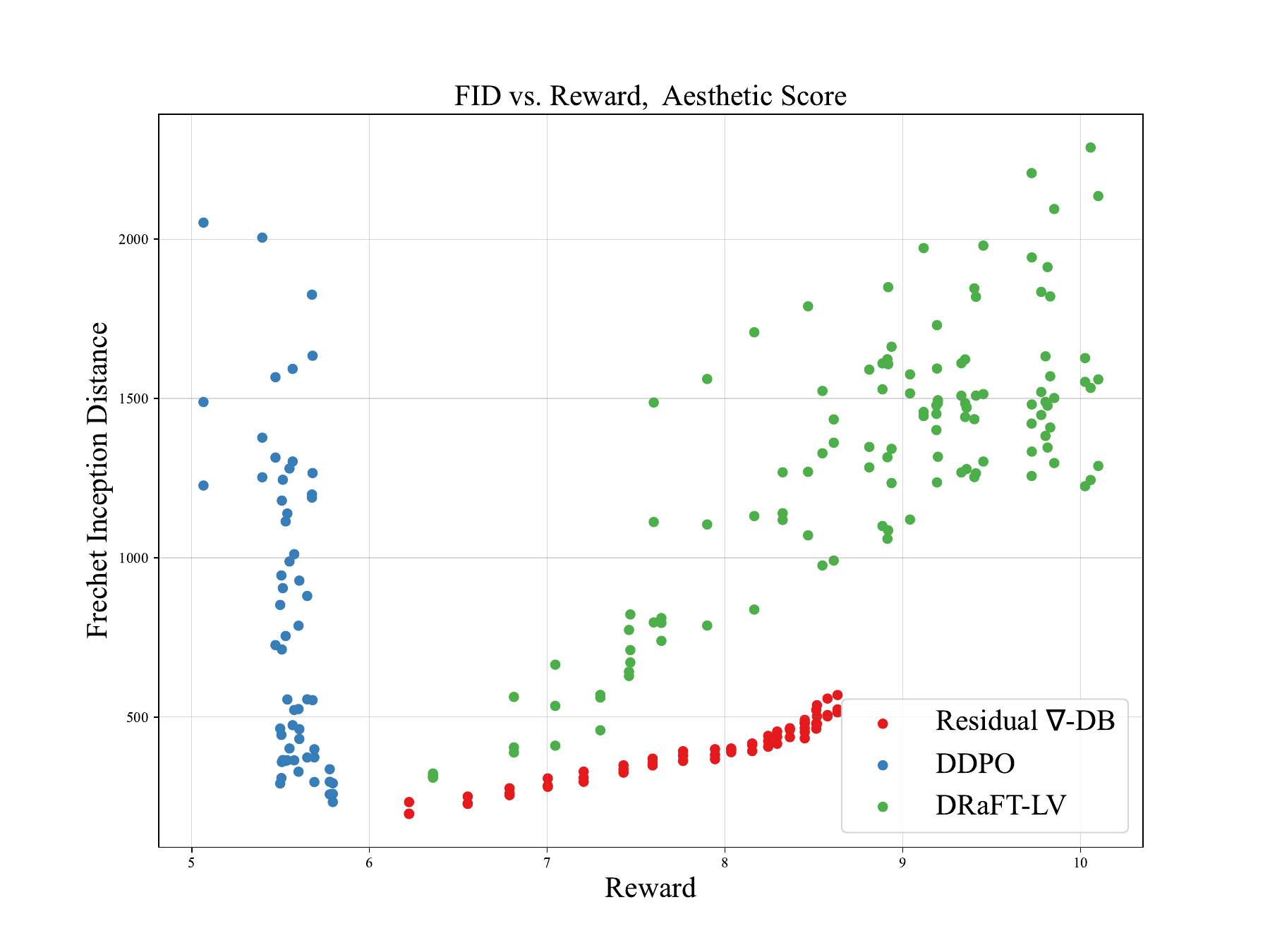}
    \vspace{-3mm}
    \caption{\footnotesize Pareto frontiers for reward, diversity and prior-preservation (measured by FID) on Aesthetic Score with the MDP constructed by SDE-DPM-Solver++ (with 20 inference steps).}
    \label{fig:pareto_dpm}
    \vspace{-2mm}
\end{figure}

\clearpage
\newpage
\section{Additional ablation experiments}
\label{sec:more_ablation}

\textbf{Effect of different prior strengths.}
It is often of interest to have slightly weaker prior instead of simply increasing the reward strength, of which the sampling objective is $R(x_T)^\beta P_F^\#(x_T)^\eta$. While it is hard to obtain an exact estimate of $P_F^\#(x_T)^\eta$ where $\eta \in (0, 1]$ is the prior strength, one can approximate it with the weighted score function $\eta \nabla_{x_t} \log P_F^\#(x_t, t)$. Denoting $F_\eta(x_t)$ the corresponding flow, we have the modified \resgraddb conditions.

\begin{align}
\label{eqn:grad-db-res-cond-eta}
        \underbrace{
                    \nabla_{x_{t+1}}\log P_F(x_{t+1} | x_{t})  \!-\! \eta \nabla_{x_{t+1}}\log P_F^\#  (x_{t+1} | x_{t}) 
            }_{\scalebox{0.8}{$\nabla_{x_{t+1}} \log \tilde{P}_F(x_{t+1} | x_t)$}: \ \text{\small residual policy score function}}
        \!=\!
        \underbrace{
                \nabla_{x_{t+1}} \log F(x_{t+1}) \!-\! \nabla_{x_{t+1}} \log F^\#_\eta(x_{t+1}) 
        }_{\scalebox{0.8}{$\nabla_{x_{t+1}} \log \tilde{F}(x_{t+1})$}: \ \text{\small residual flow score function}}.
\end{align}

\begin{align}
\label{eqn:grad-db-res-reverse-cond-eta}
        \underbrace{
                    \nabla_{x_{t}}\log P_F(x_{t+1} | x_{t})  \!-\! \eta \nabla_{x_{t}}\log P_F^\#  (x_{t+1} | x_{t}) 
            }_{\scalebox{0.8}{$\nabla_{x_{t}} \log \tilde{P}_F(x_{t+1} | x_t)$}: \ \text{\small reverse residual policy score function}}
        \!=\!
        \underbrace{
                \nabla_{x_{t}} \log F(x_{t}) \!-\! \nabla_{x_{t}} \log F^\#_\eta(x_{t+1}) 
        }_{\scalebox{0.8}{$\nabla_{x_{t}} \log \tilde{F}(x_{t})$}: \ \text{\small residual flow score function}}.
\end{align}

\begin{figure}[h]
    \centering
    \vspace{-1pt}
    \adjustbox{valign=t, max width=0.98\linewidth}{%
        \includegraphics[width=0.5\linewidth]{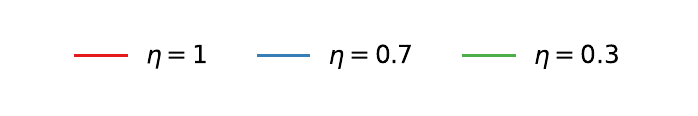}%
    }
    \vspace{-1.5em}

    \includegraphics[width=0.33\linewidth]{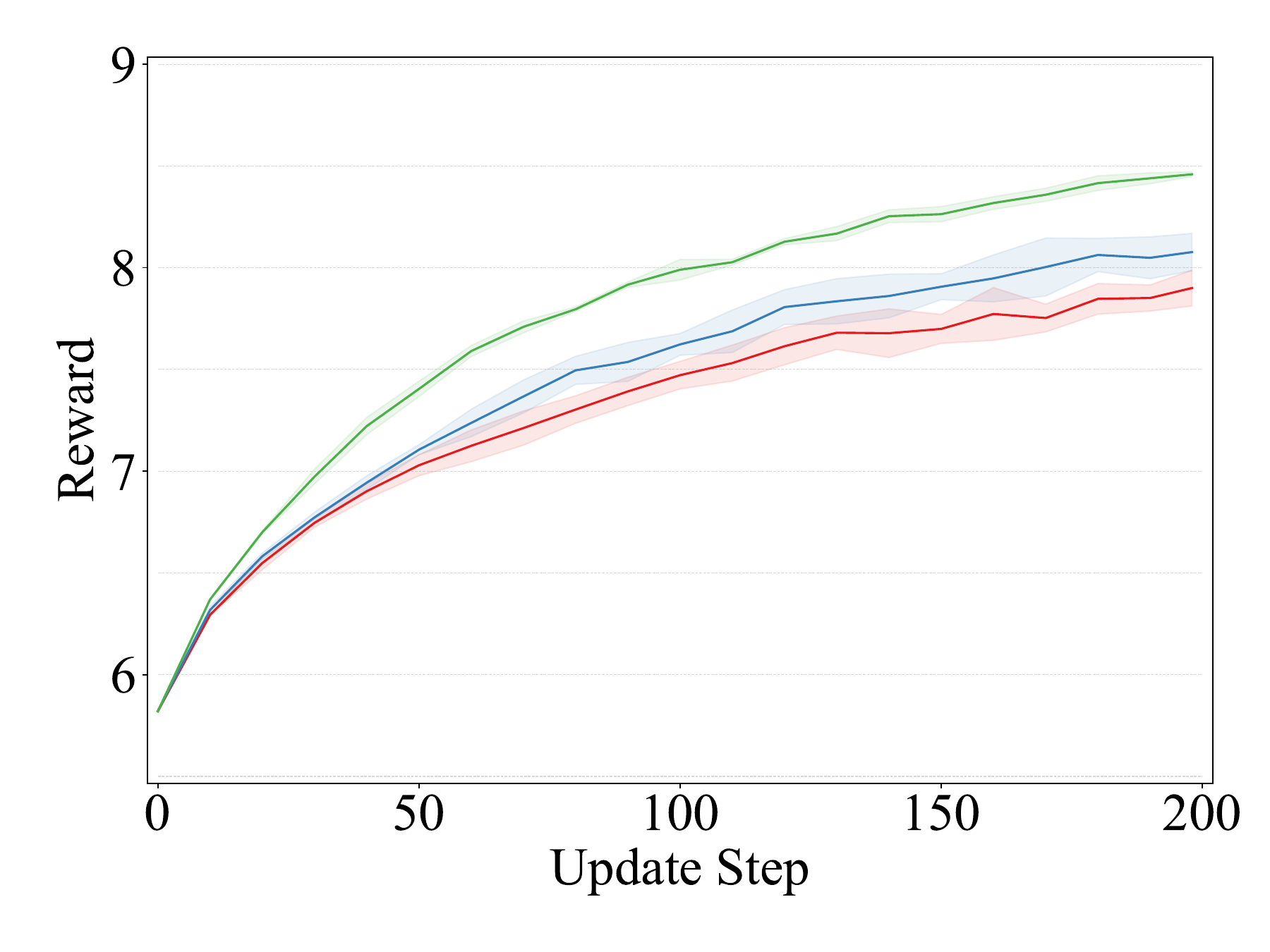}
    \hspace{-0.8em}
    \includegraphics[width=0.33\linewidth]{figs/aesthetic_eta_diversity.pdf}
    \hspace{-0.8em}
    \includegraphics[width=0.33\linewidth]{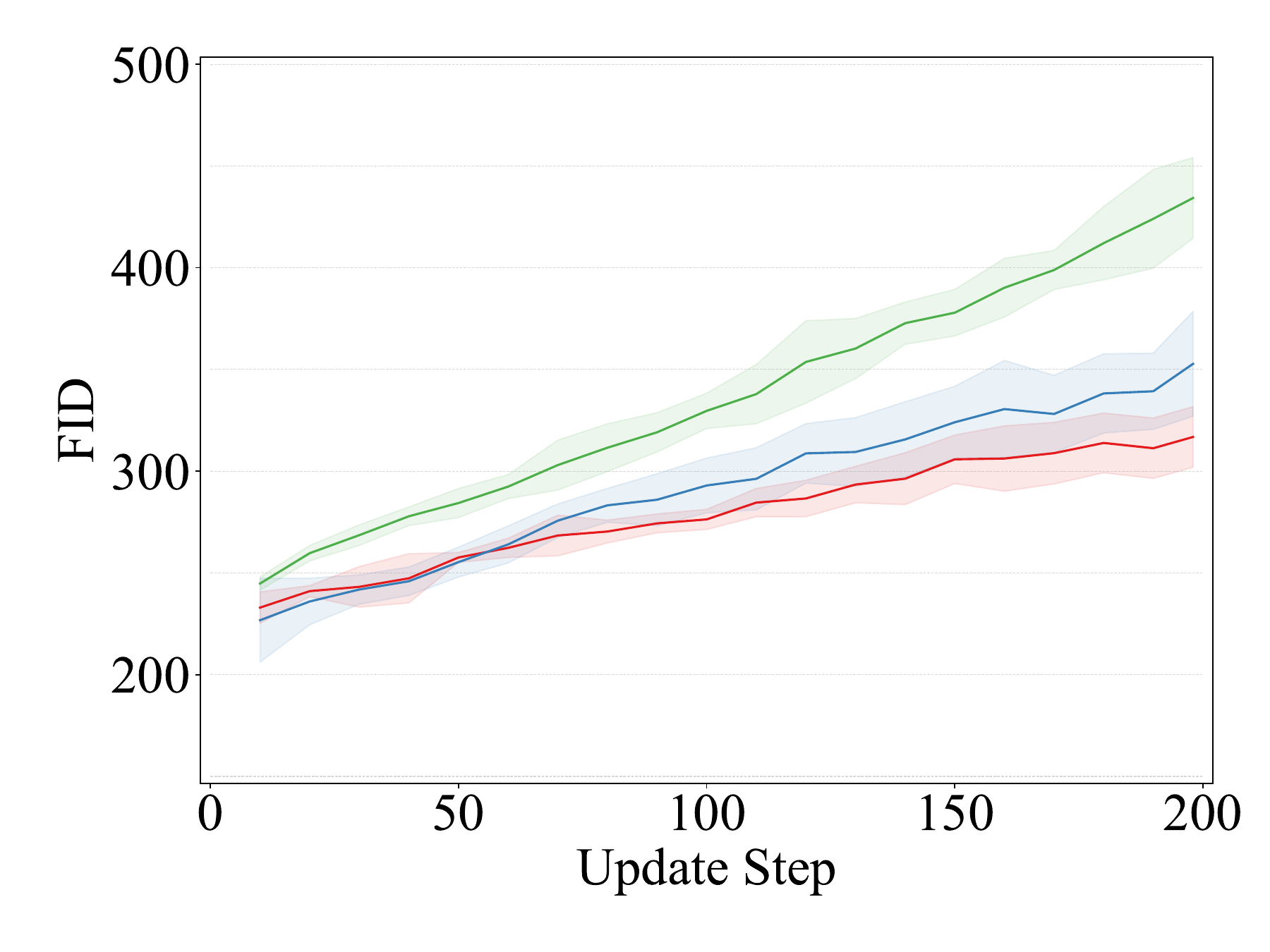}
    \caption{\footnotesize 
        Convergence curve of metrics of different methods throughout the finetuning process on Aesthetic Score with different prior strength $\eta$.
    }
    \label{fig:eta_evo}
\end{figure}

\begin{figure}[h]
    \vspace{-2mm}
    \centering
    \includegraphics[width=0.48\linewidth]{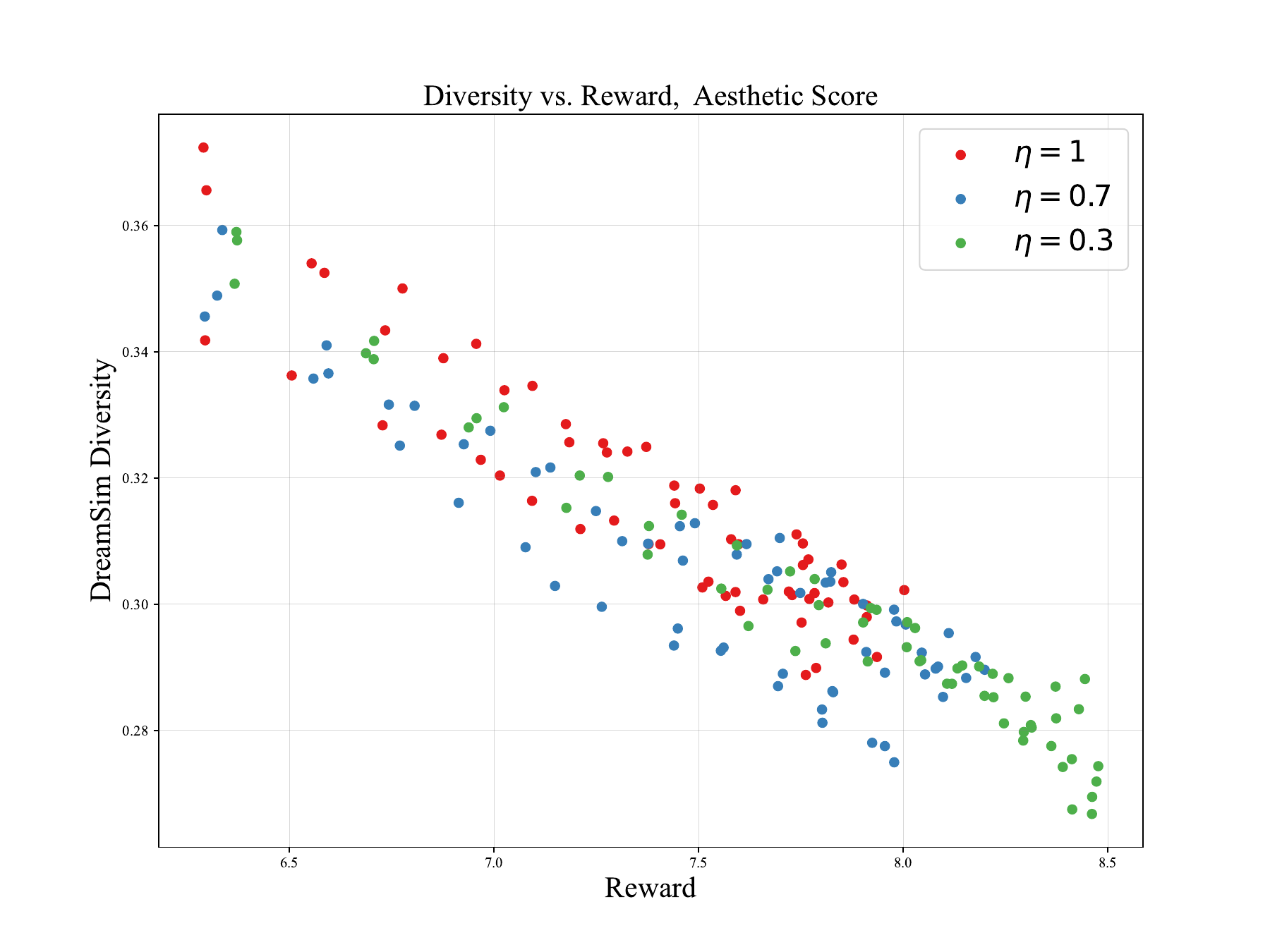}
    \hspace{-5mm}
    \includegraphics[width=0.48\linewidth]{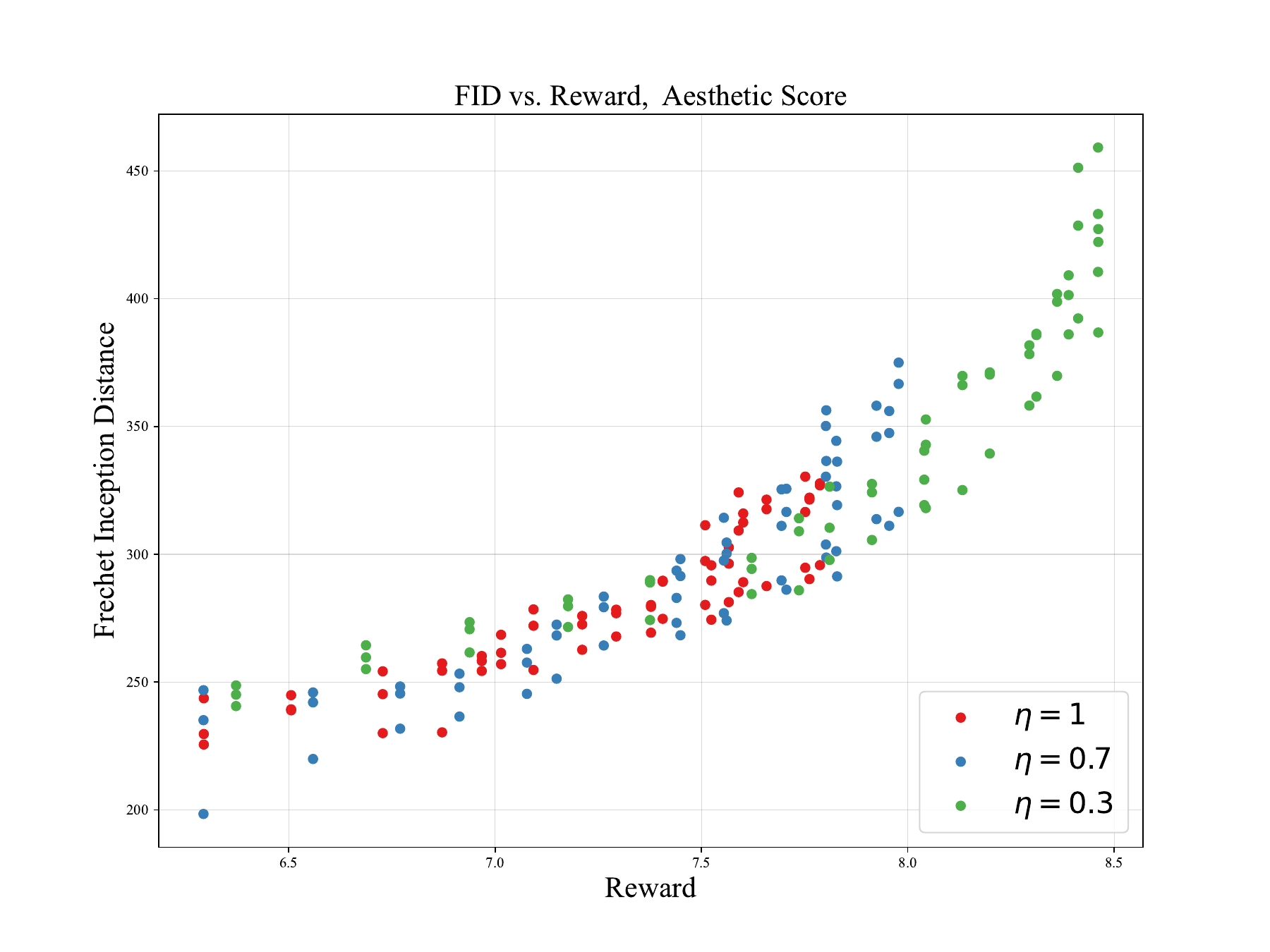}
    \vspace{-3mm}
    \caption{\footnotesize 
        Pareto frontiers for reward, diversity and prior-preservation (measured by FID) on Aesthetic Score with different prior strength $\eta$.
    }
    \label{fig:pareto_eta}
    \vspace{-2mm}
\end{figure}

We experiment with choices of prior strengths $\eta$ and observed in Fig.~\ref{fig:eta_evo} and \ref{fig:pareto_eta} that lower $\eta$ lead to better diversity-reward trade-off and faster reward convergence.

\begin{figure}[h]
    \centering
    \vspace{-1pt}
    \adjustbox{valign=t, max width=0.98\linewidth}{%
        \includegraphics[width=0.5\linewidth]{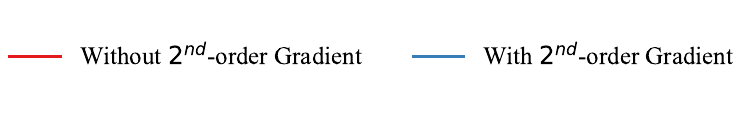}%
    }
    \vspace{-1em}

    \includegraphics[width=0.33\linewidth]{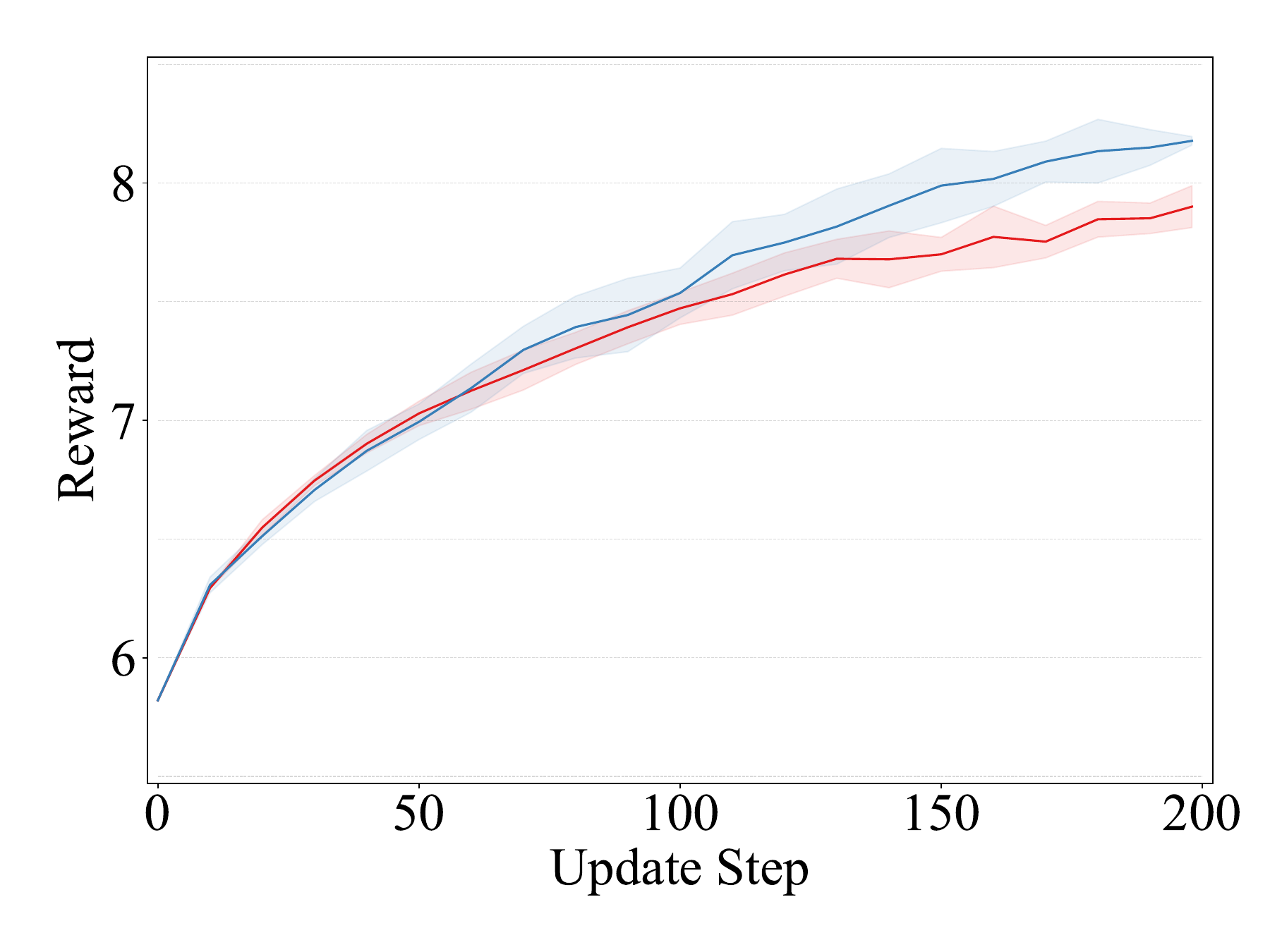}
    \hspace{-0.8em}
    \includegraphics[width=0.33\linewidth]{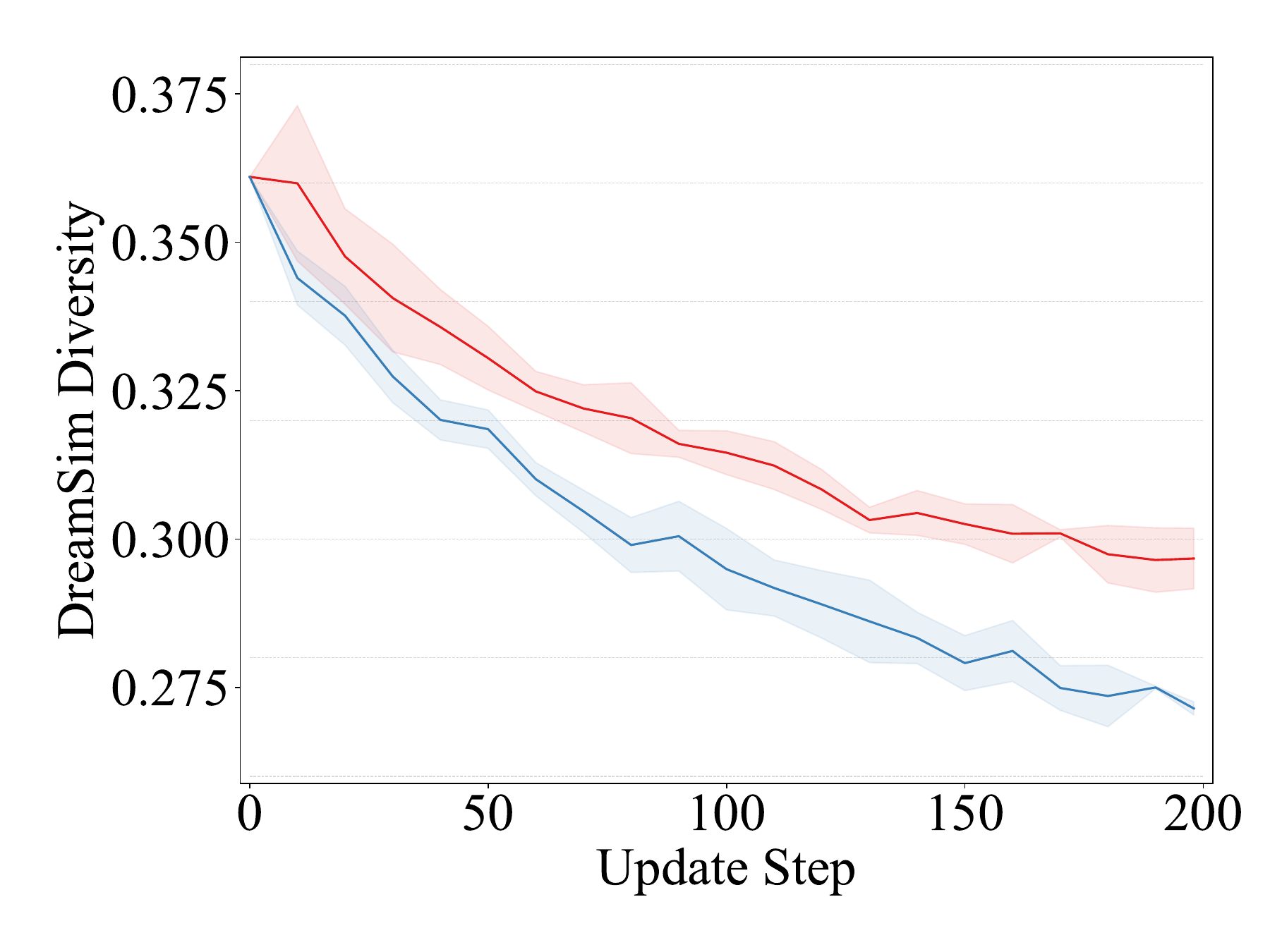}
    \hspace{-0.8em}
    \includegraphics[width=0.33\linewidth]{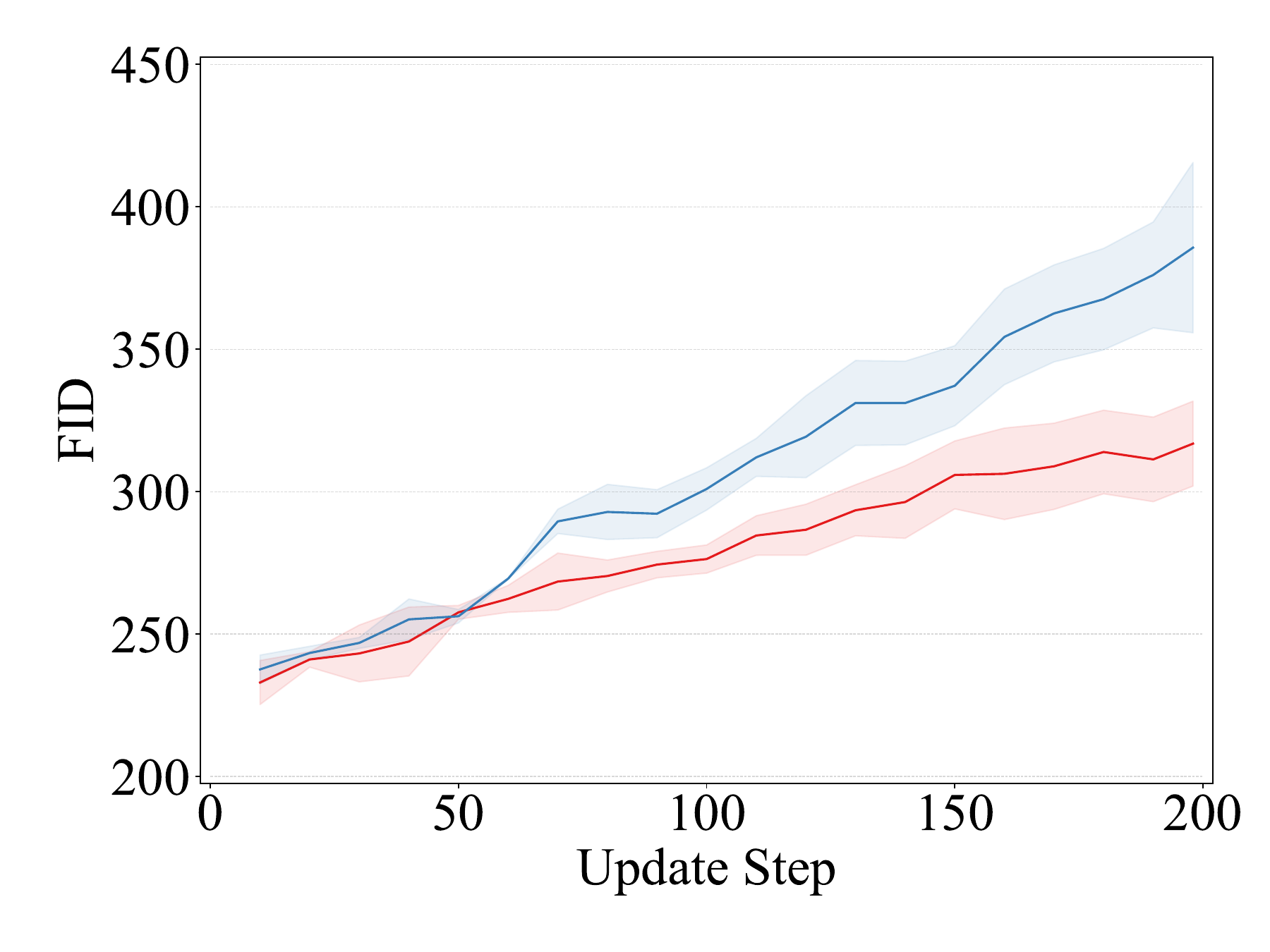}
    \caption{\footnotesize 
        Convergence curve of metrics of different methods throughout the finetuning process on Aesthetic Score with and without 2nd-order gradients.
    }
    \label{fig:aesthetic_2ndgd}

\end{figure}

\begin{figure}[h]
    \vspace{-2mm}
    \centering
    \includegraphics[width=0.48\linewidth]{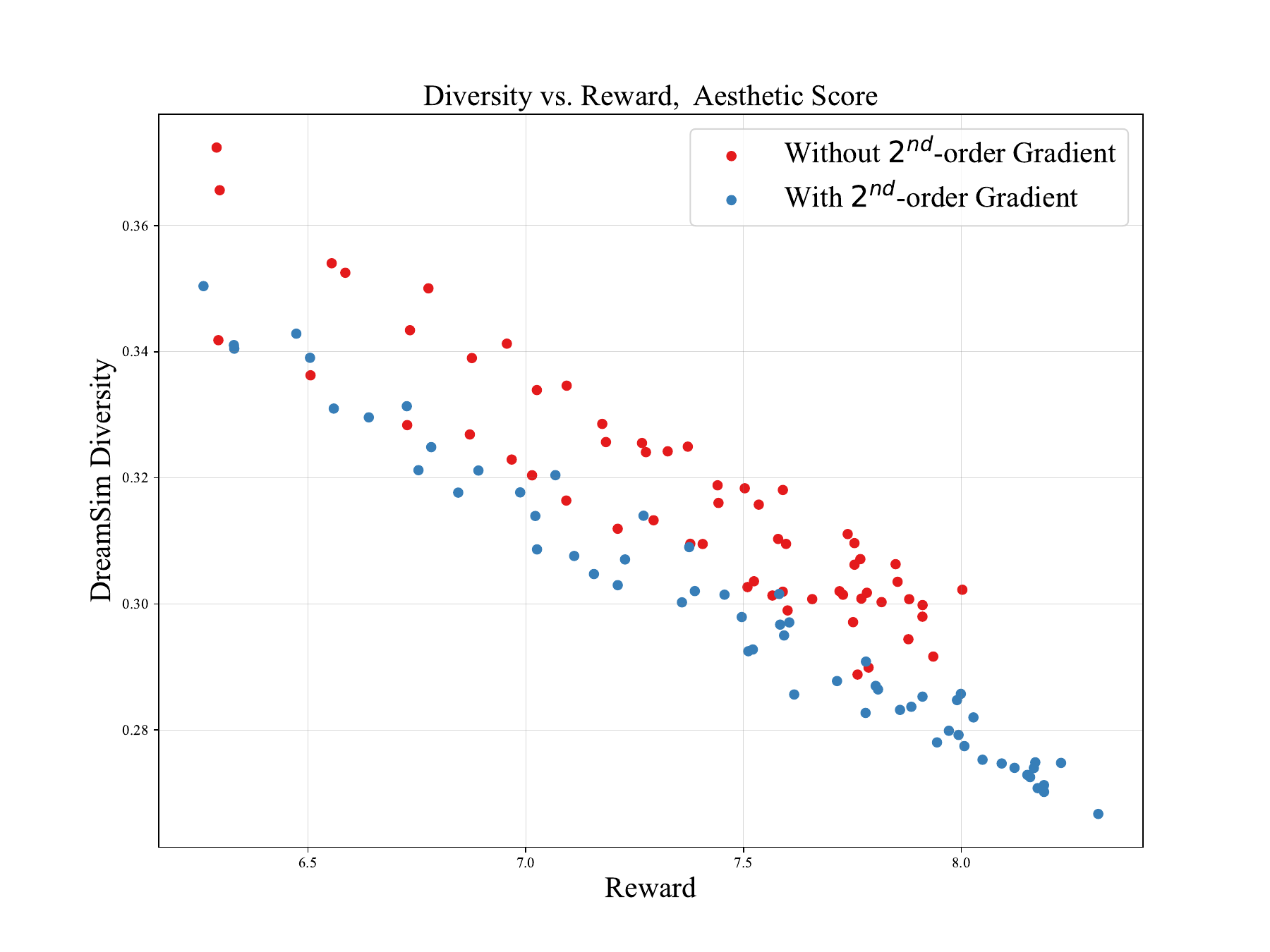}
    \hspace{-5mm}
    \includegraphics[width=0.48\linewidth]{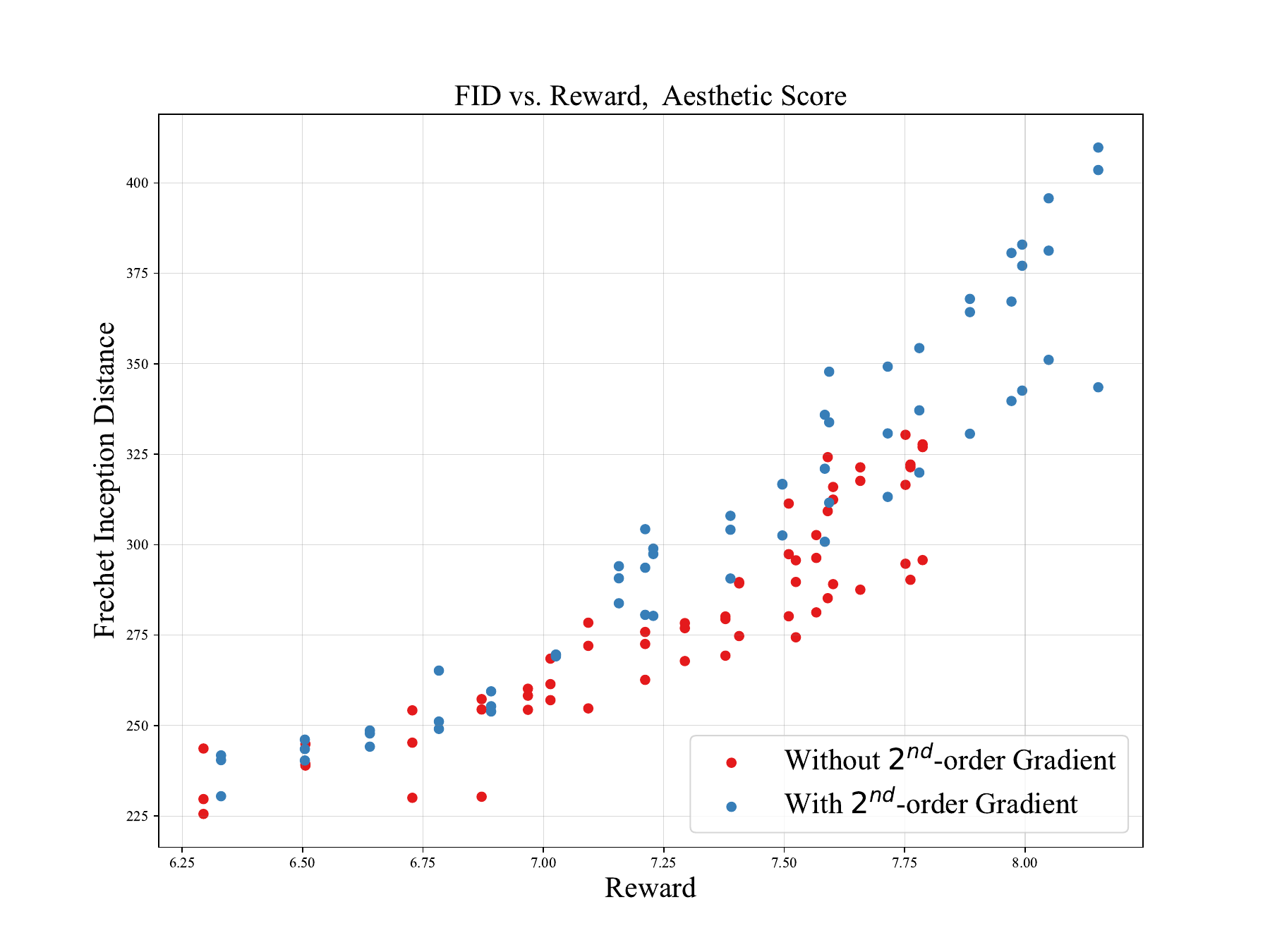}
    \vspace{-3mm}
        \caption{\footnotesize Pareto frontiers for reward, diversity and prior-preservation (measured by FID) on Aesthetic Score with and without 2nd-order gradients.}
    \label{fig:pareto_2ndgd}
    \vspace{-2mm}
\end{figure}

\textbf{Effect of 2nd-order gradients in finetuning.} In Fig.~\ref{fig:aesthetic_2ndgd} and \ref{fig:pareto_2ndgd}, we show the comparison between models with and without 2nd-order gradients, where both models are trained with $\beta=10000$. Empirically, 2nd-order gradients hurts the trade-off between reward convergence, diversity preservation and prior preservation.

\clearpage

\clearpage
\pagebreak
\newpage
\section{More samples (Aesthetic Score)}

\begin{figure}[h]
    \centering

    \includegraphics[width=0.98\linewidth]{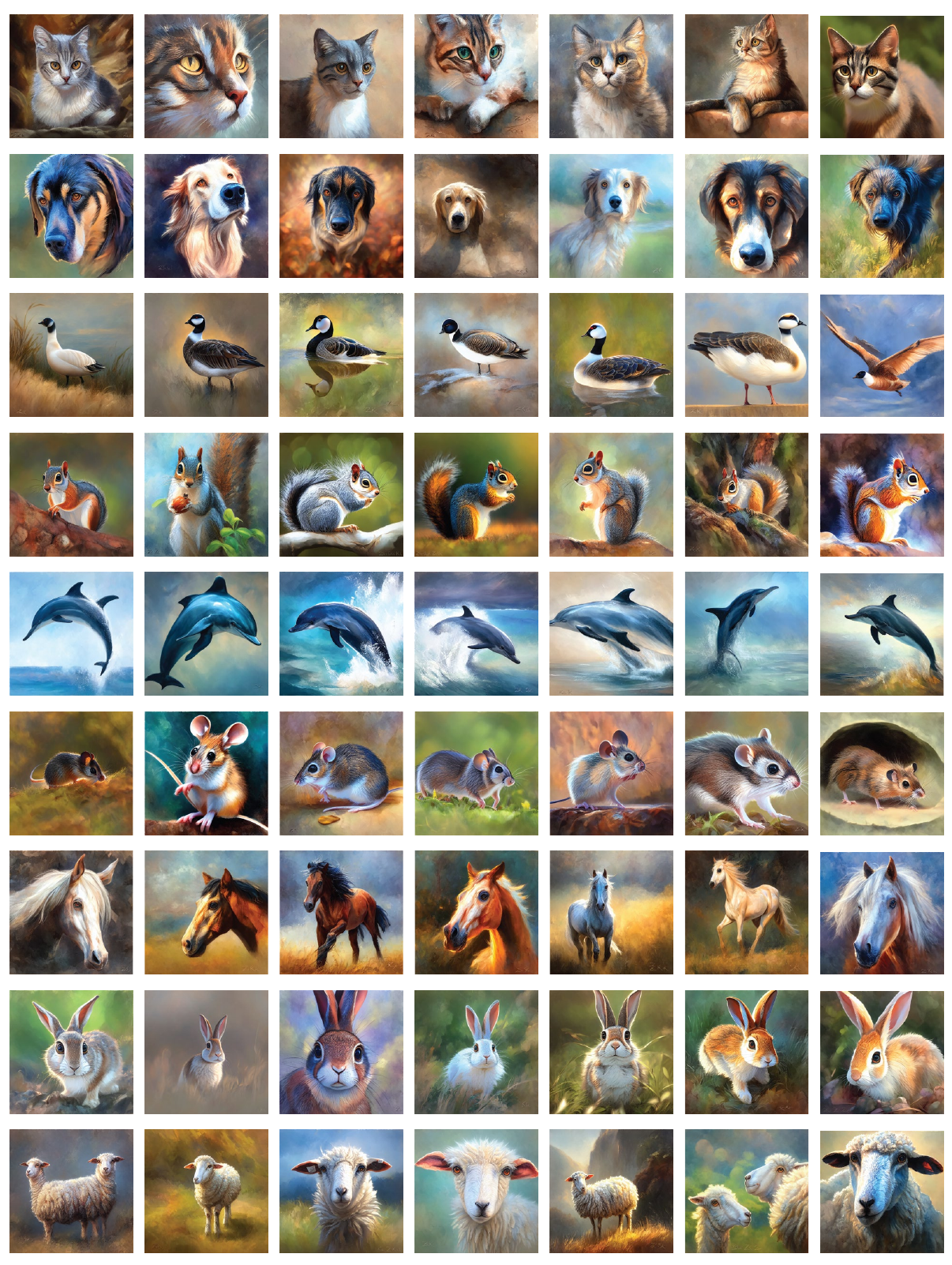}
    \hspace{-0.8em}
    \caption{\footnotesize 
        Additional uncurated samples from the model finetuned with \resgraddb on the reward model of Aesthetic Score.
    }
    \label{fig:additional_uncurated}

\end{figure}

\newpage
\section{More samples (HPSv2)}

\begin{figure}[h]
    \centering

    \includegraphics[width=0.98\linewidth]{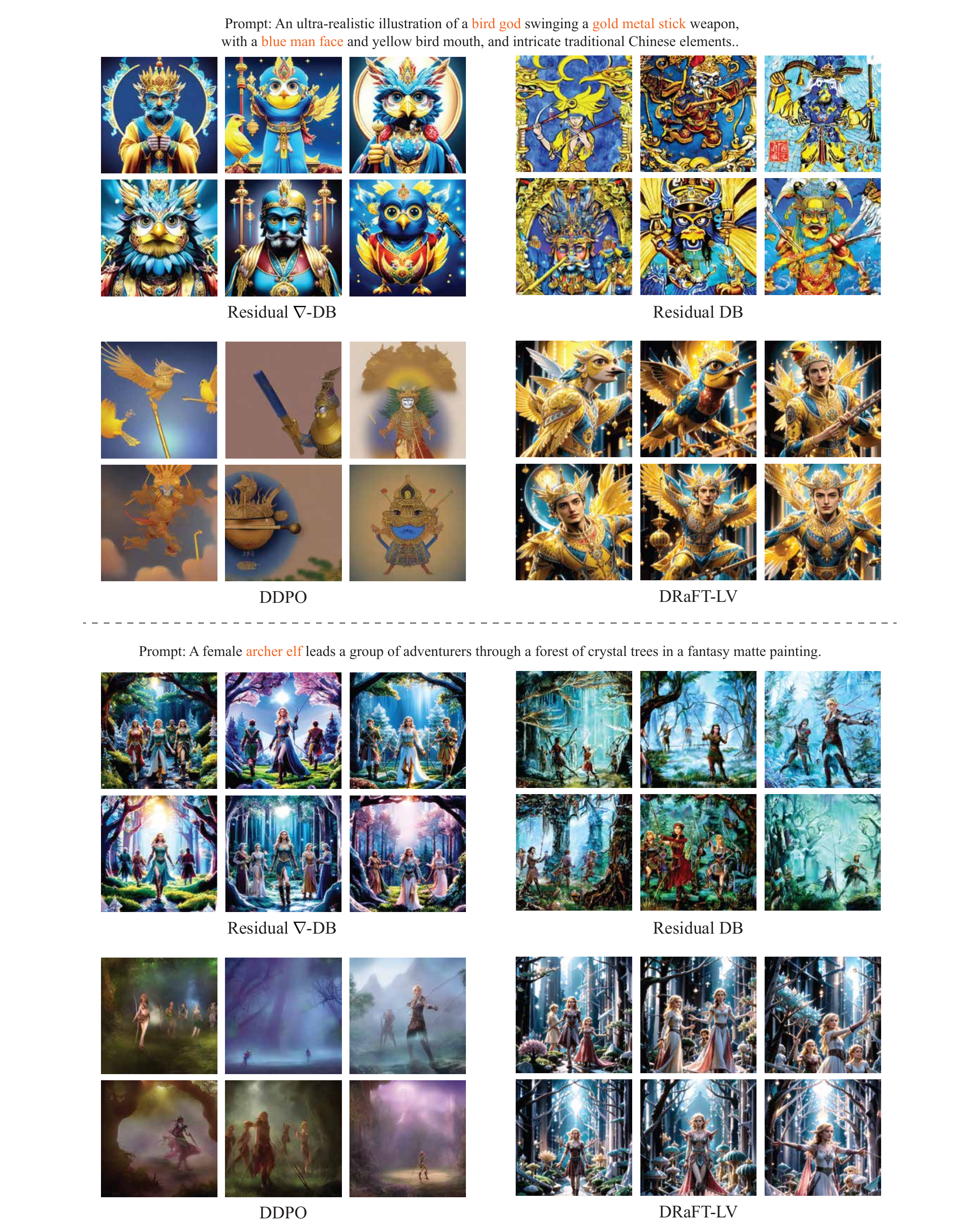}
    \hspace{-0.8em}
    \caption{\footnotesize 
        More comparison between samples generated by \resgraddb and the baseline methods. The model finetuned with \resgraddb is capable of following the instructions while generating diverse samples.
    }
    \label{fig:hps_appendix_page1}

\end{figure}

\begin{figure}[h]
    \centering

    \includegraphics[width=0.98\linewidth]{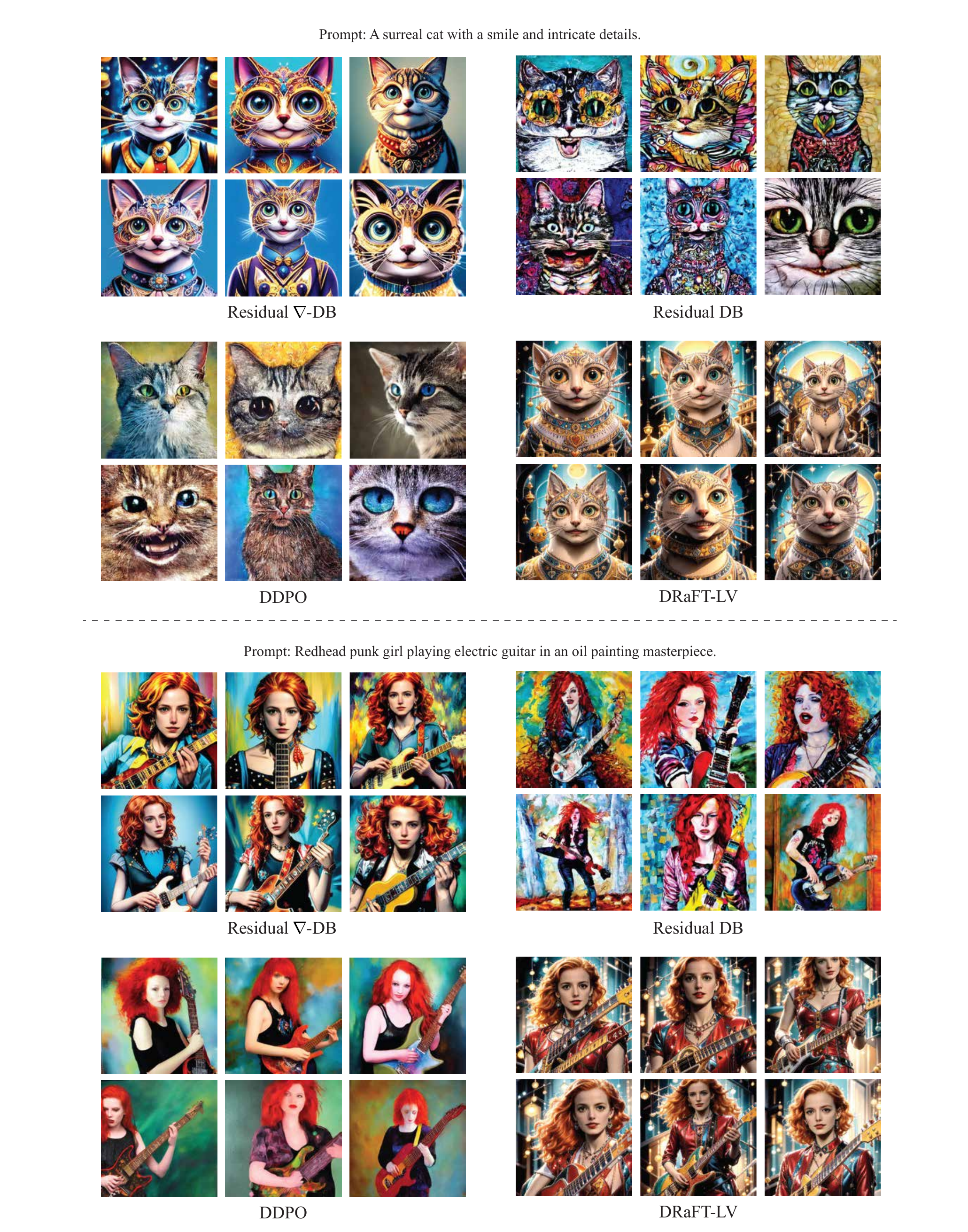}
    \hspace{-0.8em}
    \caption{\footnotesize 
        HPSv2 samples, Continued.
    }
    \label{fig:hps_appendix_page2}

\end{figure}

\clearpage
\newpage
\section{Qualitative comparison on diversity of samples generated by different methods}
\label{sec:diversity_qualitative}

Below we show uncurated samples generated by models finetuned with different methods with propmts \textit{cat, bird, rabbit, bird, kangaroo}.

\begin{figure}[h]
    \centering

    \includegraphics[width=\linewidth]{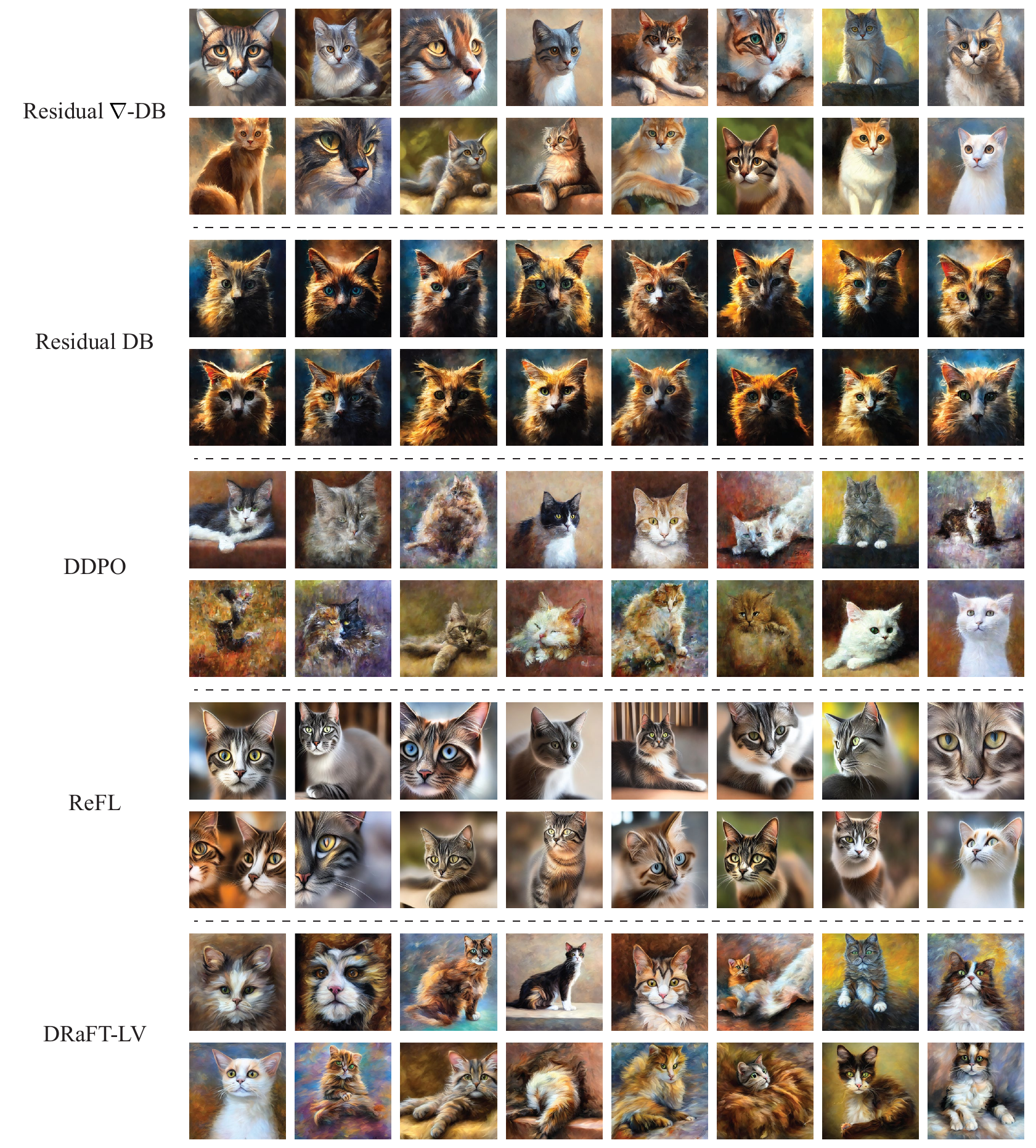}
    \caption{\footnotesize 
        Uncurated samples generated by models finetuned with different methods with prompt \textit{cat}.
    }
    \label{fig:diversity_cat}
\end{figure}

\begin{figure}[h]
    \centering

    \includegraphics[width=\linewidth]{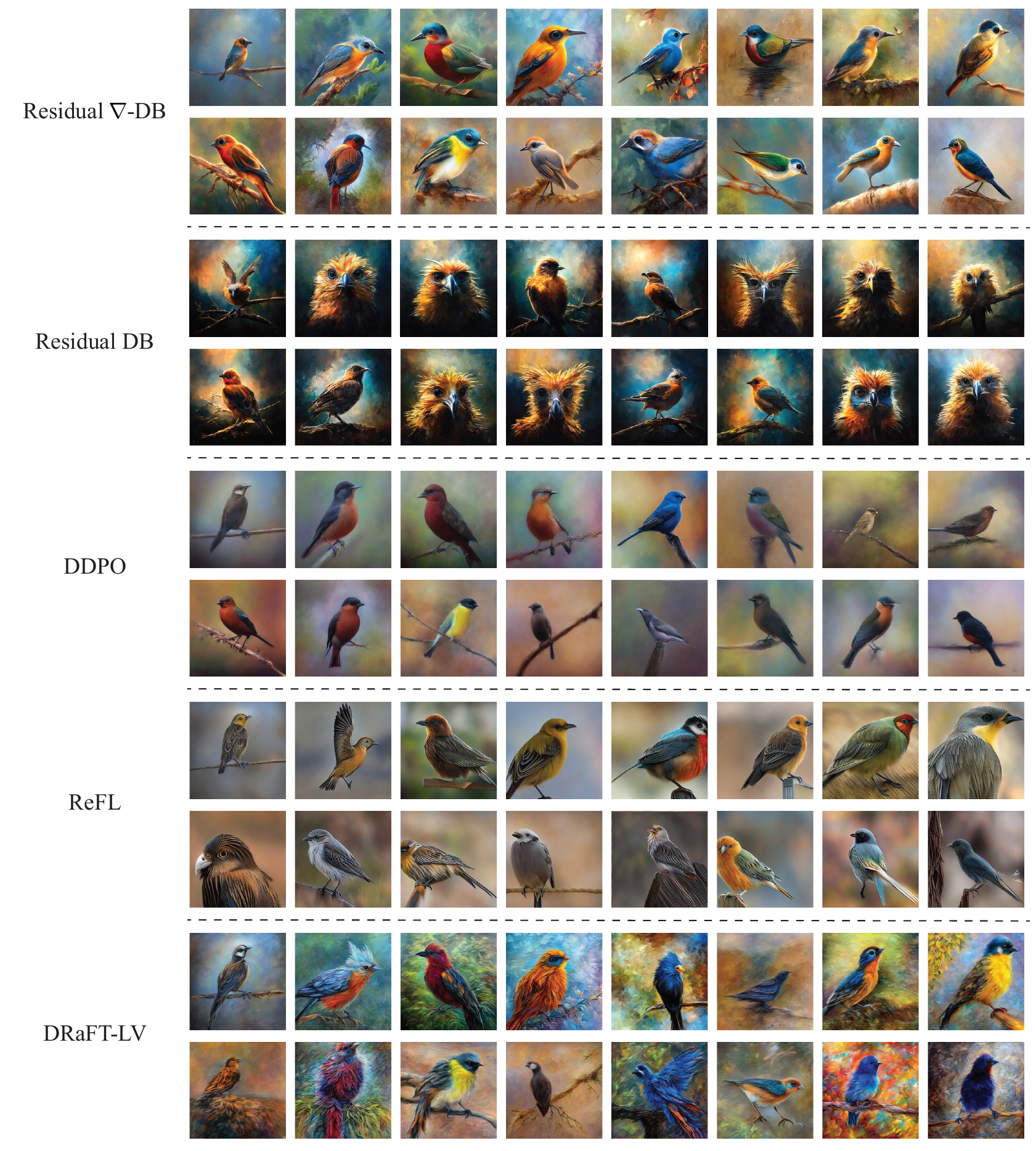}

    \caption{\footnotesize 
        Uncurated samples generated by models finetuned with different methods with prompt \textit{bird}.
    }
    \label{fig:diversity_bird}
\end{figure}

\begin{figure}[h]
    \centering

    \includegraphics[width=\linewidth]{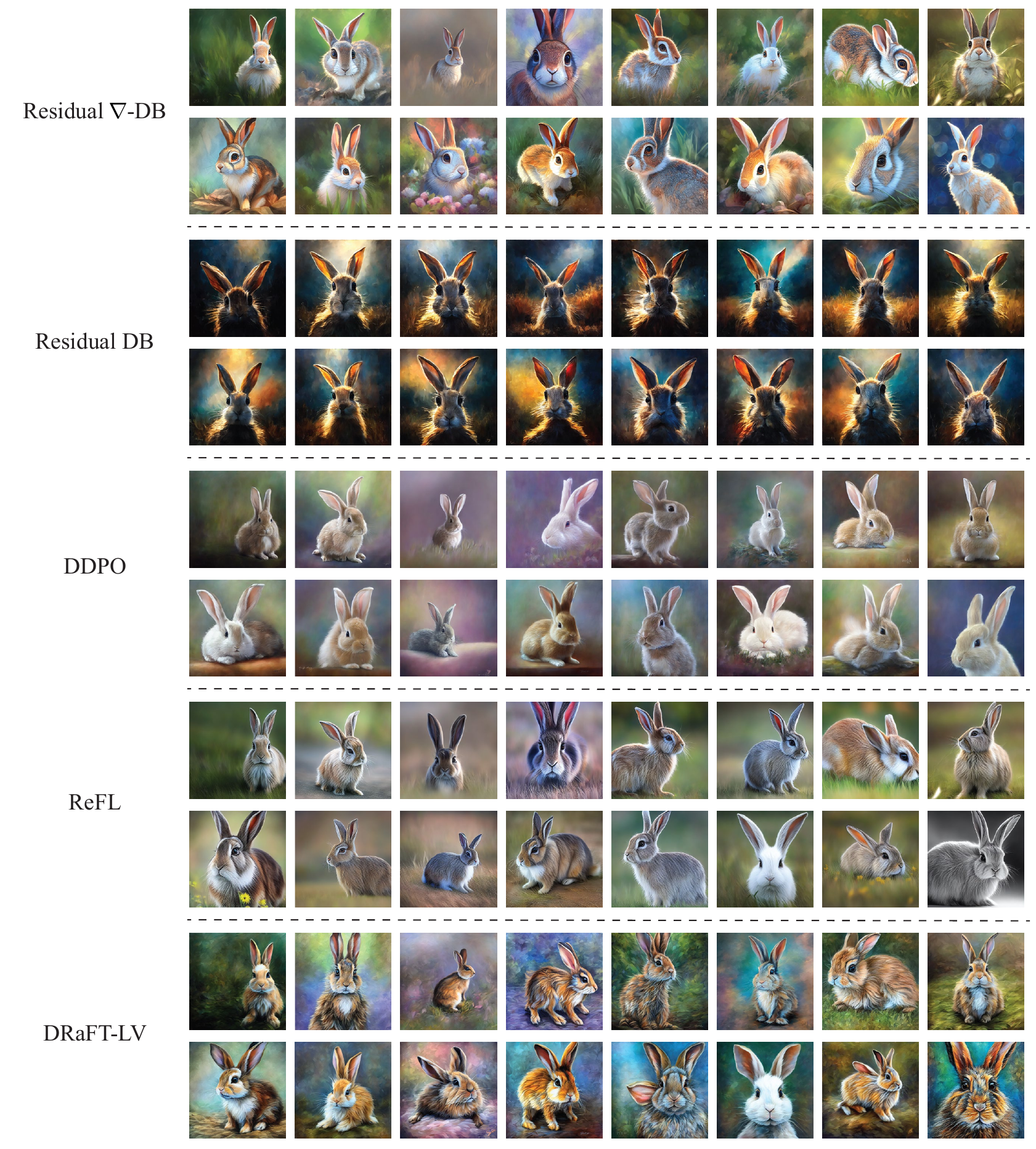}

    \caption{\footnotesize 
        Uncurated samples generated by models finetuned with different methods with prompt \textit{rabbit}.
    }
    \label{fig:diversity_rabbit}
\end{figure}

\begin{figure}[h]
    \centering

    \includegraphics[width=\linewidth]{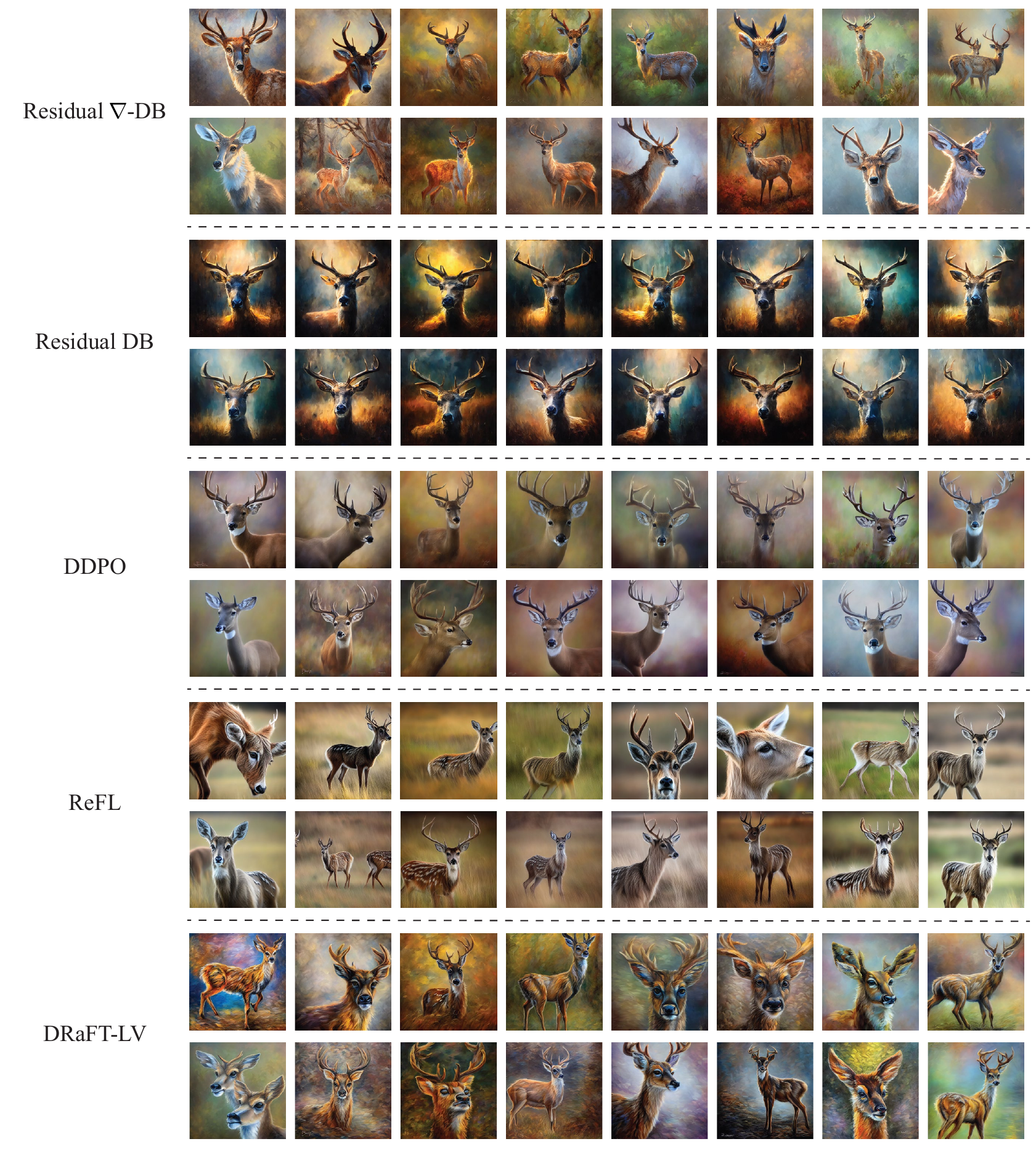}

    \caption{\footnotesize 
        Uncurated samples generated by models finetuned with different methods with prompt \textit{deer}.
    }
    \label{fig:diversity_deer}
\end{figure}

\begin{figure}[h]
    \centering
    \includegraphics[width=\linewidth]{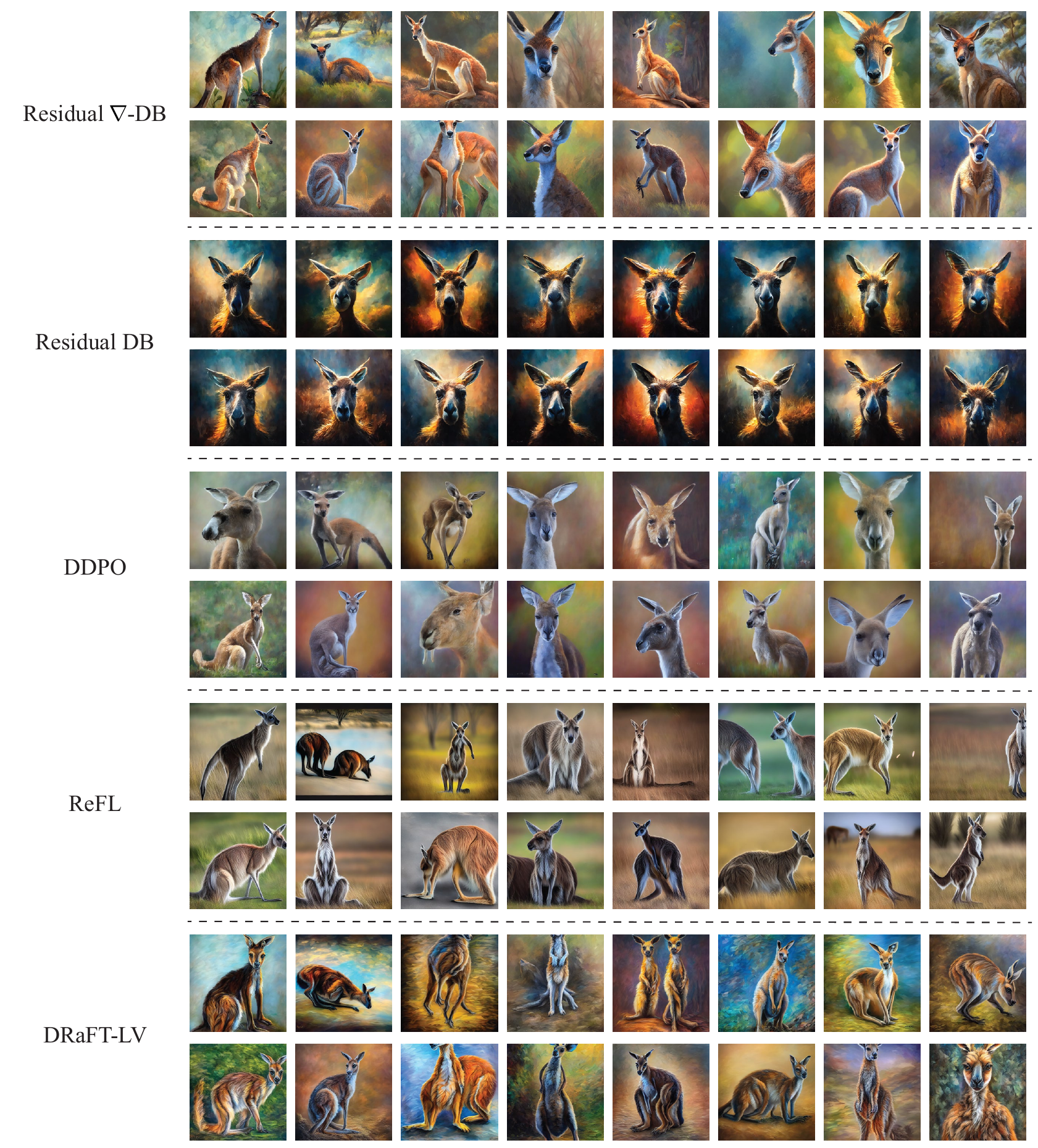}
    \caption{\footnotesize 
        Uncurated samples generated by models finetuned with different methods with prompt \textit{kangaroo}.
    }
    \label{fig:diversity_kangaroo}
\end{figure}

\end{document}